\documentclass{article}
\usepackage{amsmath,amsthm,graphicx,mlspconf}

\usepackage{epstopdf}
\usepackage{algorithm}
\usepackage{algorithmicx}
\usepackage{algpseudocode}
\usepackage{verbatim}
\usepackage{color}
\usepackage{url}
\usepackage{placeins}
\newcommand{\colorlegend}{\textcolor{red}{$\boldsymbol{\circ}$} ASAP LOE, \textcolor{blue}{$\boldsymbol{\times}$} LOE BFGS, \textcolor{green}{$\boldsymbol{\square}$} LE , \textcolor{cyan}{$\boldsymbol{\star}$} LOE MM }

\newtheorem{theorem}{Theorem}

\usepackage{mathtools}
\newcommand\mydef{\stackrel{\mathclap{\small\mbox{def}}}{=}}

\newcommand{\argmin}{\mbox{argmin}}
\newcommand{\argmax}{\mbox{argmax}}
\usepackage{amsfonts}
\usepackage{amsmath}
\usepackage{amssymb}
\newcommand{\R}{\mathbb{R}}
\newcommand{\bas}[1]{\begin{align*}#1\end{align*}}

\newcommand{\removed}[1]{\iffalse #1 \fi}

\graphicspath{{figs/}}

\title{Point Localization and Density Estimation from Ordinal kNN graphs using Synchronization}
\name{Mihai Cucuringu, Joseph Woodworth
}
\address{UCLA Mathematics Department}
\begin{document}

\maketitle

\begin{abstract} 
We consider the problem of embedding unweighted, directed k-nearest neighbor graphs in low-dimensional Euclidean space.  The k-nearest neighbors of each vertex provide ordinal information on the distances between points, but not the distances themselves.
Relying only on such ordinal information, 
 along with the low-dimensionality,
we recover the coordinates of the points up to arbitrary similarity transformations (rigid transformations and scaling).
Furthermore, we also illustrate the possibility of robustly recovering the underlying density via the Total Variation Maximum Penalized Likelihood Estimation (TV-MPLE) method.
We make existing approaches scalable by using an instance of a local-to-global algorithm based on group synchronization, recently proposed in the literature in the context of sensor network localization, 
 and structural biology,
which we augment with a scale synchronization step. 
We show our approach compares favorably to the recently proposed Local Ordinal Embedding (LOE) algorithm even in the case of smaller sized problems, and also demonstrate its scalability on large graphs.
The above divide-and-conquer paradigm can be of independent interest to the machine learning community when tackling geometric embeddings problems.
\end{abstract}

\vspace{-1mm}
\vspace{-2mm}

\newlength{\thirdcolwidth}
\setlength{\thirdcolwidth}{.31\columnwidth}
\newlength{\halfcolwidth}
\setlength{\halfcolwidth}{.47\columnwidth}
\newlength{\thirdtextwidth}
\setlength{\thirdtextwidth}{.32\textwidth}
\newlength{\sixthtextheight}
\setlength{\sixthtextheight}{.15\textwidth}
\newlength{\fourthcolwidth}
\setlength{\fourthcolwidth}{.24\columnwidth}

\section{Introduction} 

\vspace{-2mm}

Embedding unweighted $k$-nearest neighbor (kNN) graphs is a special case of ordinal or non-metric embedding, where one seeks a spatial embedding of $n$ points $\left\{\vec x_i\right\}_{i=1}^n$ in $\R^d$ such  that  
\begin{align}
\forall (i_1,j_1,i_2,j_2)\in \mathcal{C},\hspace{.1in}
\|\vec x_{i_1} -\vec x_{j_1}\|_2
<
\|\vec x_{i_2} -\vec x_{j_2}\|_2,
\end{align}
where $\mathcal{C}$ denotes the set of ordinal constraints.  Ordinal constraints are sometimes also specified as triplets \cite{agarwal2007generalized}.   
In the case of unweighted kNN graph embedding, $\mathcal{C}=\mathcal{C}(G)=\left\{(a,b,a,c)\big| ab\in E(G), ac\not\in E(G)\right\},$ where $E(G)$ is the set of directed edges in the kNN graph $G$.

Graph-based methods are of utmost importance in several modern machine learning methods with applications such as clustering, dimensionality reduction, and ranking. Many such methods rely on weighted graphs, with weights often based on similarity functions, i.e., $w_{ij}=S(x_i,x_j)$.  From a practical standpoint, storing unweighted kNN graphs instead would allow for a very sparse representation of the data.  If one could recover the source data $\{x_i\}_{i=1}^n$ from unweighted kNN graphs, such a computationally efficient sparser representation would incur no information loss.  Because of the extreme sparsity of the representation, this is generally a hard problem. Just recently, a  method for recovering data distributions from unweighted kNN graphs was introduced 
in \cite{von2013density}. 
 Another motivation for this problem comes from an instance of the popular sensor network localization problem, where each sensor is able to transmit only limited connectivity information to a central location (ID names of its k nearest neighbors), but  transmits neither the  distance measurements nor a complete list of all its neighbors within a given fixed radius. 

The original work on this problem dates back to Shepard \cite{shepard1962aanalysis} 
and Kruskal \cite{kruskal1964multidimensional, kruskal1964nonmetric}, and lately has been studied intensively in the machine learning literature \cite{agarwal2007generalized, shaw2009structure, mcfee2009partial, quist2004distributional, tamuz2011adaptively,jamieson2011low,rosales2006learning, ailon2011active, ouyang2008learning, mcfee2010metric, jamieson2011active, lan2012statistical, wauthier2013efficient}. 
In this work, we  compare  against and extend the recent Local Ordinal Embedding method \cite{terada2014local}, which enjoys several favorable comparisons with other modern methods.
 Another motivation for this problem comes from an instance of the popular sensor network localization problem, where each sensor is able to transmit only limited connectivity information to a central location, in the form of ID names of its k nearest neighbor sensors, but  transmits neither the estimated distance measurements nor a complete list of all its neighbors within a given fixed radius. Note that either of these last two scenarios renders the localization problem (of estimating the sensor coordinates) easier to solve.  Similar to the sensor network application, one could potentially apply this framework to cooperative control and sensing involving swarms of robot micro-vehicles with limited payloads communicating via radio with limited bandwidth \cite{martinez2013economical, gonzalez2011third}.
%
%
Our key ingredient is a modified version of the  As-Synchronized-As-Possible (ASAP) algorithm introduced in  \cite{asap2d}, 
which makes  existing embedding methods scalable via a divide-and-conquer, 
 non-iterative local to global approach, reduces computational complexity, allows for massive parallelization of large problems, and  increases robustness to noise.
The ASAP algorithm introduced in \cite{asap2d}, on which we rely in the present paper, renders our approach to reconstruct kNN graphs scalable to graphs with thousands or even tens of thousands of nodes, and is an example of a local-to-global approach that integrates local ordinal information into a global embedding calculation.

We detail in Section~\ref{sec:breakintopatches} the exact approach used to decompose the initial kNN graph into many overlapping subgraphs, that we shall refer to as patches from now on. 
Each resulting patch is then separately embedded in a coordinate system of its own using an ordinal embedding algorithm, such as the recent Local Ordinal Embedding (LOE) algorithm \cite{terada2014local}. In the hypothetical scenario when LOE recovers the actual ground truth coordinates of each patch, such local coordinates agree with the global coordinates up to scaling and some unknown rigid motion (such as rotation, reflection and translation), in other words, up to a similarity transformation.
However, in most practical instances, it is unreasonable to expect that the LOE algorithm will recover the exact coordinates only from ordinal data. On a related note, we point out the recent work of Kleindessner and von Luxburg \cite{von2014uniqueness}, who settled a long-known conjecture claiming that, given knowledge of all ordinal constraints of the form 
$||x_i - x_j || < ||x_k - x_l ||$ between an unknown set of points $x_1, \ldots,x_n \in \mathbb{R^d}$ (for finite $n$), it is possible to approximately recover the ground truth coordinates of the points up to similarity transformations. 
 Furthermore, the same authors show that the above statement holds even when we only have \textbf{local} information such as the distance comparisons between points in small neighborhoods of the graphs, thus giving hope for a local-to-global approach, in the spirit of the one we propose in the present paper.


Our contributions are: \textbf{1.} We present a local-to-global approach for the problem of embedding clouds of points from ordinal information, which is scalable to very large graphs, and can be computed efficiently and robustly in a distributed manner.  
Specifically, we extend the ASAP framework to the setting of ordinal embeddings, by augmenting it with a scale synchronization step.
We believe that local-to-global strategies could benefit many problems in the machine learning community.  The scale of data involved in many interesting problems poses a challenge to direct, holistic approaches.
\textbf{2.}  We extend the ordinal embedding pipeline to perform density estimation via Total Varation Maximum Penalized Likelihood Estimation.  This demonstrates the similarity between the point localization and density estimation problems.  Sufficiently simple point distributions can be well estimated by applying a short postprocessing step to an approximate embedding.
\textbf{3.} We present preliminary results for a very simple, straightforward ordinal embedding method.  

The rest of the paper is organized as follows. 
Section~\ref{sec:RelWork} is a summary of existing methods for related embedding problems.
Section \ref{sec:ASAPordinal} details the pipeline of the ASAP framework, including the scale synchronization step in Section~\ref{sec:scalingSync}.
In Section~\ref{sec:densityEst} we remark on the connection to the density estimation problem, and describe the post-processing step performed via Total-Variation Maximum Penalized Likelihood Estimation.
Section~\ref{sec:experiments} shows the results of several experiments recovering point embeddings from a variety of data sets, and compares to the existing LOE algorithm, as well as presenting results for the density estimation problem. 
In Section~\ref{sec:lp} we discuss an entirely different approach to ordinal embedding, and present some preliminary results which suggest more modifications are needed.
We conclude our primary discussion in Section~\ref{sec:conclusion} and summarize in Appendix \ref{sec:rigidity} some related basic notions from the rigidity theory literature.

\vspace{-4mm}
\section{Related Work}  \label{sec:RelWork}
\vspace{-3mm}
\subsection{Multidimensional Scaling}
\vspace{-2mm}
Broadly speaking, multidimensional scaling (MDS) refers to a number of related problems and methods.  In Classical Multidimensional Scaling (CMDS) \cite{torgerson1958theory}, one is given all Euclidean Squared-Distance 
measurements $\Delta_{ij}=\|\vec x_i-\vec x_j\|_2^2$ on a set of points $X=\{\vec x_i\}_{i=1}^n$ and wishes to approximate the points, assuming that they approximately lie in a low-dimensional space $d\ll n$.
Note that the solution for the coordinates is unique only up to rigid transformations, and that solutions do not exist for all possible inputs $\Delta$.  

One can generalize CMDS to incorporate additional nonnegative weights $W_{ij}$ on each distance, useful  when some distances are missing, or most distances are noisy, but some are known.  The optimization involves minimizing an energy known in the literature as \textit{stress} \cite{kruskal1978multidimensional}.
One approach to minimize stress is to iteratively minimize a majorizing function of two variables.  
A further generalization of MDS is non-metric MDS, or Ordinal Embedding, in which the input $D$ is assumed to be an increasing function applied to distance measurements \cite{shepard1962aanalysis}.
This may be the case if $D$ represents dissimilarity between points, as opposed to measured distances.  The problem can again be expressed with stress functionals and is usually solved with isotonic regresion \cite{kruskal1964nonmetric}.

\subsection{Semidefinite Programming methods}

Semidefinite Programming methods (SDP) have been applied frequently to MDS and related problems.  Classical MDS can be stated as an SDP, with a closed form solution.  Any formulation of the problem that optimizes over the Gram matrix requires the semidefinite constraint $K\in\mathbb{S}^n_+.$  Indeed, for metric MDS, if one penalizes the squared error on the squared distance measurements, the problem can be written as
\bas{
&\min_{X\in\R^{d\times n}} {\sum_{ij}W_{ij}(\Delta_{ij} -\Delta_{ij}(X) )^2 }\\
=&
\min_{K\in\mathbb{S}^n_+,X\in\R^{d\times n}} {\sum_{ij}W_{ij}(\Delta_{ij} -(K_{ii}-2K_{ij}+K_{jj}) )^2 }\\
&\mbox{ s.t. }K=X^TX.
}
Constraints of the form $K=X^TX$ are usually not allowed however, and are typically relaxed to  \cite{snlsdp, snlsdpArx}  
\bas{
\left[\begin{matrix} I & X \\ X^T & K\end{matrix}\right]\succeq 0.
}
via Schur's Lemma. 
  Furthermore, one encourages $K$ to be approximately low-rank by introducing a nuclear norm or trace penalty $\|K\|_*=\|\sigma(K)\|_1=tr(K),$ as a convex relaxation of a rank constraint.  Intuitively, since the $\ell_1$ norm promotes sparsity, the nuclear norm should promote few nonzero singular values.  Elsewhere \cite{weinberger2006introduction}, it is argued that one should maximize $tr(K)$, in the spirit of the popular Maximum Variance Unfolding approach \cite{weinberger2006introduction}.  Neither minimizing nor maximizing the trace actually imposes an exact rank constraint, which is non-convex and NP-hard.  One approach that could achieve exact rank constraints would be to use the Majorized Penalty Approach of Gao and Sun \cite{gao2010majorized} with an alternating minimization method.

A group of methods have studied the graph realization problem, where one is asked to recover the configuration of a cloud of points given a sparse and noisy set of pairwise distances between the points \cite{BiswasYe,biswas_stress_sdp,biswas, sdpAngle, univRig}.  
One of the proposed approaches involves minimizing the following energy 
\begin{equation}
 \min_{p_1,...,p_n\in \mathbb{R}^2 } \sum_{(i,j)\in E}  \left( \| p_i - p_j \|^2 - d_{ij}^2 \right)^2.
\label{E1}
\end{equation}
which unfortunately is nonconvex, but admits a convex relaxation into a SDP program. 
We refer the reader to Section 2 of \cite{asap2d} for several variations of this approach, some of which have been shown to be more robust to noise in the measured distances.

\vspace{-4mm}
\subsection{Local Ordinal Embedding}
\vspace{-2mm}

Terada and von Luxburg \cite{terada2014local} have recently proposed an algorithm for ordinal embedding and kNN embedding specifically, called Local Ordinal Embedding (LOE), which minimizes a soft objective function that penalizes violated ordinal constraints.
\vspace{-2mm}
\begin{equation}
\min_{X\in \R^{d\times n}} \hspace{-4mm}  {\sum_{i<j,k<l, (i,j,k,l)\in \mathcal{C}}  \hspace{-4mm} \max\left[0,D_{ij}(X)+\delta-D_{kl}(X)\right]^2 }.
\end{equation}
The energy takes into account not only the number of constraints violated, but
the distance by which the constraints are violated, penalizing large violations
more heavily.

An advantage of this energy in contrast to ones that normalize by
the variance of $X$ (to guarantee nondegeneracy) is its relatively simple
dependence on $X$, making the above energy easier to minimize. Instead, the
scale parameter $\delta$ guarantees nondegeneracy, and fixes the scale of the
embedding (which is indeterminable from ordinal data alone).

The authors introduce algorithms to minimizing the above energy, based on
majorization minimization and the Broyden-Fletcher-Goldfarb-Shanno (BFGS)
approximation of Newton's method, and prove that ordinal embedding is possible
when only local information is given (e.g. a $k$ neareast neighbor graph).  The
algorithm recovers not only the ordinal constraints, but the density structure
of the data as well. The algorithm applies to ordinal constraints associated
with $k$NN graphs as well more general sets of ordinal constraints. We will use
this crucial property when solving subproblems in the method presented here, as
the corresponding subgraphs are generally not $k$NN graphs.

\vspace{-5mm}
\section{ASAP \& Scale Synchronization  for Ordinal Embeddings}
\label{sec:ASAPordinal}
\vspace{-3mm}

In this section we detail the steps of the ASAP algorithm, central to the divide-and-conquer algorithm we propose for the ordinal embedding problem. Note that the difference between the original ASAP algorithm introduced in \cite{asap2d} and our approach lies in the 
decomposition method from Section \ref{sec:breakintopatches} and the 
scale synchronization step from Section \ref{sec:scalingSync}. The ASAP approach starts by decomposing the given graph $G$ into overlapping subgraphs (referred to as \textit{patches}), which are then embedded via the method of choice (in our case LOE). To every local patch embedding, there corresponds a scaling and an element of the Euclidean group Euc(d) of $d$-dimensional rigid transformations, and our goal is to estimate the scalings and  group elements that will properly align all the patches in a globally consistent framework. The local optimal alignments between pairs of overlapping patches 
yield noisy measurements for the ratios of the above unknown group elements. Finding group elements from noisy measurements of their ratios is also known as the group  synchronization problem.
for which Singer \cite{sync} introduced spectral and semidefinite programming (SDP) relaxations 
over the group SO(2) of planar rotations, 
which is a building block for the ASAP 
algorithm \cite{asap2d}.

Table~\ref{overview} gives an overview of the steps of our approach. 
The inputs are an ordinal graph (we consider kNN graphs) 
 $G=(V,E)$, where edge $ij\in E$ and non-edge $il \not\in E$ indicates that $d_{ij} \le d_{il},$  the max patch size parameter MPS, the target dimension $d$, and a base-case ordinal embedding method $OrdEmbed : G \mapsto X\in \R^{d\times n} $ for embedding each patch, such as LOE.

\begin{algorithm*}
  \begin{center}
	\small
    \begin{tabular}{| p{0.3\columnwidth} | p{1.6\columnwidth} |}
      \hline
      INPUT & $G=(V,E), \; |V|=n, \; |E|=m, \;$ $MPS, \; d, \; OrdEmbed(\cdot)$\\
      \hline
      \hline
      Choose Patches &
      1. Break $G$ into $N$ overlapping globally rigid patches $P_1,\ldots,P_N$ following the steps in Sec.~\ref{sec:breakintopatches}.\\
      Embed Patches &
      2. Embed each patch $P_i$ separately based on the ordinal constraints of the corresponding subgraph of $G$ using $OrdEmbed(\cdot).$
      \\
      \hline
      Step 1 & 1. If a pair of patches $(P_i, P_j)$ have enough nodes in common, let $\Lambda_{ij}$ be the median of the ratios of distances realized in the embedding of $P_i$ and their corresponding distances in $P_j$ as in (\ref{def:scalingMtx}); otherwise set $\Lambda_{ij}=0$.\\ 
      Scale 
       & 2. Compute the eigenvector $v_1^{\Lambda}$ corresponding to the largest eigenvalue of the sparse matrix $\Lambda$.  \\
       & 3. Scale each patch $P_i$ by $v_1^{\Lambda}(i)$, for $i = 1, \ldots, n$ \\
      \hline
      Step 2 & 1. Align all pairs of patches $(P_i, P_j)$ that have enough nodes in common. \\
      Rotate \& Reflect & 
      2. Estimate their relative rotation and possibly reflection $H_{ij} \in O(d) \subset \R^{d\times d}$.\\
      & 3. Build a sparse $dN\times dN$ symmetric matrix $H=(H_{ij})$ where entry $ij$ is itself a matrix in $O(d)$.\\
      & 4. Define $\mathcal{H} = D^{-1} H$, where $D$ is a diagonal matrix with \newline
      $D_{1+d(i-1),1+d(i-1)}=\ldots=D_{di,di} = deg(i)$, $i = 1,\ldots,N$, where $deg(i)$ is the node degree of patch $P_i$. \\
      & 5. Compute the top $d$ eigenvectors $v_i^\mathcal{H}$ of $\mathcal{H}$  satisfying  $ \mathcal{H} v_i^\mathcal{H}= \lambda_i^\mathcal{H} v_i^\mathcal{H}, i=1,\ldots,d$. \\
      & 6. Estimate the global reflection and rotation of patch $P_i$ by the orthogonal matrix $\hat{h}_i $ that is closest to $\widetilde{H}_i $ in Frobenius norm, where $\widetilde{H}_i $ is the submatrix corresponding to the i$^\text{th}$ patch in the $dN \times d$ matrix formed by the top $d$ eigenvectors $[v_1^\mathcal{H} \ldots v_d^\mathcal{H}]$. \\
      & 7. Update the embedding of patch $P_i$ by applying the above orthogonal transformation $\hat{h}_i $ \\ 
      \hline
      Step 3 Translate &  Solve $m\times n$ overdetermined system of linear equations (\ref{eq:translation}) for optimal translation in each dimension.\\
      \hline
      \hline
      OUTPUT & Estimated coordinates $\hat{x}_1,\ldots,\hat{x}_n$ \\
      \hline
    \end{tabular}
  \end{center}
  \caption{Modified ASAP algorithm that incorporates the scale synchronization step.}
  \label{overview}
  \end{algorithm*}

\normalsize

\vspace{-5mm}
\subsection{Break up the kNN graph into patches and embed}
\label{sec:breakintopatches}
\vspace{-2mm}

The first step we use in breaking the kNN graph into patches is to apply normalized spectral clustering \cite{von2007tutorial} to a symmetrized version of the graph.  Normalized spectral clustering partitions  the nodes of a graph into $N\ll n$ clusters by performing k-means 
on the embedding given by the top 
$N$ eigenvectors of the random-walk normalized graph Laplacian.  It is shown \cite{von2007tutorial} that normalized spectral clustering minimizes a relaxation of the normalized graph cut problem.  Next, we enlarge the clusters by adding the graph-neighbors of each node, so that the resulting patches have significant overlap, a prerequisite for the ASAP synchronization algorithm. The higher the overlap between the patches, the more robust the pairwise group ratio estimates would be, thus leading overall to a more accurate final global solution. 
Finally, we use an iterative procedure to remove nodes from the graph relying on tools from rigidity theory. 
\footnote{A graph is globally rigid if all realizations of it are congruent up to a rigid transformation.}
If a patch is not globally rigid, we drop a constant fraction of the added nodes. At each round we choose to drop a quarter of the nodes with the lowest degree while retaining all nodes that were in the original cluster generated by k-means in the corresponding patch.  This uses the heuristic that low-degree nodes tend to render a graph not globally rigid.  After dropping nodes, we check the remaining patch for globally rigidity again.  We stop the pruning process when the patch contains fewer than $4/3$ the number of nodes in the original cluster, or the patch is globally rigid.

  We refer the readers to Appendix~\ref{sec:rigidity} for for a brief description of global rigidity, and relevant results in the literature, and use the remainder of this section as a brief discussion of the main definitions. 
In 
the \textit{graph realization problem} (GRP), one is given a graph $G=(V,E)$ together with a non-negative distance measurement $d_{ij}$ associated with each edge, and is asked to compute a realization of $G$ in $\mathbb{R}^d$. In other words, for any pair of adjacent nodes $i$ and $j$, the distance $d_{ij} = d_{ji}$ is available, and the goal is to find a $d$-dimensional embedding $p_1, p_2, \ldots, p_n \in \mathbb{R}^d$ such that $\|p_i-p_j \| =d_{ij}, \mbox{ for all } (i,j) \in E$. The main difference between the GRP and the problem we aim to address in  our paper is the input information available to the user. Unlike the GRP problem where distances are available to the user, here we only have information of the adjacency matrix of the graph and have the knowledge that it represents a kNN graph. Both problems aim to recover an embedding of the initial configuration of points.

A graph is globally rigid in $\mathbb{R}^d$ if there is a unique (up to the trivial Euclidean isometries) embedding of the graph $\mathbb{R}^d$ such that all distance constraints are preserved. It is well known that a necessary condition for global rigidity is 3-connectivity of the graph. Since the problem at hand that we are trying to solve is harder (as we do not have distance information available) we require that the patches we generate are globally rigid graphs. Even in the favorable scenario when we do have available distance measurements (which we do not in the present problem, but only ordinal information), any algorithm seeking an embedding of the graph would fail if the graph were to have multiple non-congruent realizations.

\vspace{-4mm}
\subsection{Scale Synchronization}
\label{sec:scalingSync}
\vspace{-2mm}
Before applying the original ASAP algorithm to the embedded patches, we introduce an additional step that further improves our approach and is independent of the dimension $d$. In the \textit{graph realization problem} which motivated ASAP, one is given a graph $G=(V,E)$ and non-negative distance measurement $d_{ij}$ associated with each edge $ij\in E(G)$,
and is asked to compute a realization of $G$ in $\mathbb{R}^d$. The distances are readily available to the user and thus the local embedding of each patch is on the same scale as the ground truth. 
However, in the kNN embedding problem, distances are  unknown and the scale of one patch relative to another  must be approximated. Any ordinal embedding approach has no way of relating the scaling of the local patch to the global scale. To this end, we augment the ASAP algorithm with a step where we synchronize local scaling information to recover an estimate for the global scaling of each patch, thus overall synchronizing over the group of similarity transformations.

We accomplish this as follows.  Given a set of patches, $\{P_i\}_{i=1}^N$, we create a patch graph in which two patches are connected if and only if they have at least $d+1$ nodes in common. 
We then construct a matrix $\Lambda\in\R^{N\times N}$ as
\vspace{-2mm}
\begin{equation}
\Lambda_{ij} = \begin{cases}
\mbox{median}\left\{ \frac{D_{a,b}^{P_i}}{D_{a,b}^{P_j}}\right\}_{a\ne b\in P_i \cap P_j} & \vspace{-3mm} \mbox{if $P_i \sim P_j$, $i\le j,$}\\
1/\Lambda_{ji} & \mbox{if $P_i \sim P_j$, $i> j,$}\\
0 &\mbox{otherwise},
\end{cases}
\label{def:scalingMtx}
\end{equation}
where $D_{a,b}^{P_i}$ is the distance between nodes $a$ and $b$ as realized in the embedded patch $P_i$.
The matrix $\Lambda$ approximates the relative scales between patches.  If all distances in all patches were recovered correctly up to scale, and all patches had sufficient overlap with each other, then each row of $\Lambda$ would be a scalar multiple of the others and each column of $\Lambda$ would be scalar multiple of the others, thus rendering $\Lambda$  a rank-1 matrix. For the noisy case, in order to get a consistent estimate of global scaling, we compute the best rank-1 approximation of $\Lambda$,  
given by its leading eigenvector $v_1^{(\Lambda)}$.  We use this approximation of global scaling to rescale the embedded patches before running ASAP. Note that the 
connectivity of the patch graph and the non-negativity of $\Lambda$ guarantee, via the Perron-Frobenius Theorem, that all entries of $v_1^{(\Lambda)}$ are positive. We refer the reader to Figure \ref{fig:scalecomparison}, which  illustrates on an actual example the importance of this scaling  synchronization step.

\vspace{-4mm}
\subsection{Optimal Rotation, Reflection and Translation}
\begin{figure}[t]
\center
\includegraphics[width=0.7\columnwidth]{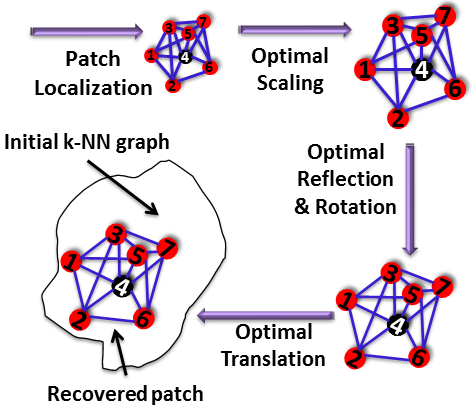}
\vspace{-5mm}
\caption{ASAP and scale synchronization pipeline.}
\vspace{-5mm}
\label{fig:diagram}
\end{figure}

\vspace{-2mm}
After applying the optimal scaling to each patch embedding, we use the original ASAP algorithm 
to integrate all patches in a global framework, as illustrated in the pipeline in Figure~\ref{fig:diagram}. We estimate, for each patch $P_i$, an element of the Euclidean group Euc($d$) $=$ O(d) $ \times \mathbb{R}^d $  which, when applied to that patch embedding $P_i$, aligns all patches as best as possible in a single coordinate system. 
In doing so, we start by estimating, for each pair of overlapping patches $P_i$ and $P_j$, their optimal relative rotation and reflection, i.e., an element $H_{ij}$ of the orthogonal group O(d) that best aligns $P_j$ with $P_i$. Whenever the patch embeddings perfectly match the ground truth, $H_{ij} = O_i O_j^{-1}$. We refer the reader to \cite{asap2d} for 
several methods on aligning pairs of patches and computing their relative reflections and rotations  $H_{i,j}$. Finding group elements $\{O_i\}_{i=1}^N$ from noisy measurements $H_{ij}$ of their ratios is also known as the group synchronization problem. Since this problem is NP-hard,  
we rely on the spectral relaxation \cite{sync} of 
\vspace{-3mm}
\begin{align}\label{eq:O2sync}
\min_{O_1,\dots,O_N\in O(d)} \sum_{P_i\sim P_j}\| O_{i} O_j^{-1} - H_{ij}\|_F^2.
\end{align}
for synchronization over O(2), and estimate a consistent global rotation 
of each patch 
from the top $d$ eigenvectors of the 
graph Connection Laplacian, as in Step 2.4 in Table \ref{overview}. 
We estimate the optimal translation of each patch by solving, in a least squares sense, $d$ overdetermined linear systems 
\vspace{-2mm}
\begin{equation}
     x_i - x_j = x^{(k)}_i - x^{(k)}_j,\quad (i,j) \in E_k,\quad k=1,\ldots,N, 
\label{eq:translation}
\end{equation}
where $x_i$, respectively $x^{(k)}_i$, denote the unknown location of node $i$ we are solving for, respectively, the known  location of node $i$ in the embedding of patch $P_k$. 
We refer the reader to \cite{asap2d} for a  description of computing the optimal translations.

\vspace{-4mm}
\subsection{Extension to higher dimensions}
\vspace{-2mm}
Although we present experiments here on 2D and 3D data, the ASAP  approach extends naturally to higher dimensions. In the 3D case, ASAP has been recently used as a scalable robust approach to the \mbox{molecule problem} in structural biology \cite{asap3d}. For the $d$-dimensional general case, one can extend ASAP by first
using the same approach for scaling synchronization from Section \ref{sec:scalingSync}, 
then synchronizing over O($d$),  
and finally estimating the optimal translations over $R^d$ by solving
 $d$ overdetermined systems of linear equations via least-squares.
The LOE approach that can be used to obtain the local patch embeddings required by ASAP, has a natural extension to the $d$-dimensional case, thus rendering the entire pipeline amenable to dealing with higher-dimensional data.

\section{Density Estimation}   \label{sec:densityEst}

In this section, we remark on the explicit connection between the graph embedding problem considered in this paper and the density estimation problem. 
In particular, one may approach the problem of recovering the unknown coordinates underlying the kNN graph by first aiming to estimate the density function that generates the coordinates.  Suppose for example that one is able to estimate the pointwise density $u:\Omega\subseteq\R^d \rightarrow[0,1]$, up to some constant multiple, evaluated at each vertex of the graph, $x_i$.  Next, as outlined in \cite{von2013density}, one can assign weights to the originally unweighted kNN graph, defined by $w(x_i,x_j)=(u^{-1/d}(x_i)+u^{-1/d}(x_j))/2$. Furthermore, it can be shown that the shortest path distance in the resulting weighted kNN graphs converges to the  Euclidean distance of the original points as the number of points increases.  In other words, applying multidimensional scaling to the shortest path distances on the weighted kNN graph will yield increasingly accurate embeddings of the original points $\{x_i\}_{i=1}^n$ as $n\rightarrow +\infty$.

In contrast to finding an approximate embedding from a density estimate, under certain conditions, the reverse process is also straightforward.  With sufficiently many points and sufficiently strong priors on the distribution, the methodology of Maximum Penalized Likelihood Estimation (MPLE) applies \cite{EggLaR01}.  One first assumes that the locations correspond to points drawn independently identically distributed according to some unknown underlying spatial distribution.  MPLE approximates the most likely spatial distribution given the points observed and some assumed prior distribution on the space of distributions.  The data fidelity term comes in the form of a log-likelihood term, a function of the distribution estimate and the point locations, and is given by
\bas{
L(u,\{x_i\}_{i=1}^n)
=
\sum_{i=1}^n \log(u(x_i)),
}
and the penalty term, $P(u)$ enforces the prior distribution on the space of distributions.  
Typical choices for $P(u)$ include the $H^1$-seminorm regularizer, $P(u)=\frac{\lambda}{2}\int_\Omega |\nabla u|^2dx$, enforcing smoothness, and Total Variation (TV) norm regularization, $P(u)=\lambda\int_\Omega |\nabla u|dx$, which enforces smoothness, but also allows for edges.
Therefore, general MPLE seeks to optimize the following energy over all probability distributions on the spatial domain $\Omega \subseteq \R^d$
\bas{
\hat u 
=
\argmax_{u\ge 0, \int_\Omega u dx =1}{L(u,\{x_i\}_{i=1}^n)-P(u) }.
}
The form and scale of $P$ encodes different types and amounts of regularity in the resulting density estimate $u$.  In practical settings, cross-validation should be performed to determine the appropriate amount of regularity to impose on a given data set.

For the purpose of using kNN graphs to recover densities, we will include a post-processing step for a subset of the embedding experiments, to which we apply a standard implementation of TV MPLE \cite{mohler2011fast} to the embedded points.  TV is a good choice of penalty because we will be applying it to points that are drawn from a piecewise constant density.  The good density estimates based on good embeddings shown in Section~\ref{sec:experiments} illustrate that there is in fact a strong connection between the embedding and density estimation problems.  

The actual implementation of the TV MPLE relies on the Split Bregman (equivalently Alternating Direction Method of Multipliers)  minimization technique in which one introduces a splitting and equality constraints that are enforced by performing saddle-point optimization of the augmented Lagrangian.  This results in an iterative update procedure given by Algorithm~\ref{alg:tvmple}.
\begin{algorithm}
  \caption{TV MPLE}
  \label{alg:tvmple}

INPUT : $\{x_i\}, \rho, \gamma$\\
$y = 0, z = 0$\\
For $numberIterations$ 
$\{\\$
\bas{\left(\hat u,\hat d\right) = &\\
 \argmin_{u\ge 0,d}\bigg\{&\|d\|_1-\sum_{i=1}^n \log(u(x_i)) \\
  &+\frac{\rho}{2}\|\nabla u-d+y\|_2^2+\frac{\gamma}{2}\left(\|u\|_1-1+z\right)^2\bigg\}\\
y=&y+\tilde \nabla \hat u-\hat d\\
z=&z+\|\hat u\|_1-1
  }
$\}$

\end{algorithm}
The first minimization step is actually replaced by minimizing over $u$, and $d$ individually, making use of the shrinkage proximal operator associated with the $\ell^1$ norm.

\vspace{-4mm}
\section{Experiments}
\label{sec:experiments}
\vspace{-3mm}


Our experiments compare 
embeddings of points drawn from three different 2D synthetic densities: piecewise constant half-planes (\textbf{PC}), piecewise constant squares (\textbf{PCS}), and Gaussian (\textbf{Gauss}), and a 3D synthetic denisty : piecewise constant half-cubes (\textbf{halfcube}), each with $n=\{500,1000,5000\}$ points, as well as points drawn uniformly from a 3D donut shape (\textbf{Donut}) with  $n=500$, and the actual 2D coordinates of $n=1101$ cities in the US (\textbf{US cities}).  For a given set of data points, we use its kNN adjacency matrix as input to each ordinal embedding method.  Separate from these datasets with a clear correct geometric embeddings, we find embeddings of points in a co-authorship network of network scientists (\textbf{NetSci2010}) with $n=552$ (see Section~\ref{sec:netsci2010}).
We test Laplacian Eigenmaps \cite{Belkin_Niyogi}, the LOE BFGS and LOE MM methods  \cite{terada2014local}, and ASAP with LOE BFGS used for the patch embeddings.  As LOE was already compared with several methods in \cite{terada2014local}, attaining better performance than LOE may suggest better performance than a number of relevant methods including Kamada and Kawai \cite{kamada1989algorithm}, and Fruchterman and Reingold \cite{fruchterman1991graph}. We remark that our approach deals with a different input than that of the \textbf{t-SNE} method in \cite{van2012stochastic}, which is generally used for embeddings of high dimensional data where some of the constraints are deliberately violated, which is not necessarily the case in our setting. 
We evaluate the methods based on (wall-clock) runtime and two different error metrics, Procrustes alignment error\cite{sibson1978studies}, and 
\textit{A-error} ($\mathcal{E}_A$) defined as the percentage of edge disagreements between the kNN adjacency matrix of the proposed  embeddeding $\tilde X$ and the  ground truth 
\vspace{-2mm}
\begin{equation}
  \label{eq:Aerr}
\mbox{error}(\tilde X,X): \;\;  \mathcal{E}_A \;\; \mydef \;\; \frac{1}{n^2}\sum_{i,j=1}^n\left|\left(A^k_{\tilde X}\right)_{ij}-\left(A^k_{X}\right)_{ij}\right|,
\end{equation}
where $A_{X}^k\in \{0,1\}^{n\times n}$ denotes the adjacency matrix of the corresponding kNN graph.
We set varying limits on the number of LOE iterations $\{5,10,50,100,300,500\}$, and we use varying maximum patch sizes (MPS) for ASAP .
The LOE and ASAP methods give, for each distribution and values $n$ and $k$, an error-runtime Pareto curve (with low values in both coordinates being best).
In Table~\ref{table:notation}, we establish some shorthand notation for the
methods and parameters used in this section. 
 For fair comparisons, we pass the same randomly sampled data to each of the methods. 
Ideally, one would run these experiments many times over and average the results (to get an estimate of average performance), but this is effect already partially accomplished by running the LOE and ASAP methods with multiple parameters to get a more holistic measurement of performance.
It is worth mentioning that while LOE BFGS and LOE MM are iterative methods which should converge to the best estimate of the solution as the number of iterations increases, ASAP is not iterative and the results of ASAP LOE with a given MPS, do not inform the results of ASAP LOE with another MPS. This aspect, combined with the randomized k-means spectral clustering used to choose patches means that we do not generally expect the recovery errors of ASAP LOE to be monotonically decreasing with MPS or time (as higher MPS generally leads to longer computational time). 
A principled way of choosing the best MPS for a given application of ASAP LOE could be of further interest.
\begin{table}[h]
  \center
  \small
\begin{tabular}{|l | p{5cm}|}
  \hline
  Recovery Method&\\
  \hline
  LE & Laplacian Eigenmaps embedding\\
  LOE MM & Local Ordinal Embedding using majorization minimization\\
  LOE BFGS &  Local Ordinal Embedding using BFGS \\
  ASAP LOE & ASAP \& LOE BFGS patch embeddings  \\ 
  \hline
  Parameters & \\
  \hline
  sparse k & $k=\lceil 2\log(n)\rceil$\\
  dense k &  $k=\lceil \sqrt{n\log(n)}\rceil$\\
  MPS & maximum patch size (for ASAP)\\ 
  Iter. & number of iterations (of LOE)\\
\hline
  Data sets & \\
  \hline
  \textbf{PC} 		& 2D piecewise constant half-planes\\
  \textbf{PCS}   	& 2D piecewise constant squares  \\
  \textbf{Gauss} 	& 2D Gaussian  \\
  \textbf{halfcube} 	& 3D piecewise constant half-cubes \\
  \textbf{Donut} 	& 3D Donut  \\
  \textbf{US cities} & 2D coordinates of US cities  \\
  \textbf{NetSci2010} &  co-authorship network of scientists\\
\hline
\end{tabular}
\vspace{-2mm}
\caption{Notation for plotting experimental results.}
\vspace{-2mm}
\label{table:notation}
\end{table}

\subsection{The need for scale synchronization}
\begin{figure}[h]
\center
\begin{tabular}{c}
\includegraphics[height=\sixthtextheight]{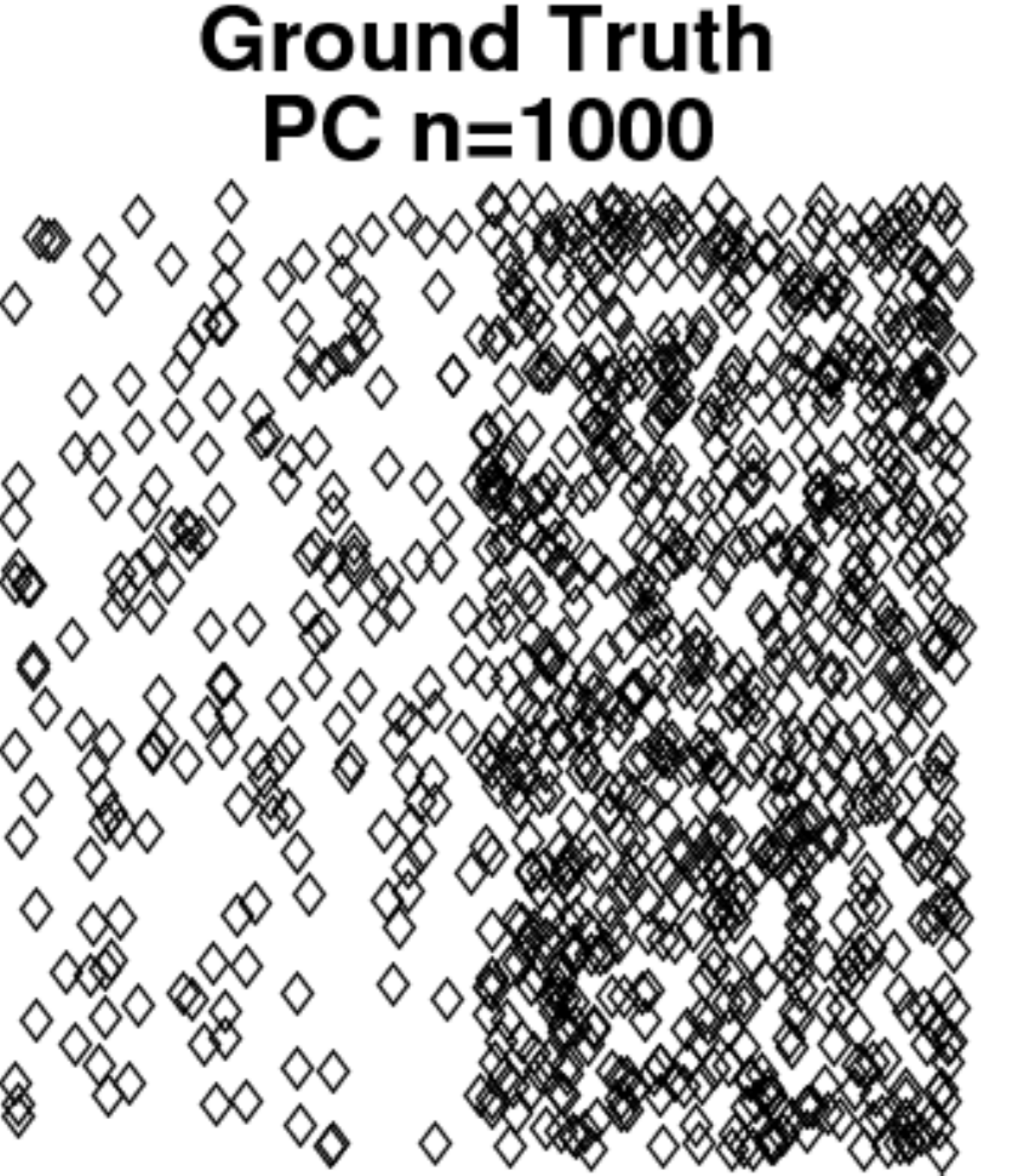}
\includegraphics[height=\sixthtextheight]{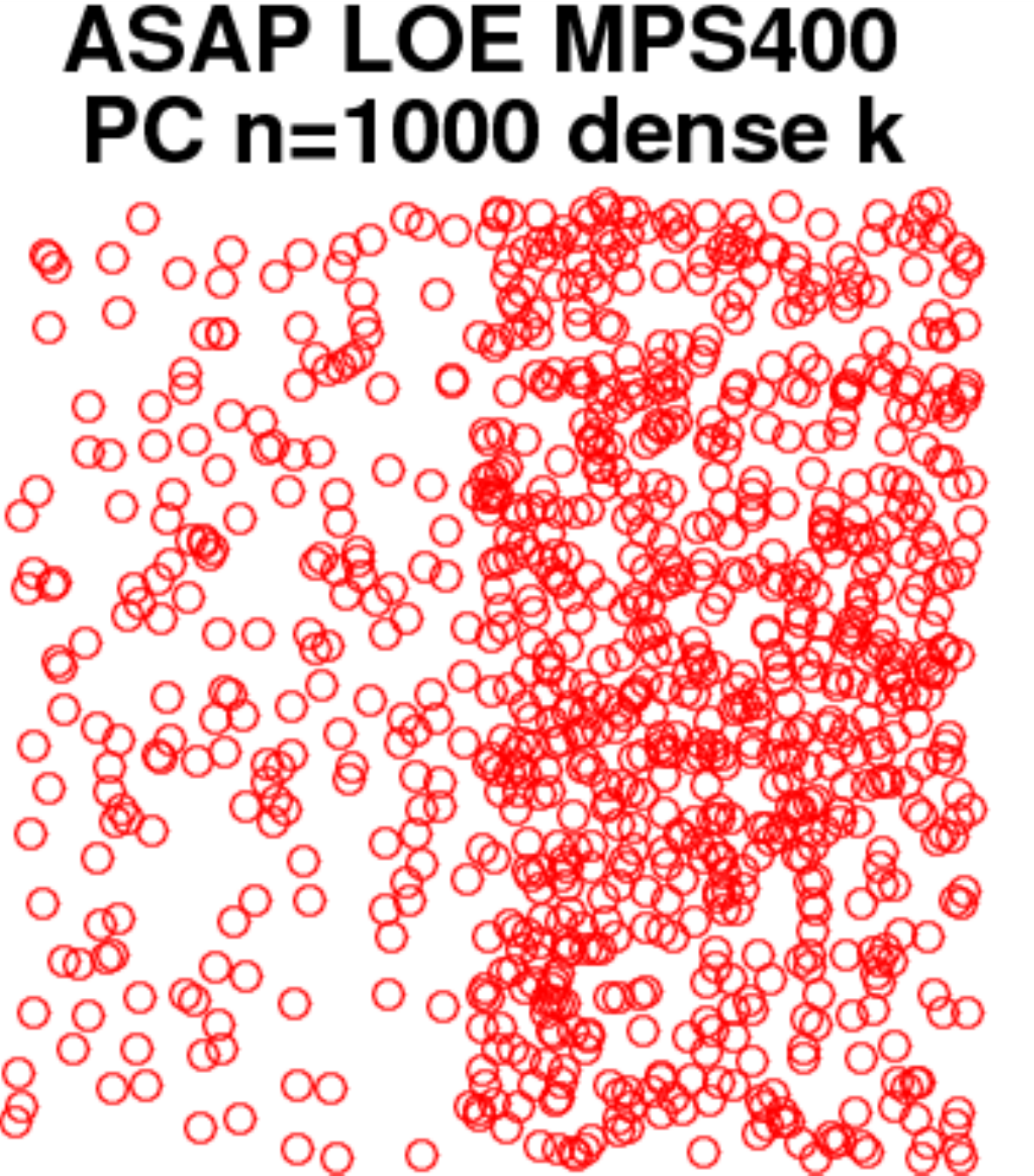}
\includegraphics[height=\sixthtextheight]{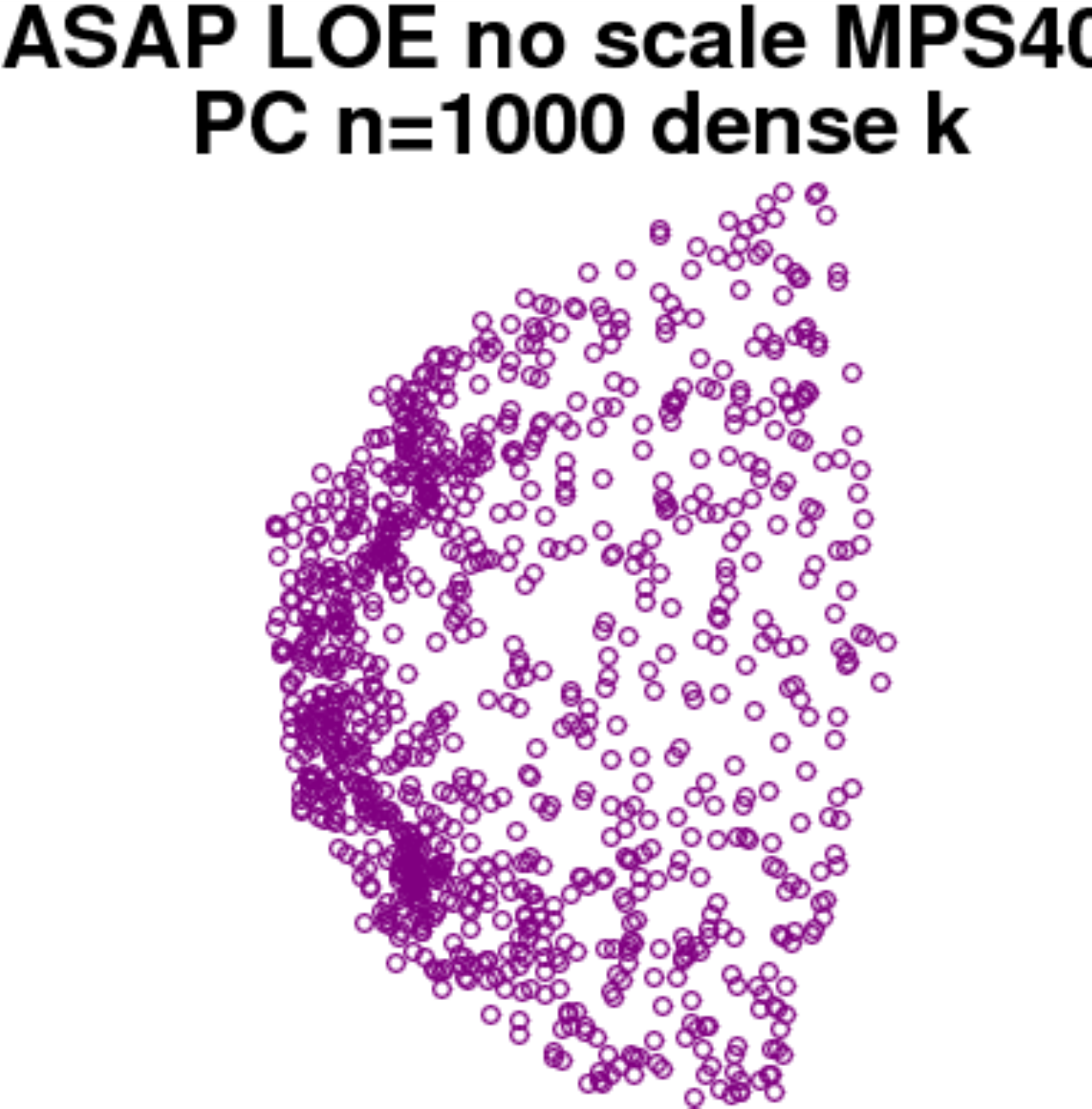}
\end{tabular}
\vspace{-2mm}
\caption{Left: Ground truth, $n=1000, k=14$. Middle: ASAP LOE 
with scale synchronization: 
 $\mathcal{E}_A = 0.007$. 
Right: ASAP LOE without scale synchronization: 
 $\mathcal{E}_A = 0.038$. 
}
\label{fig:scalecomparison}
\end{figure}

First, to illustrate the importance of the scale synchronization  introduced in Section \ref{sec:scalingSync}, we compare in Figure~\ref{fig:scalecomparison}  ASAP synchronized embeddings 
with and without this step. Clearly, this 
 step significantly improves the recovered solutions.

\subsection{Simulations with $n=500, 1000, 5000$ with sparse and dense $k$}
\begin{figure}[h]
\center
\begin{tabular}{c}
\includegraphics[width=\halfcolwidth]{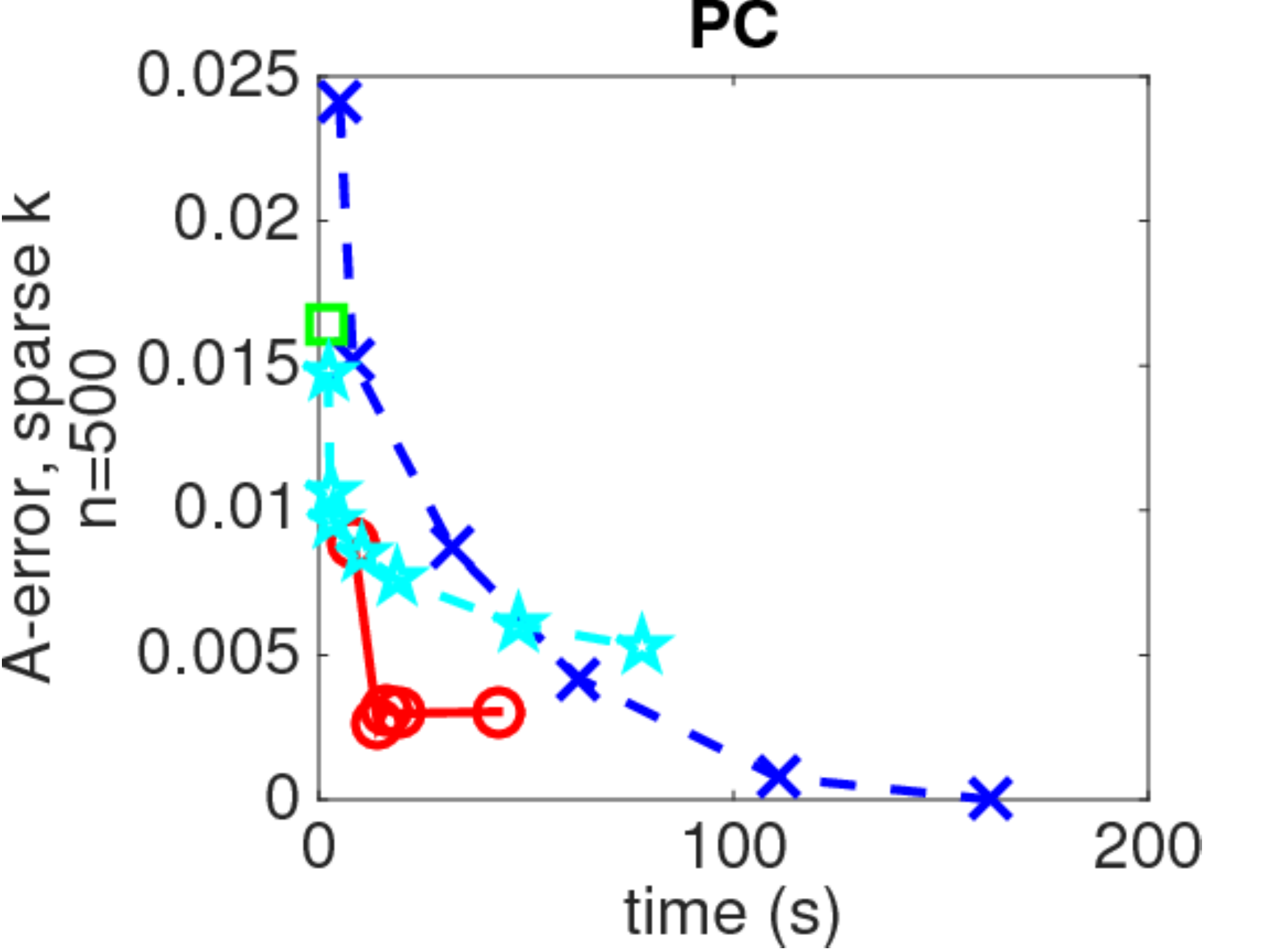}
\includegraphics[width=\halfcolwidth]{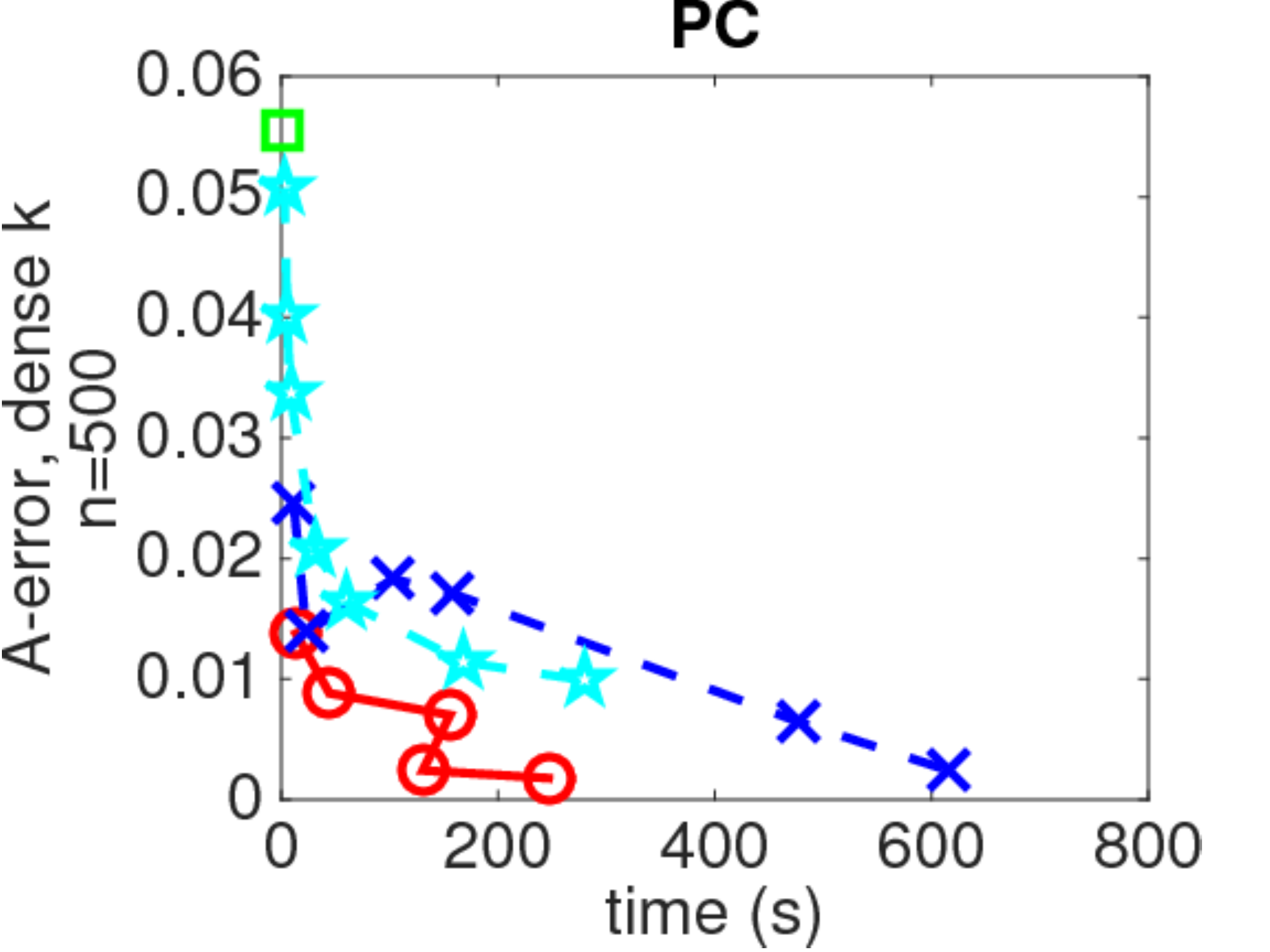}\\
\includegraphics[width=\halfcolwidth]{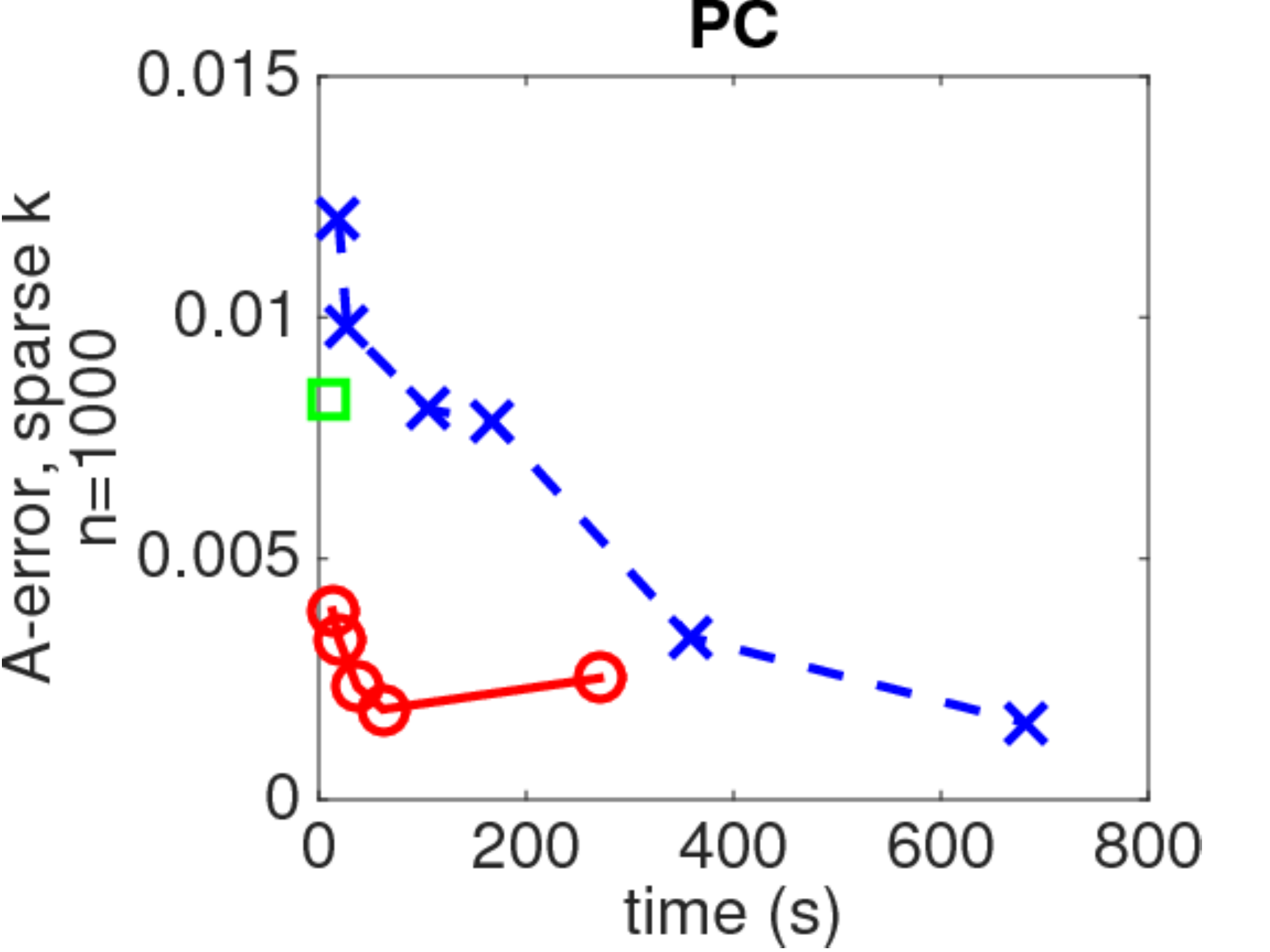}
\includegraphics[width=\halfcolwidth]{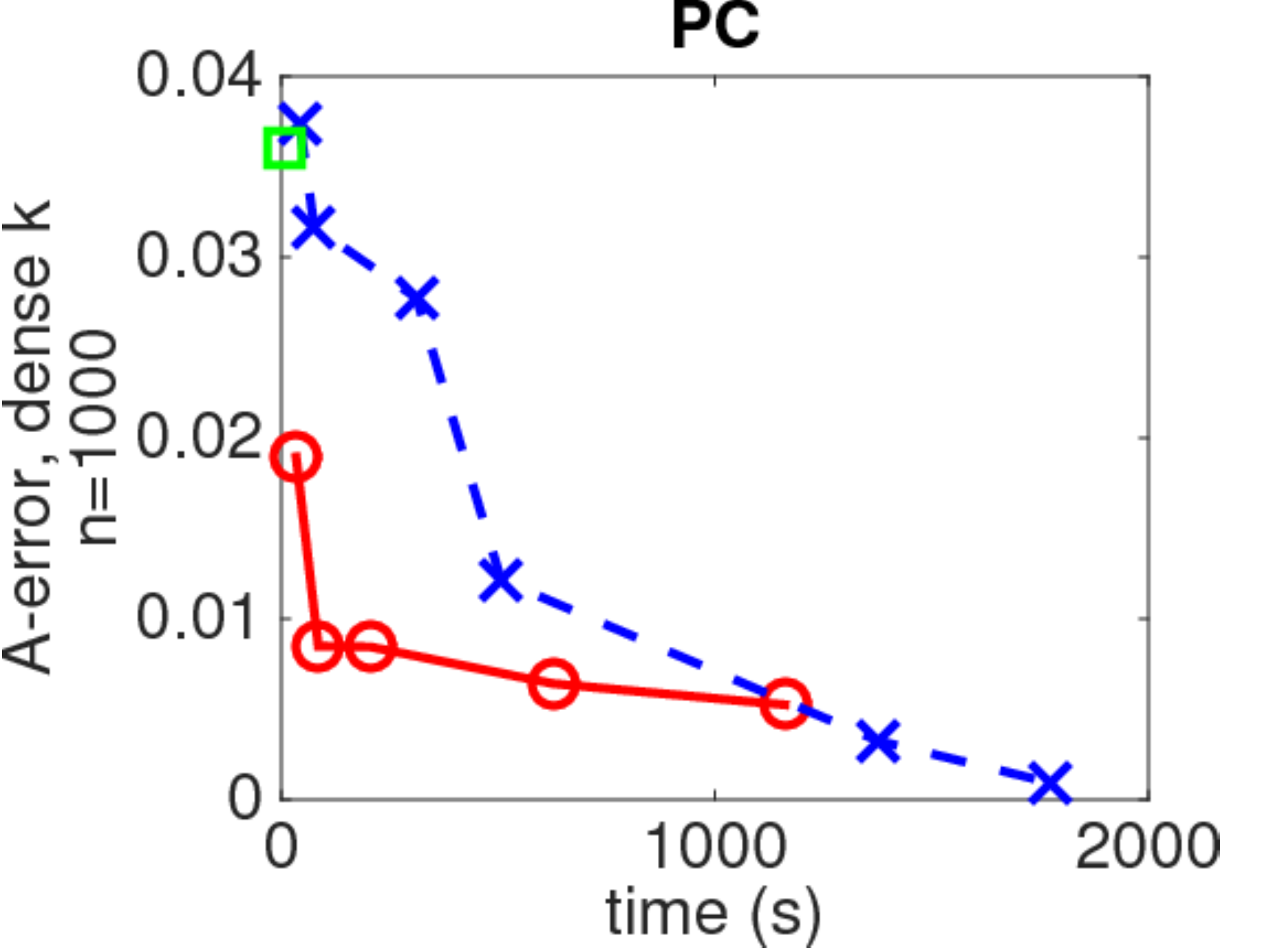}\\
\includegraphics[width=\halfcolwidth]{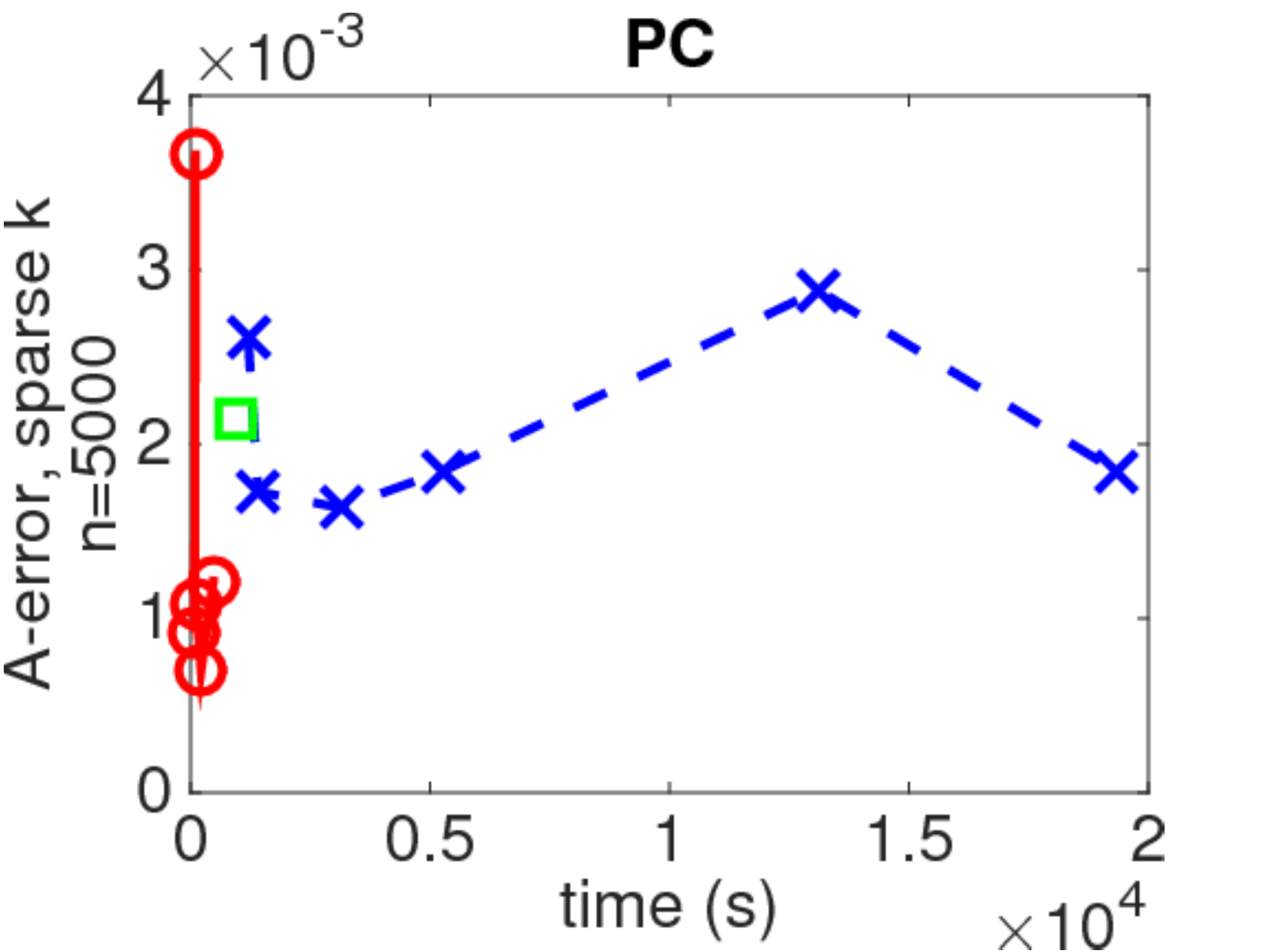}
\includegraphics[width=\halfcolwidth]{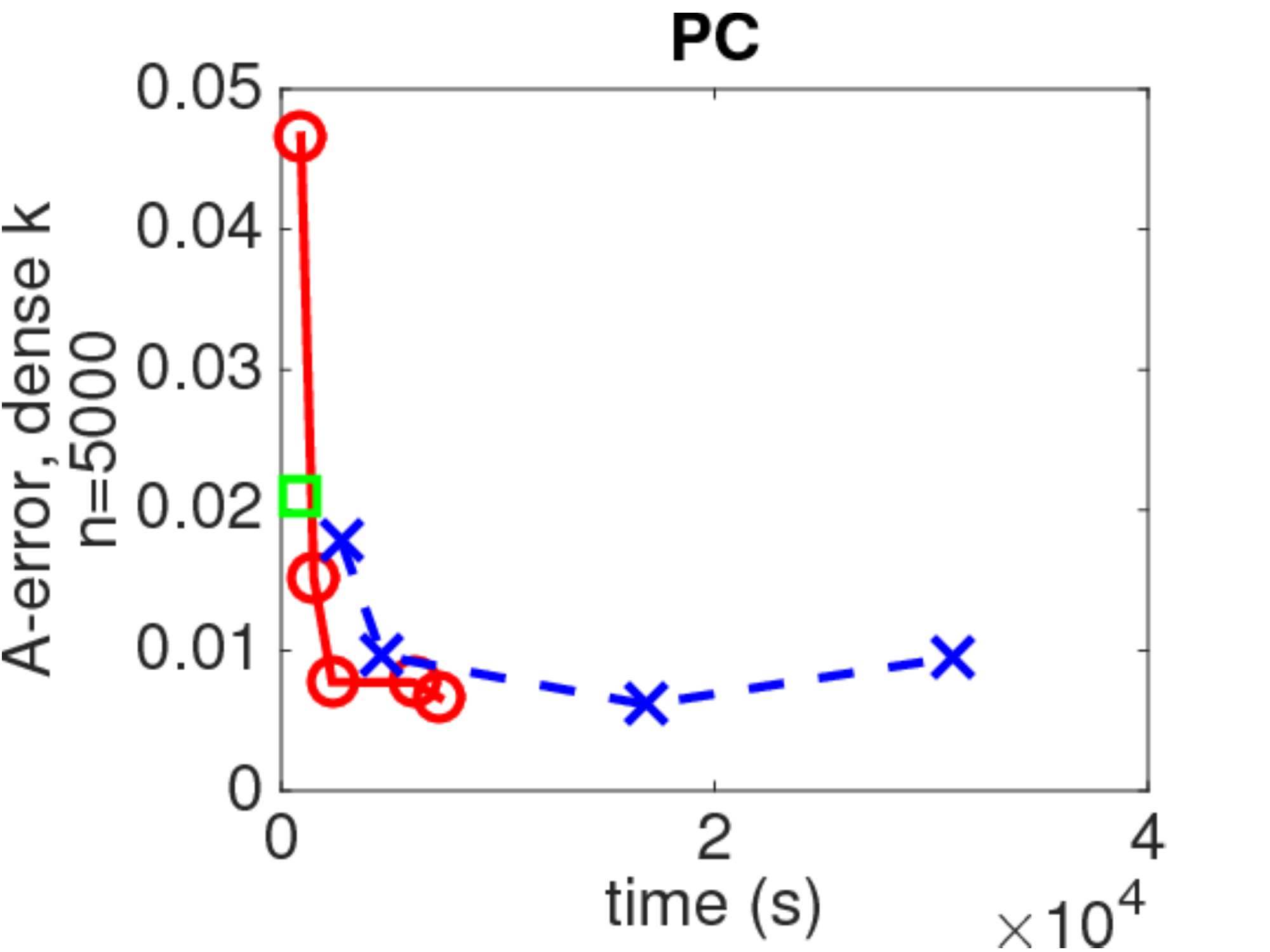}
\end{tabular}
\vspace{-4mm}
\caption{$\mathcal{E}_A$  vs. time, $n=\{500,1000,5000\}$, Left : $k$ sparse, Right : $k$ dense, piecewise constant half-planes, 
\colorlegend}
\vspace{-3mm}
\label{fig:AerrPC}
\end{figure}

\begin{figure}[h]
\center
\begin{tabular}{c}
\includegraphics[width=\halfcolwidth]{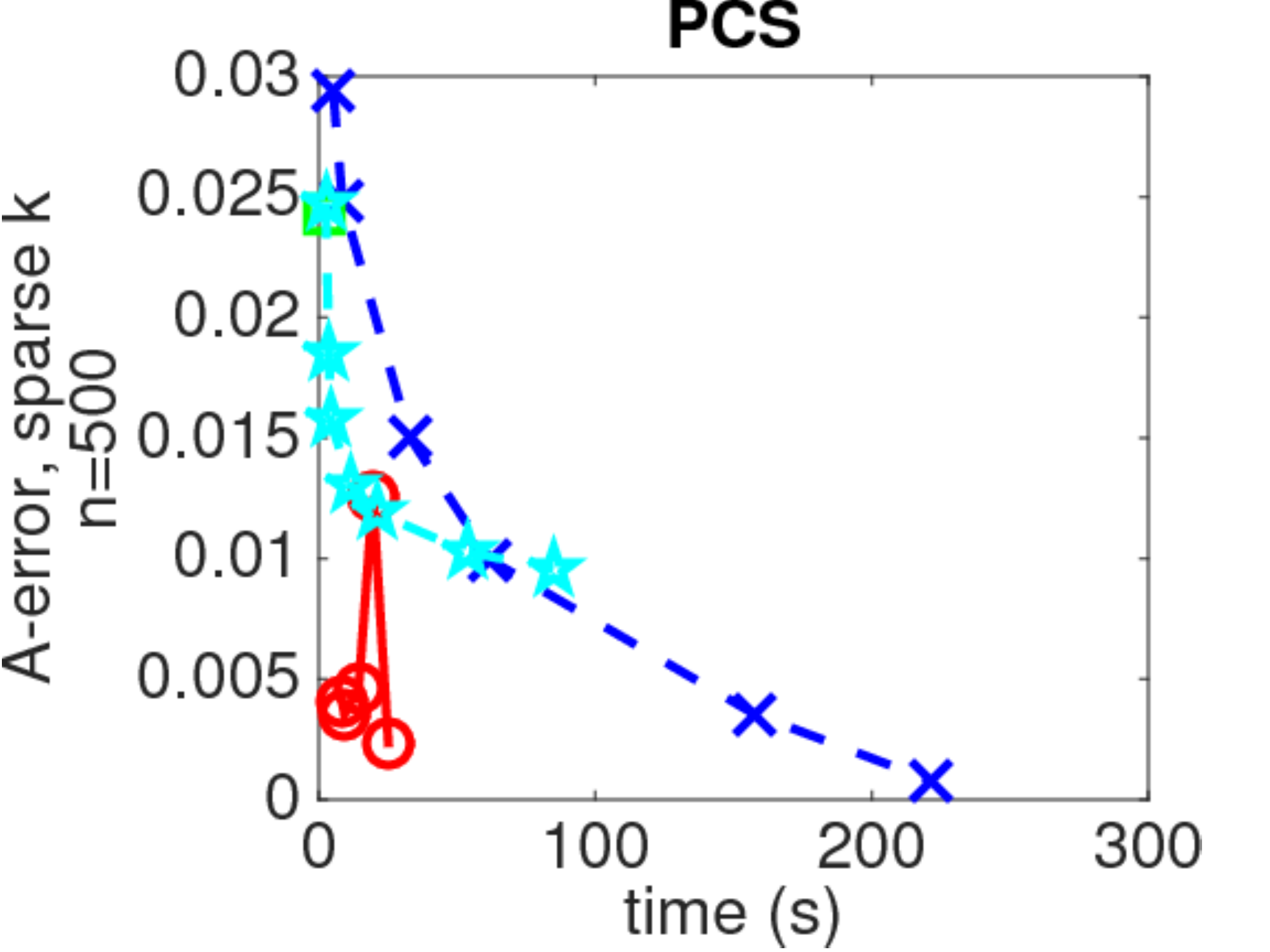}
\includegraphics[width=\halfcolwidth]{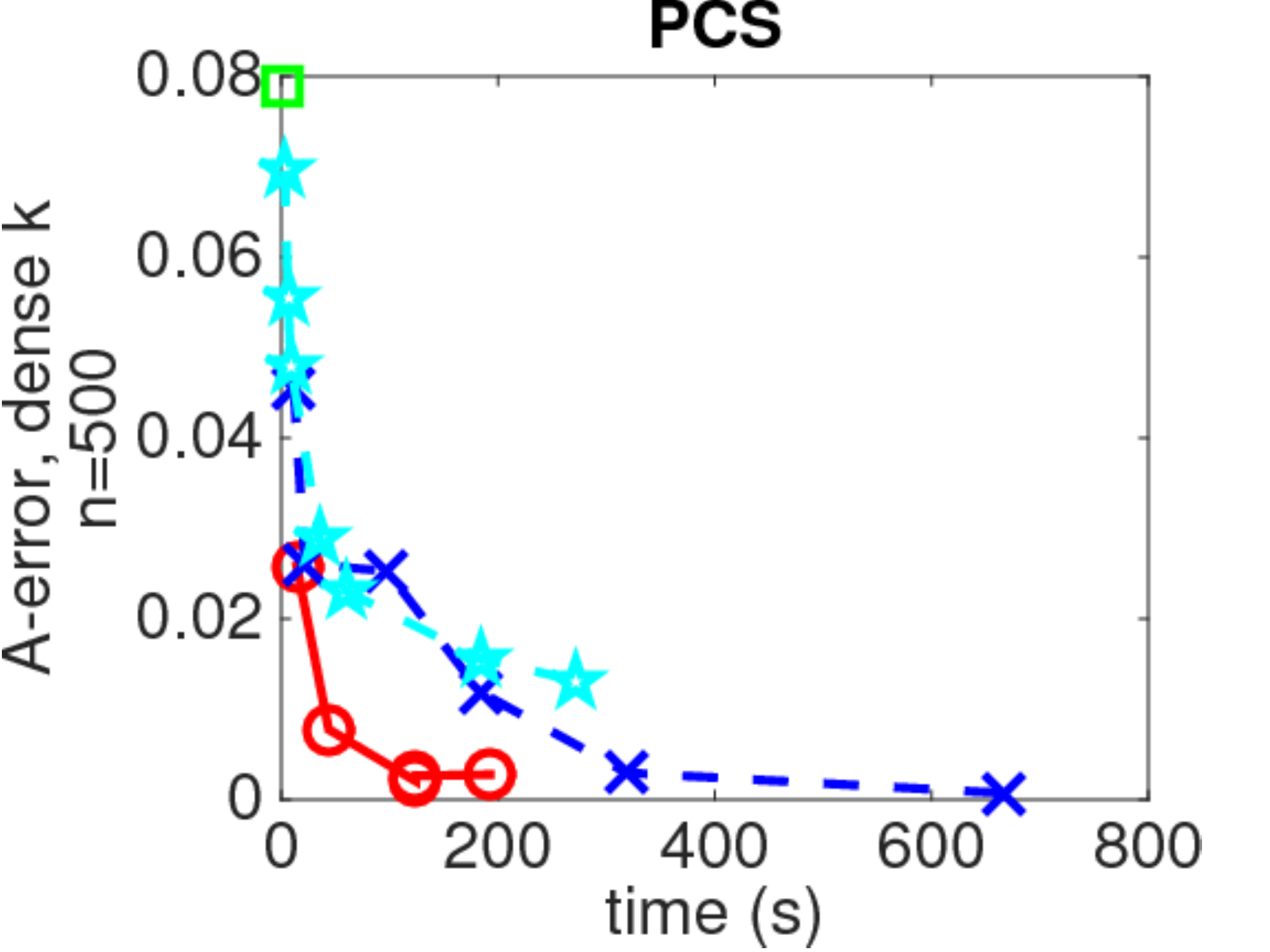}\\
\includegraphics[width=\halfcolwidth]{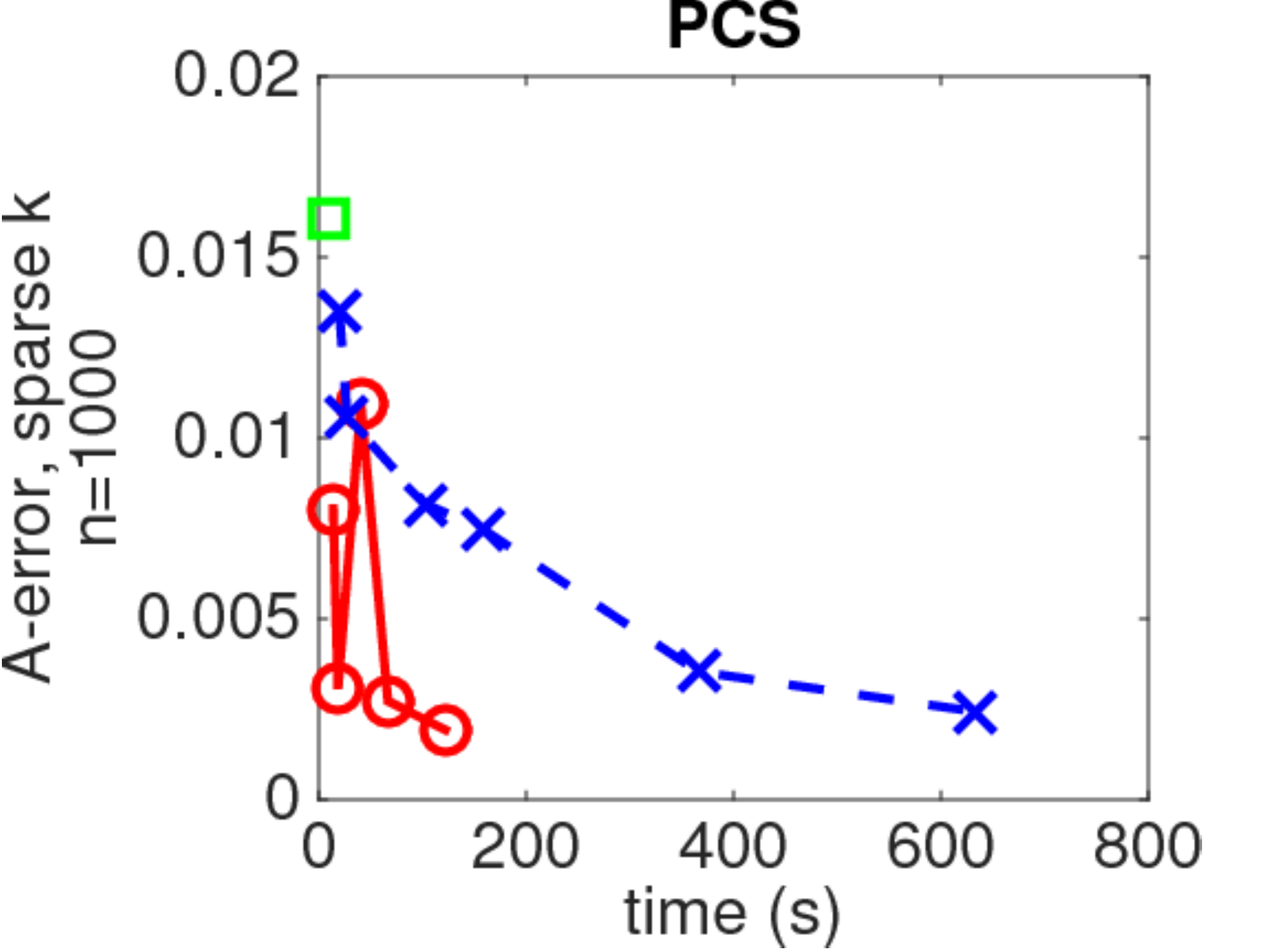}
\includegraphics[width=\halfcolwidth]{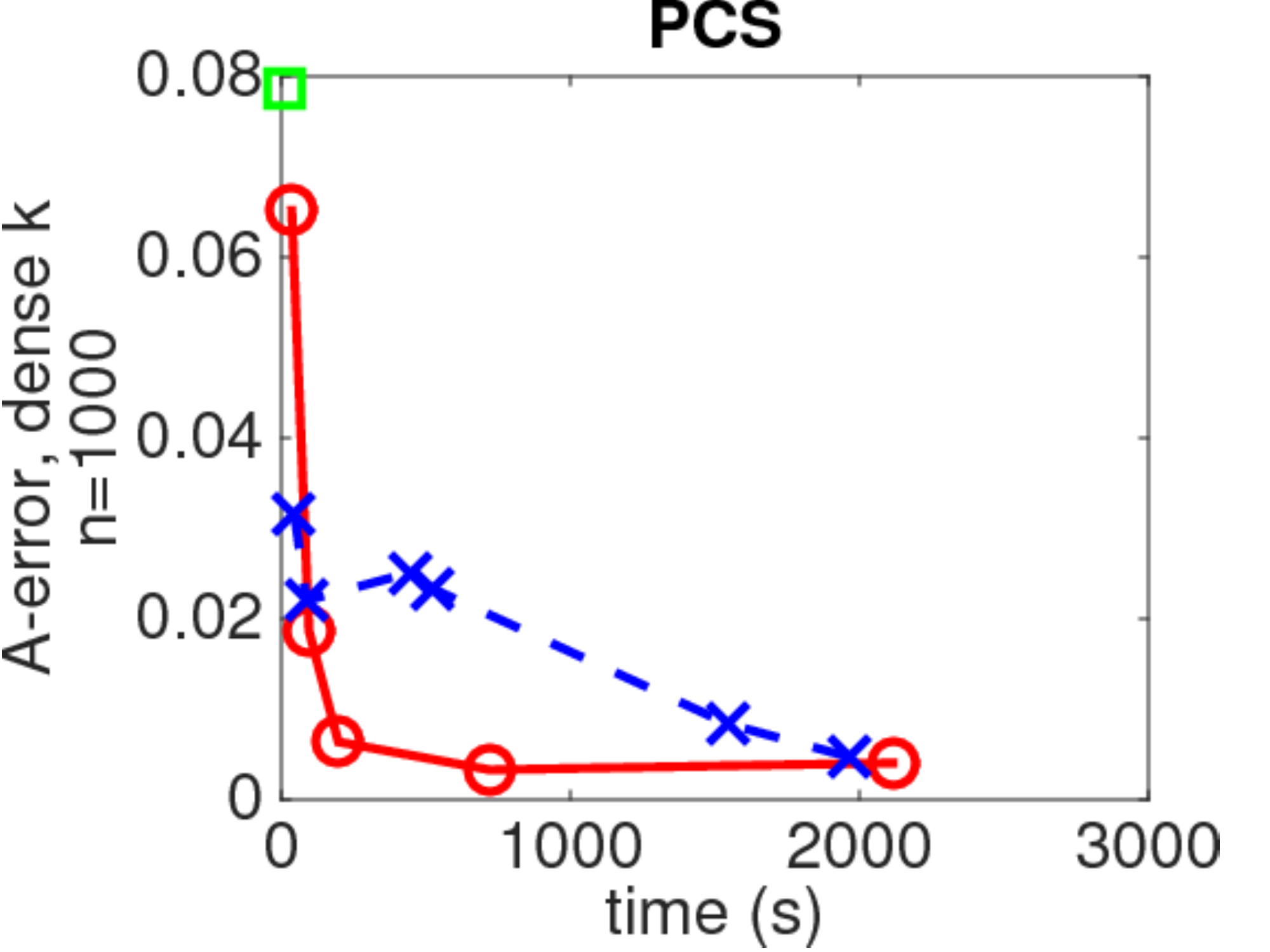}\\
\includegraphics[width=\halfcolwidth]{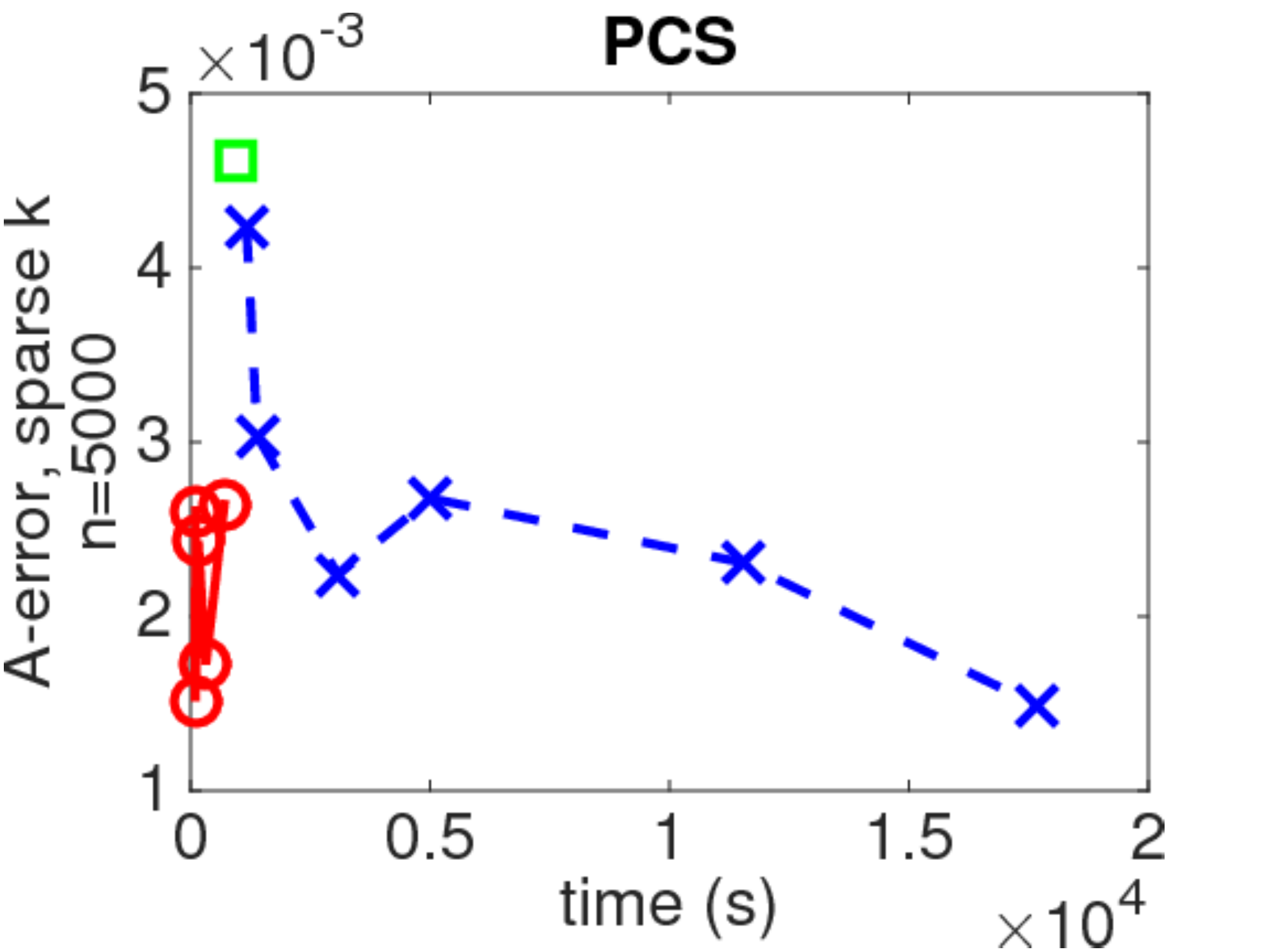}
\includegraphics[width=\halfcolwidth]{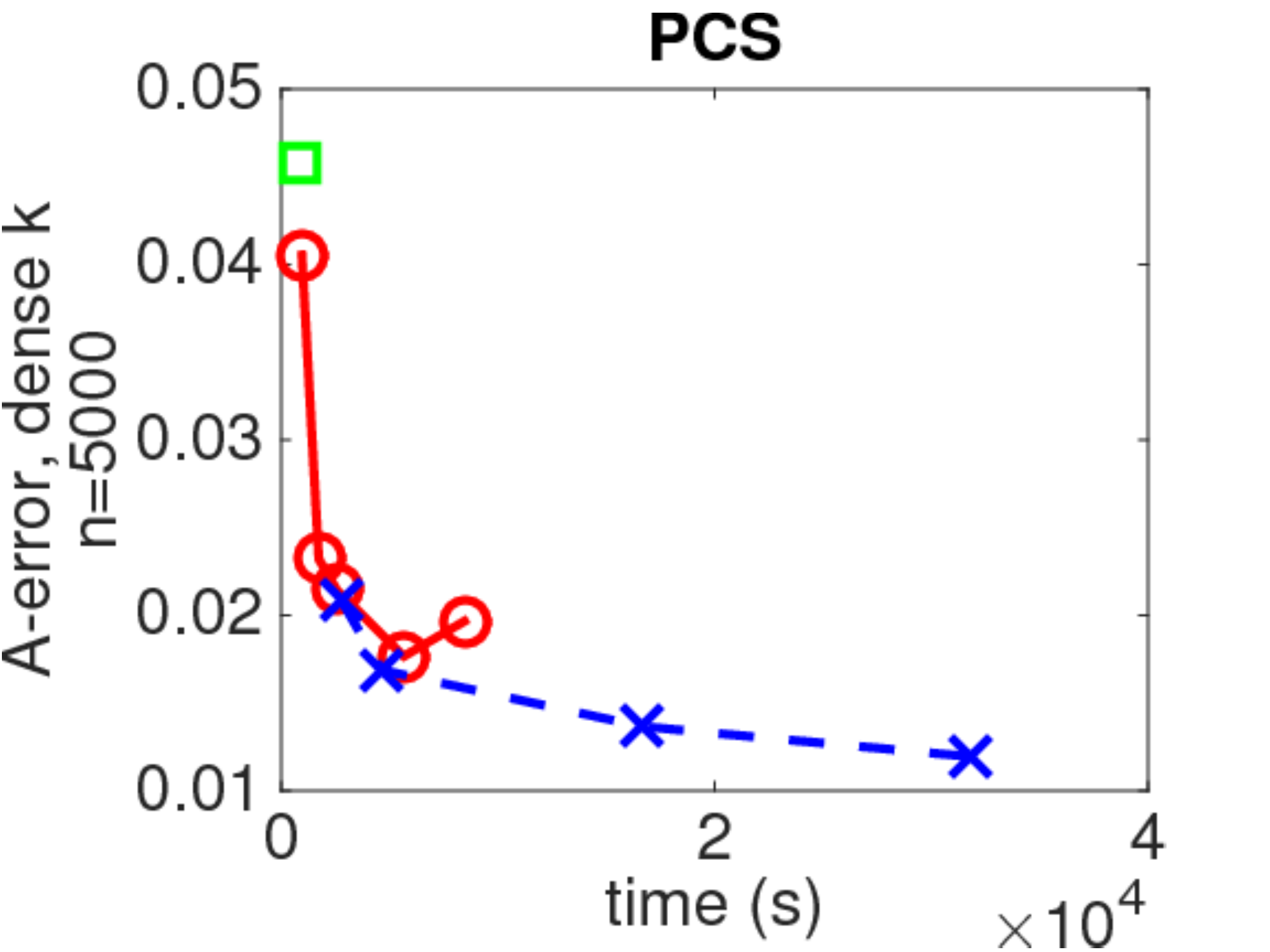}
\end{tabular}
\vspace{-4mm}
\caption{$\mathcal{E}_A$  vs. time, $n=\{500,1000,5000\}$, Left : $k$ sparse, Right : $k$ dense, piecewise constant half-planes, 
\colorlegend}
\vspace{-3mm}
\label{fig:AerrPCS}
\end{figure}

\begin{figure}[h]
\center
\begin{tabular}{c}
  \includegraphics[width=\halfcolwidth]{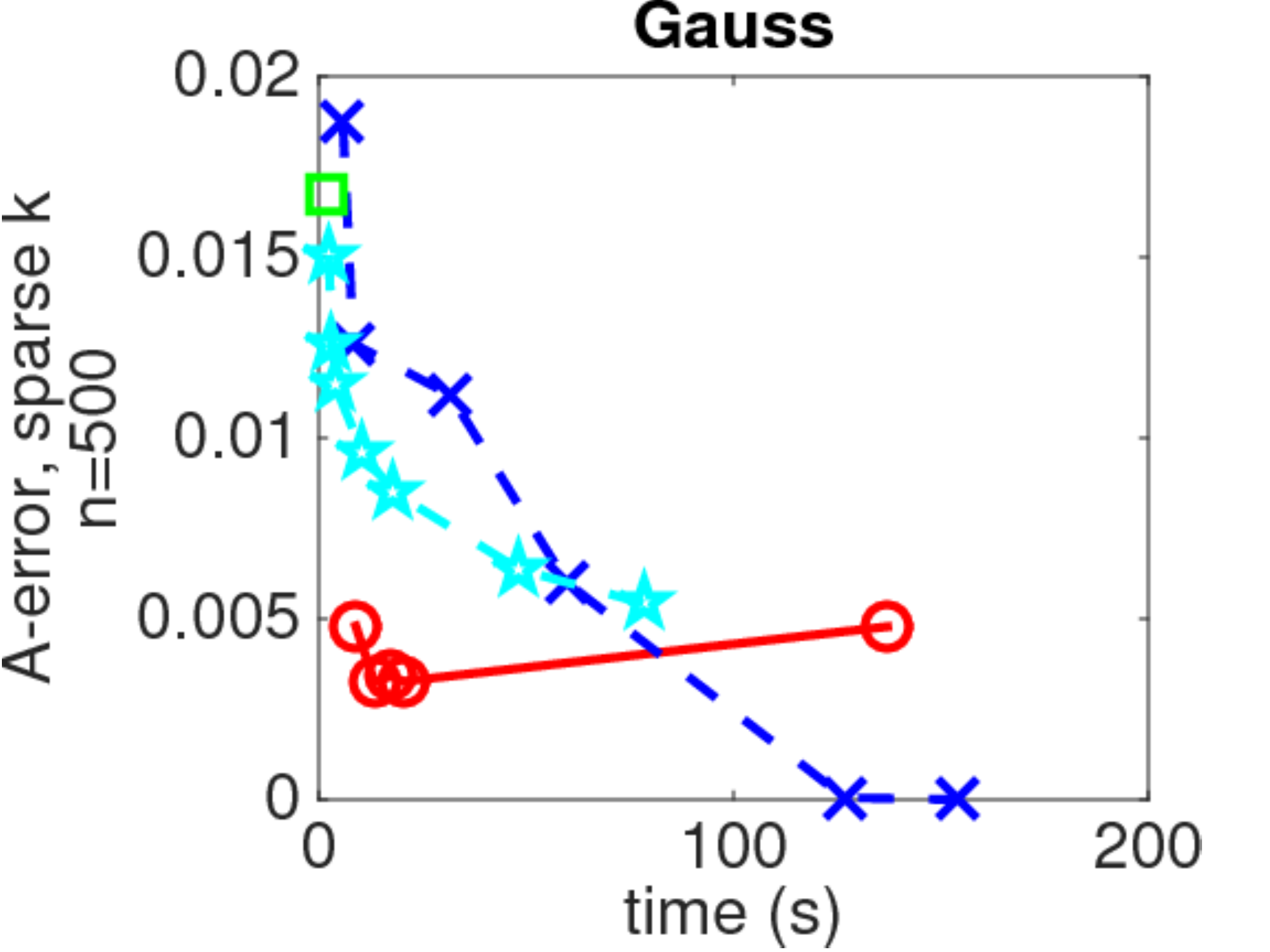}
  \includegraphics[width=\halfcolwidth]{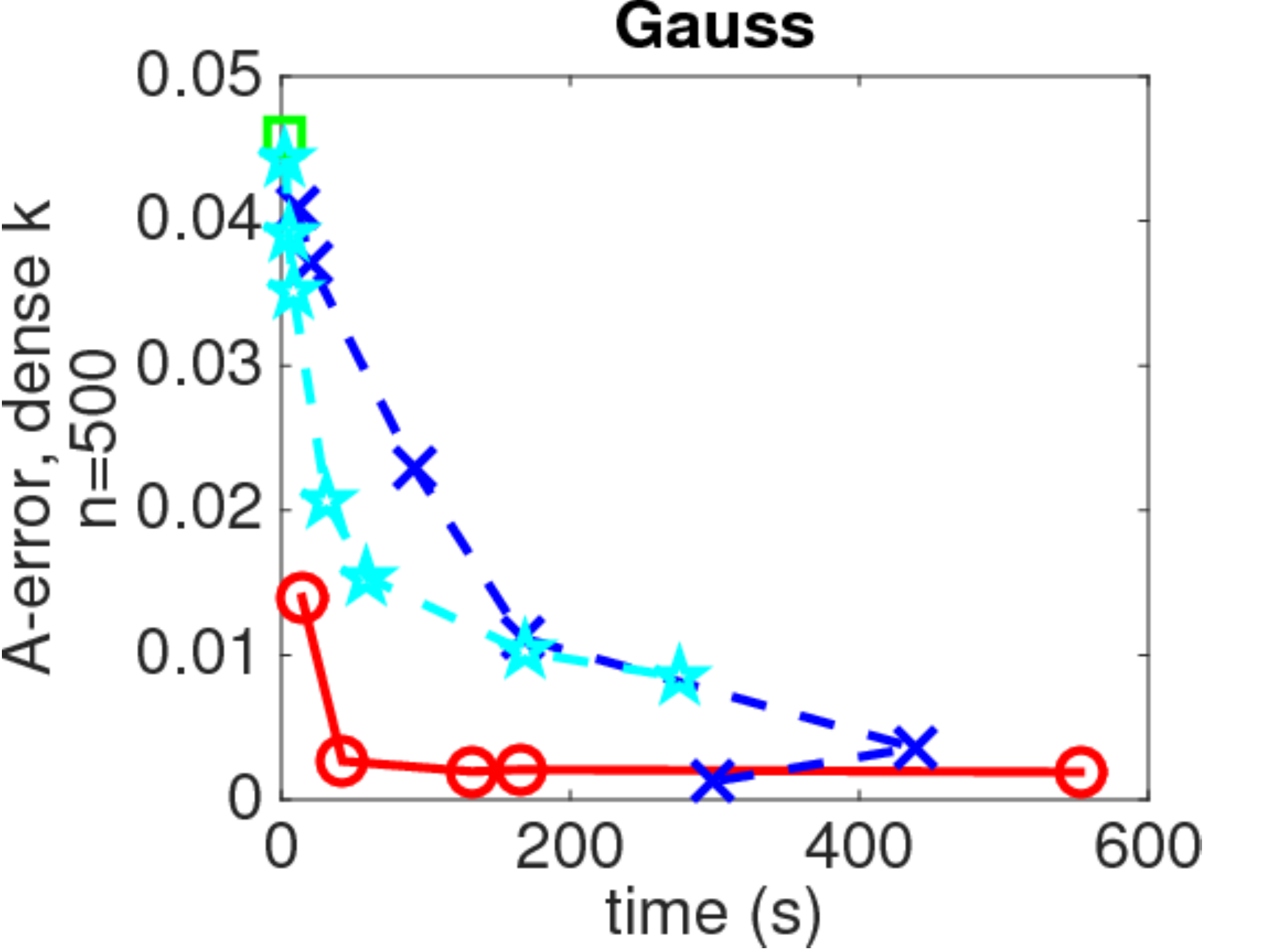}\\
  \includegraphics[width=\halfcolwidth]{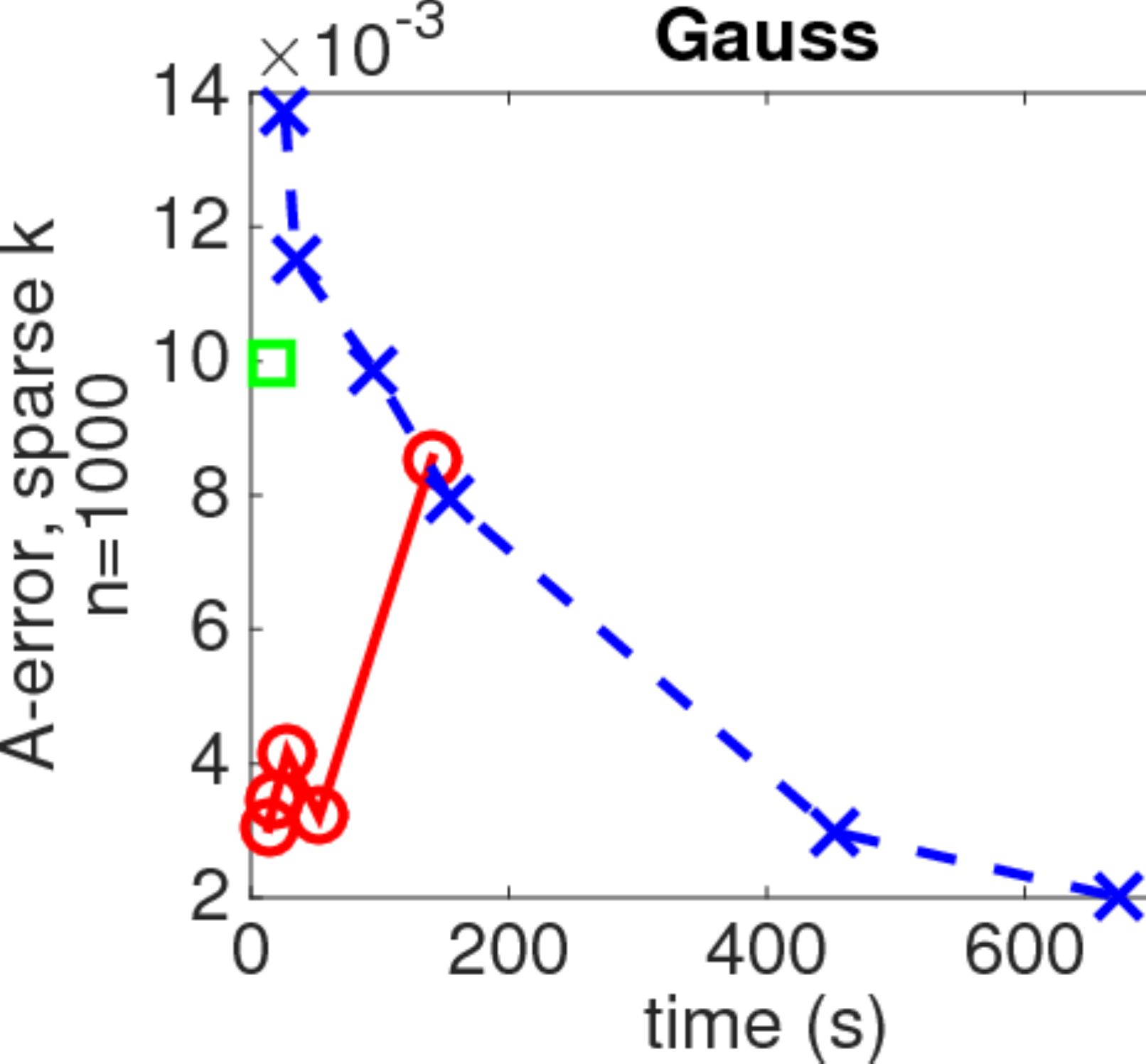}
  \includegraphics[width=\halfcolwidth]{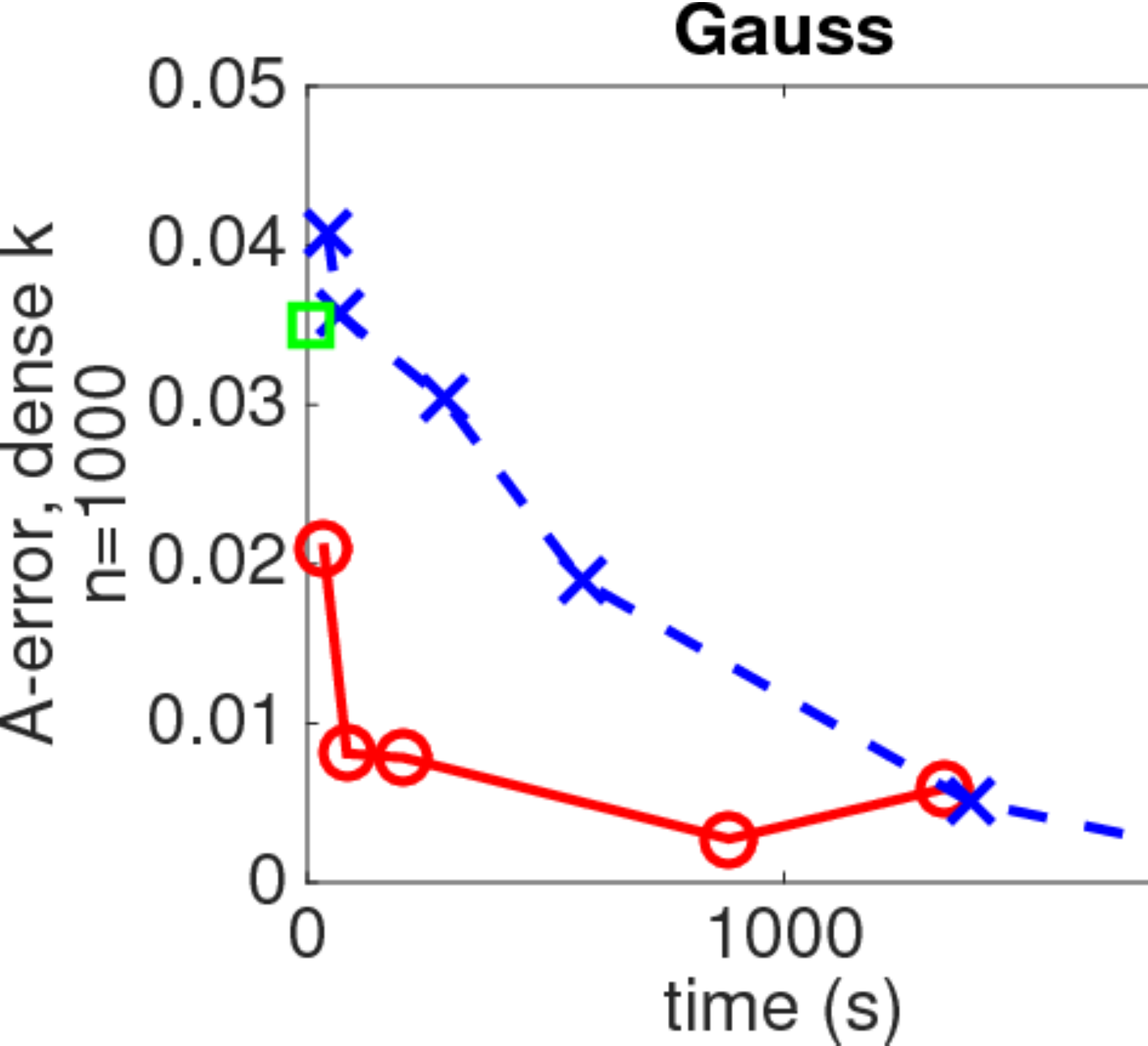}\\
  \includegraphics[width=\halfcolwidth]{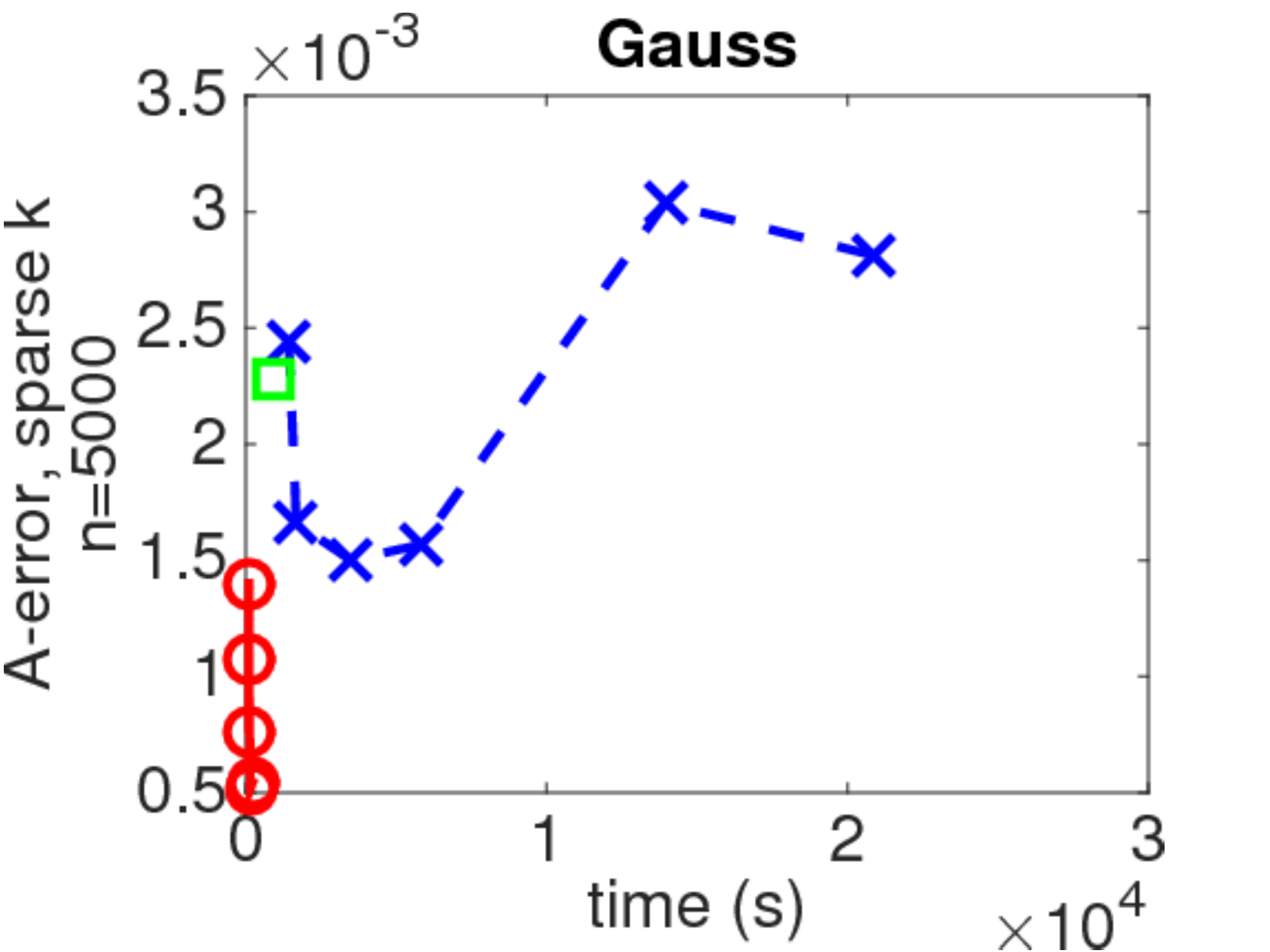}
  \includegraphics[width=\halfcolwidth]{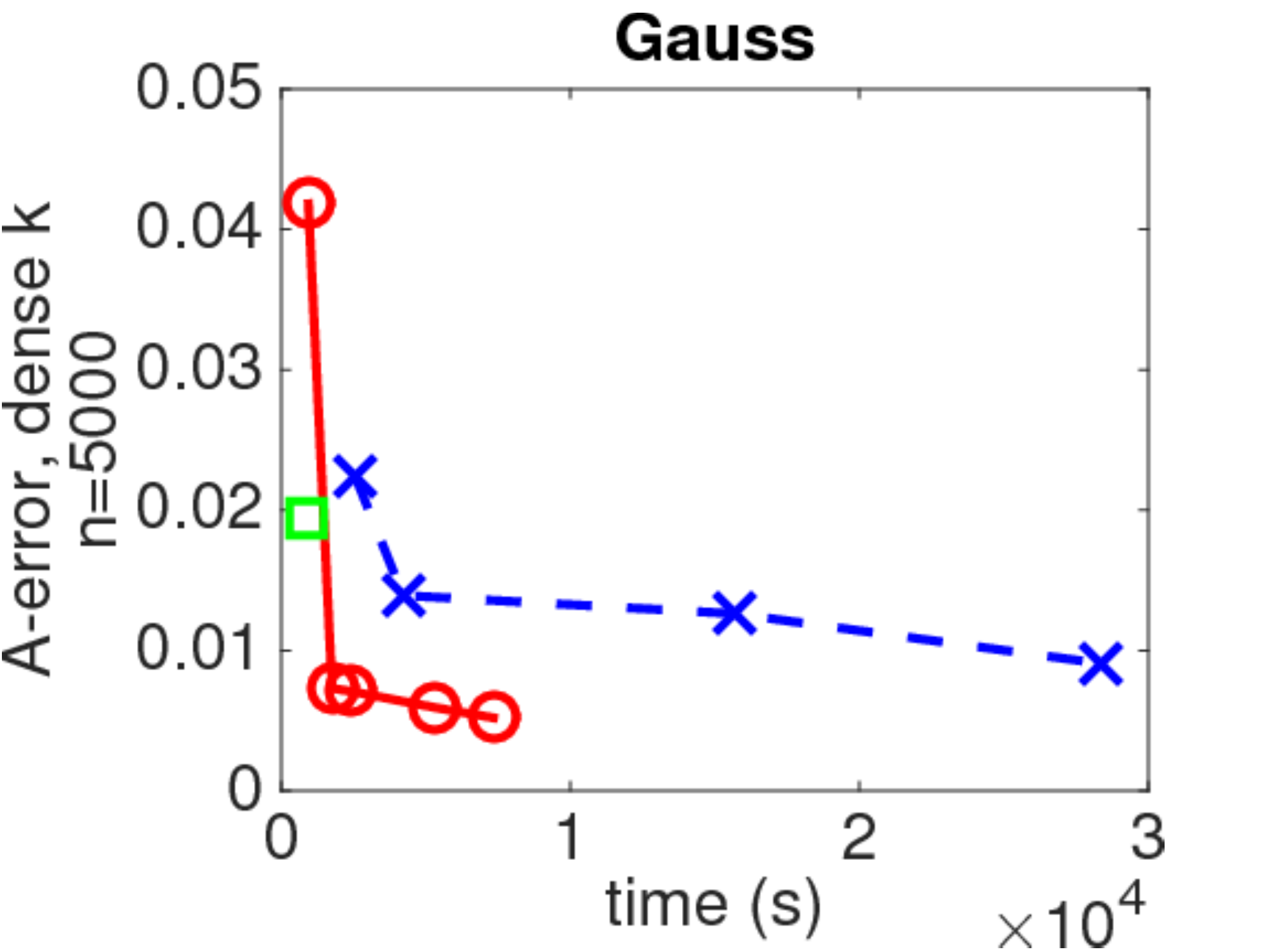}
\end{tabular}
\vspace{-4mm}
\caption{$\mathcal{E}_A$  vs. time, $n=\{500,1000,5000\}$, Left : $k$ sparse, Right : $k$ dense, Gaussian density 
\colorlegend}
\vspace{-3mm}
\label{fig:AerrGauss}
\end{figure}

\begin{figure}[h]
\center
\begin{tabular}{c}
  \includegraphics[width=\halfcolwidth]{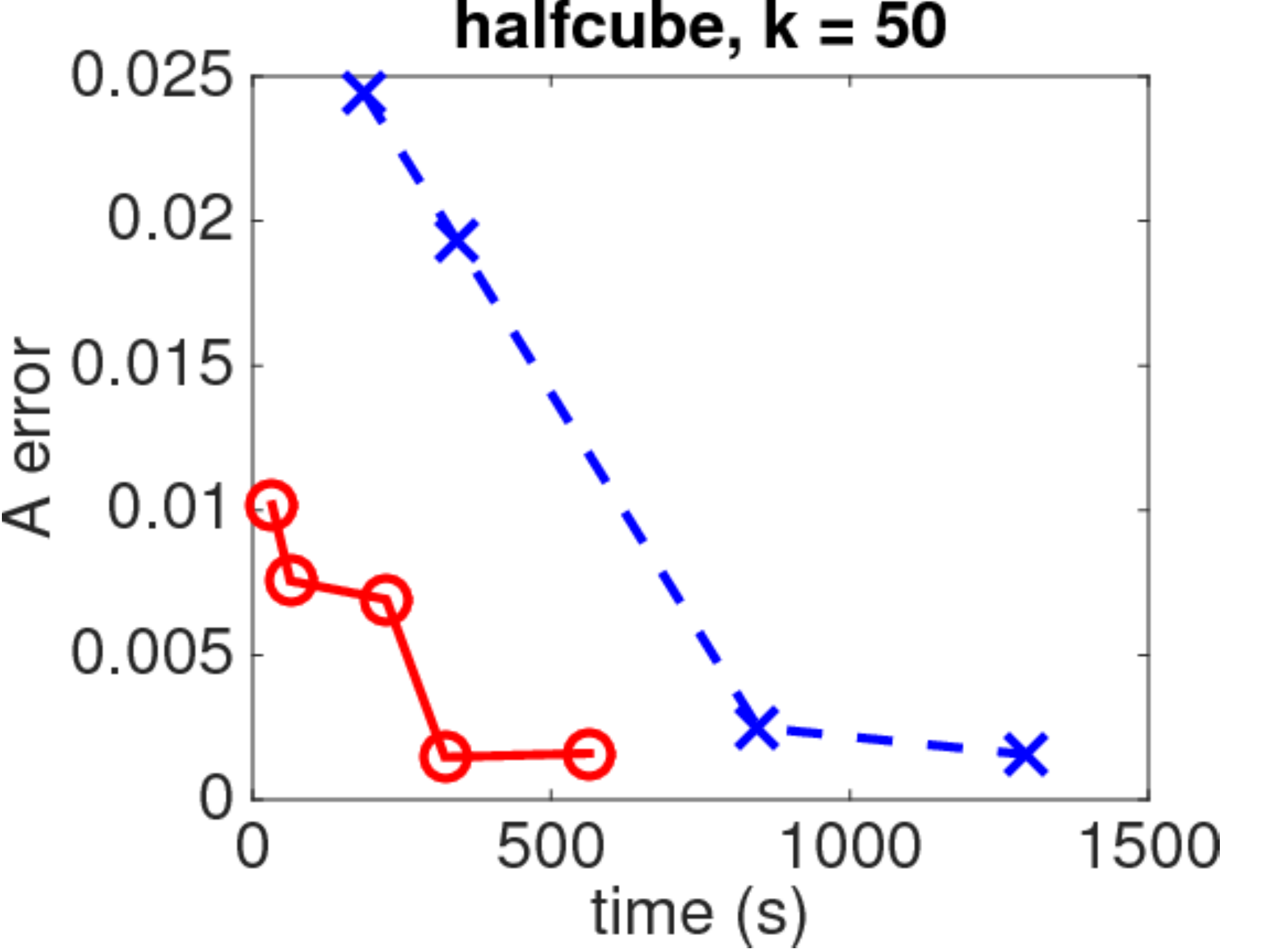}
  \includegraphics[width=\halfcolwidth]{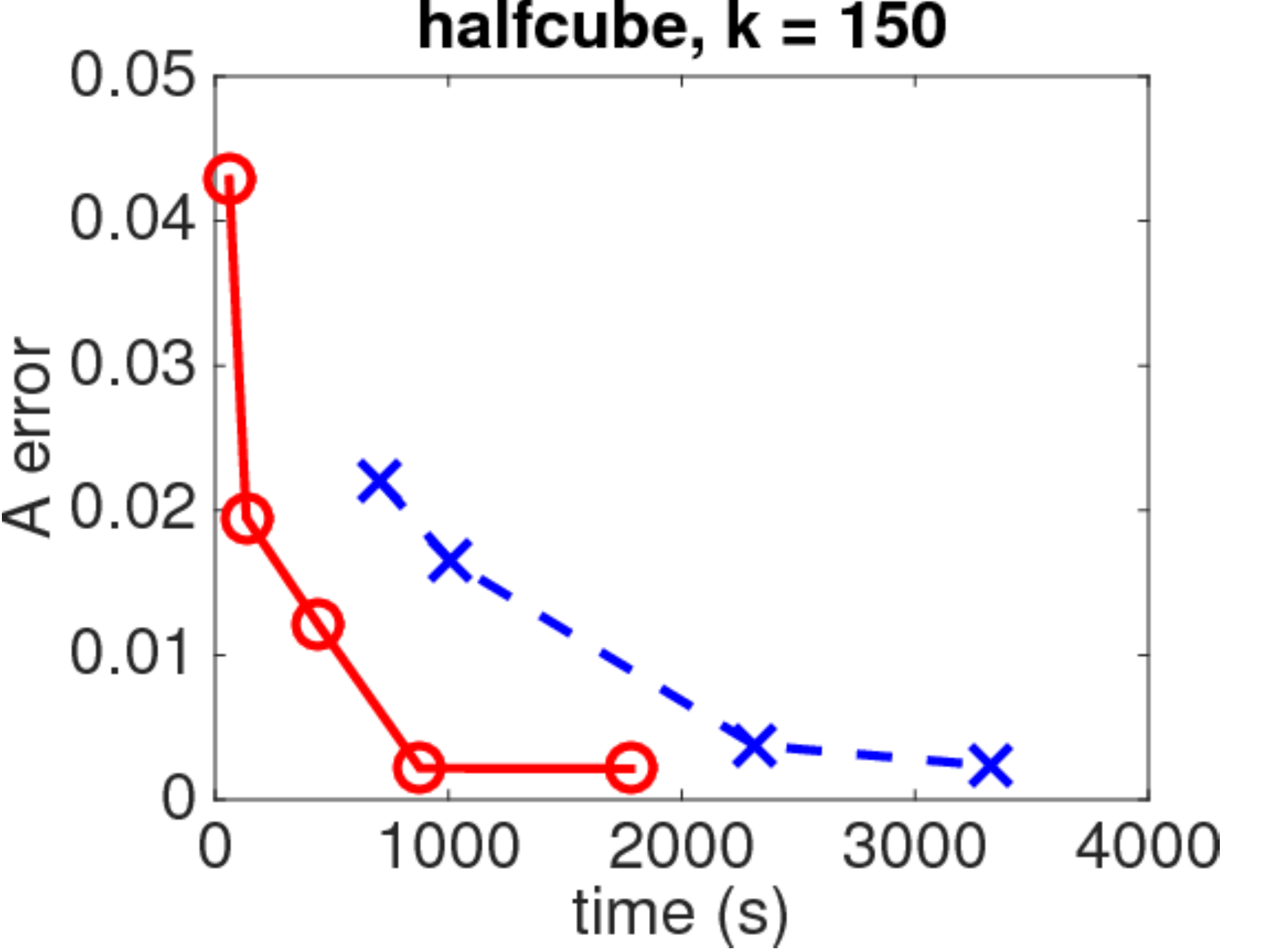}\\
  \includegraphics[width=\halfcolwidth]{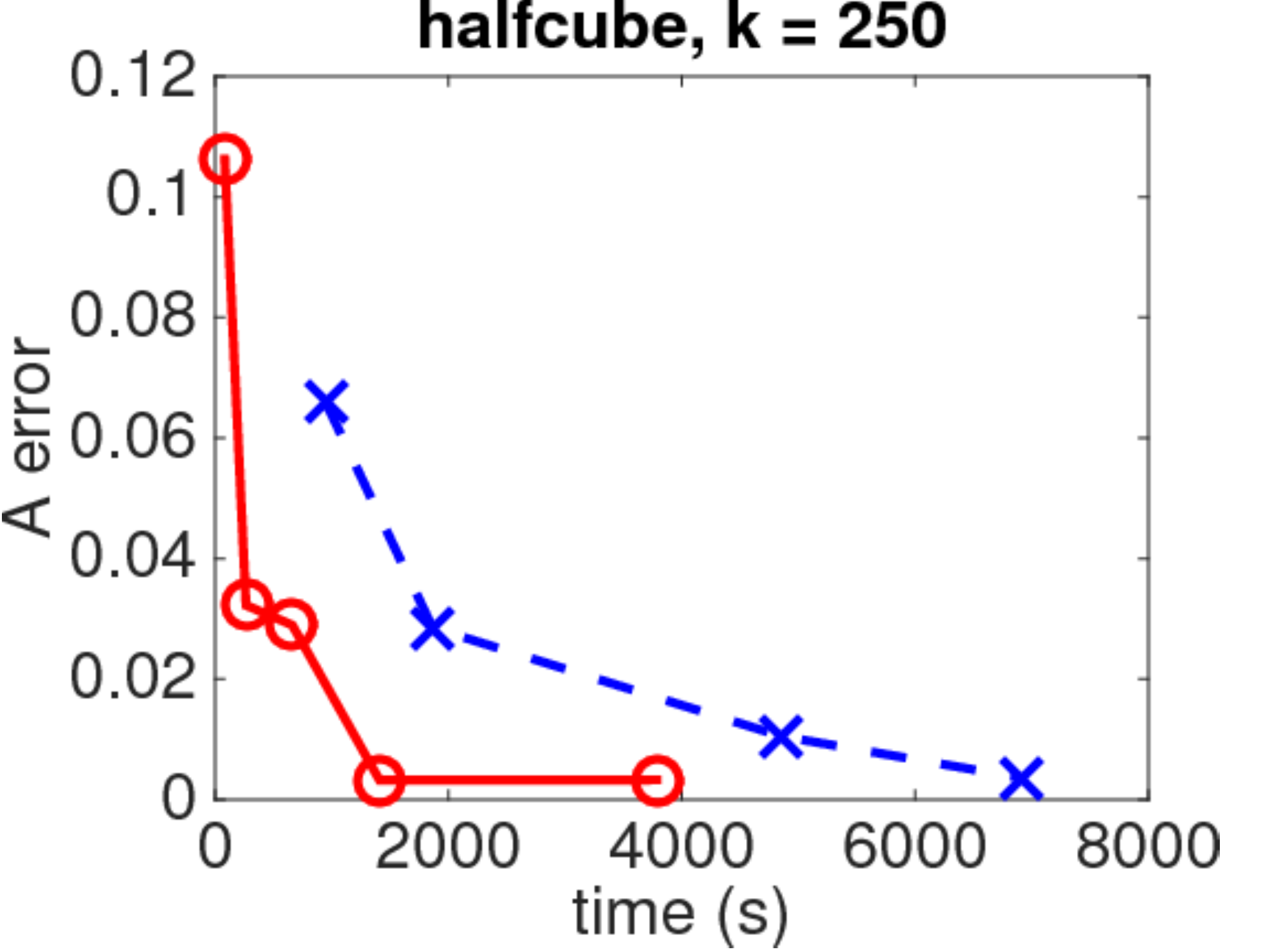}
  \includegraphics[width=\halfcolwidth]{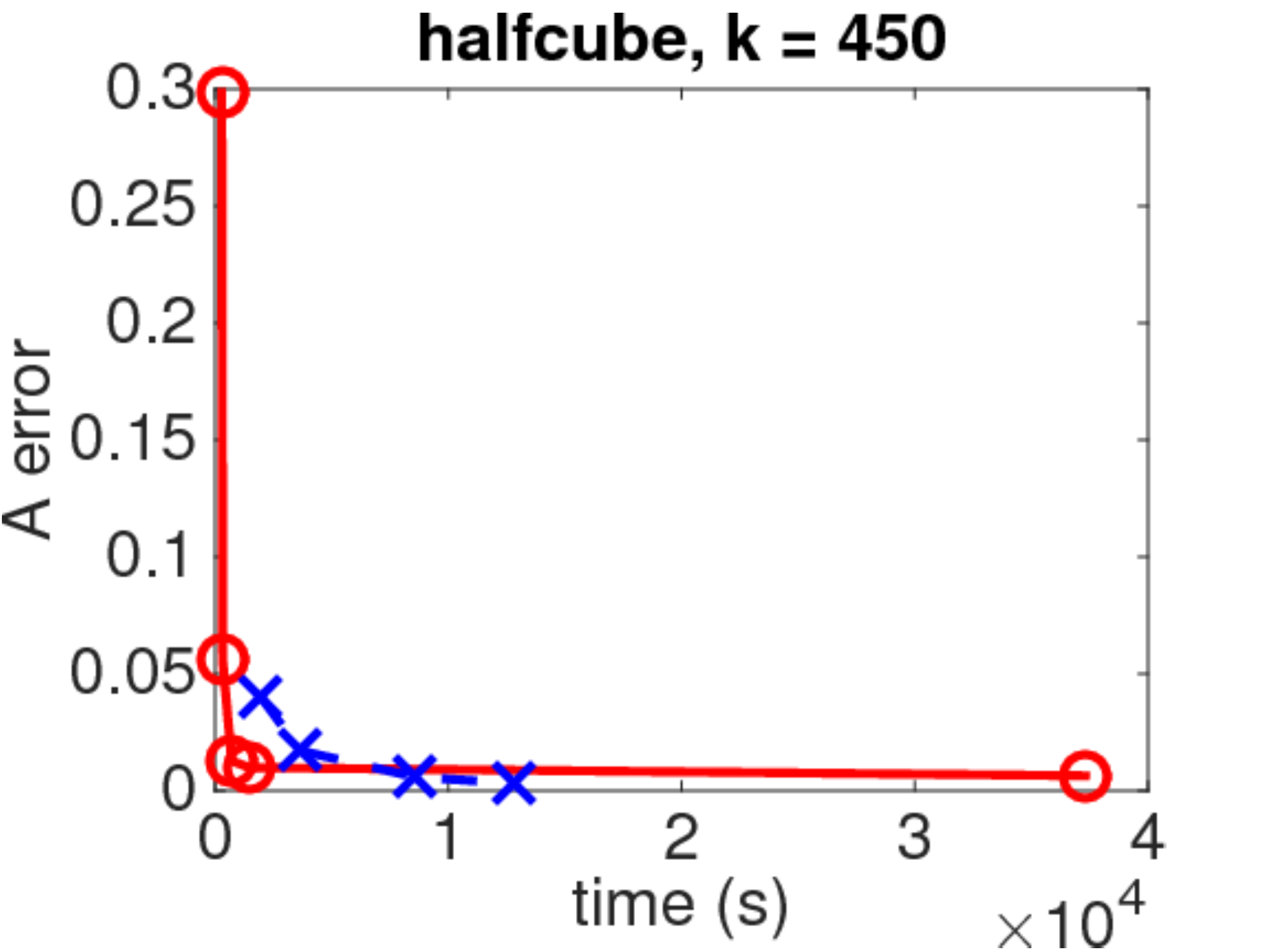}
\end{tabular}
\vspace{-4mm}
\caption{$\mathcal{E}_A$  vs. time, $n=\{500,1000,5000\}$, $k = 50, 150, 250, 450$, 3D half-cube density
\colorlegend}
\vspace{-3mm}
\label{fig:AerrHalfcube}
\end{figure}

We show $\mathcal{E}_A$ versus runtime for recovering $n=\{500,1000,5000\}$ points sampled from the PC (Figure~\ref{fig:AerrPC}), PCS (Figure~\ref{fig:AerrPCS}), and Gaussian (Figure~\ref{fig:AerrGauss}), with each figure showing results in the sparse and dense $k$ regime (see Table~\ref{table:notation}). 
We also show $\mathcal{E}_A$ versus runtime for $n=\{500,1000,5000\}$ points drawn from the  halfcube (Figure~\ref{fig:AerrHalfcube}) distribution for $k = 50, 150 250, 450$.
Even for lower values of $n$, we find that ASAP LOE is often either faster than or better-performing than LOE BFGS, or both.  This seems to be especially true in the sparse $k$ domain.
This is partly due to the massively parallel embedding step in ASAP, which can take advantage of multiple cores as the problem scales.  One would expect that as $n$ continues to grow, if more processors are made available and memory increases sufficiently, the advantage of embedding parallelization would continue to increase.

\begin{figure}[h]
  \center
  \includegraphics[width=\thirdcolwidth]{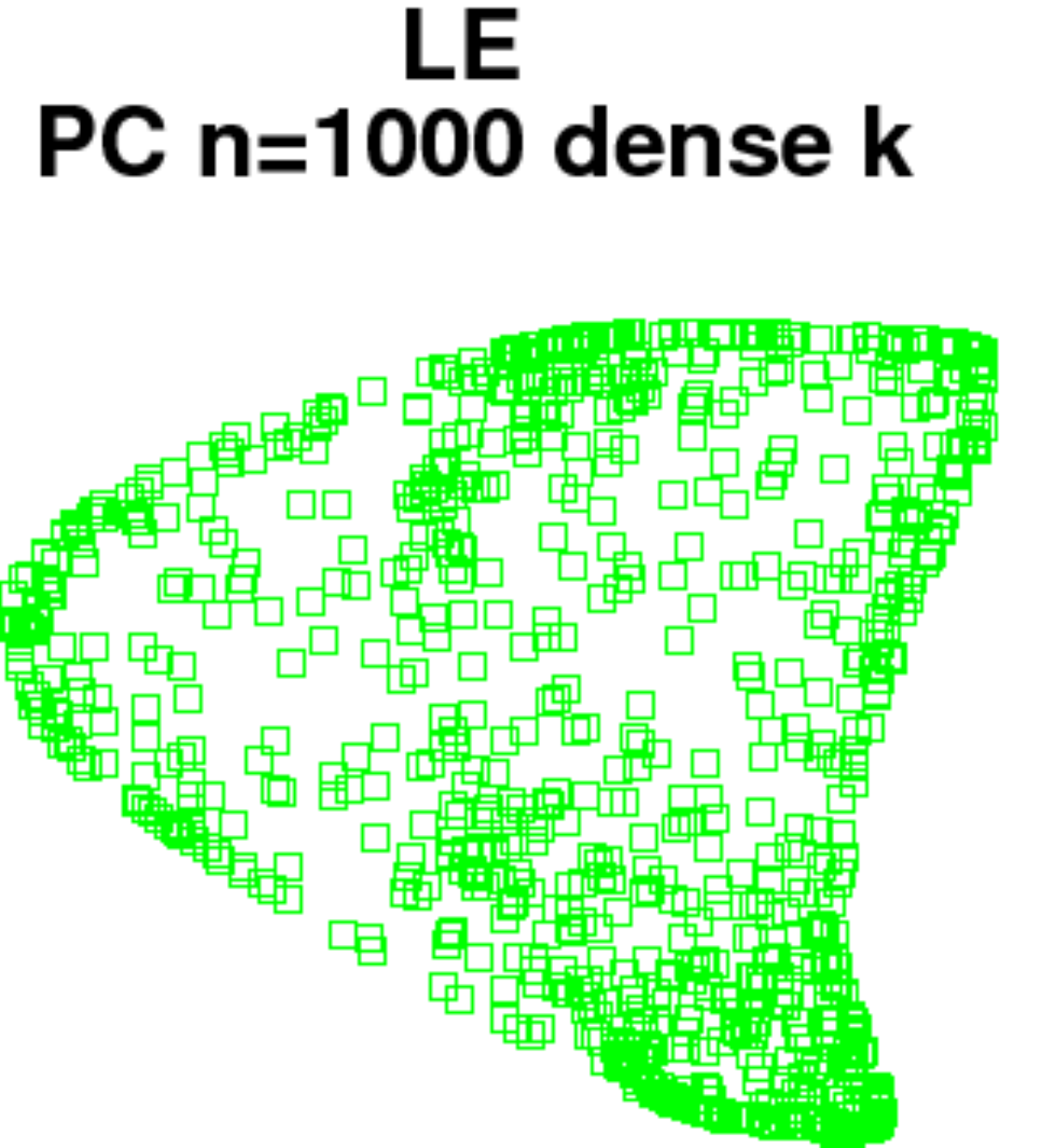}
  \includegraphics[width=\thirdcolwidth]{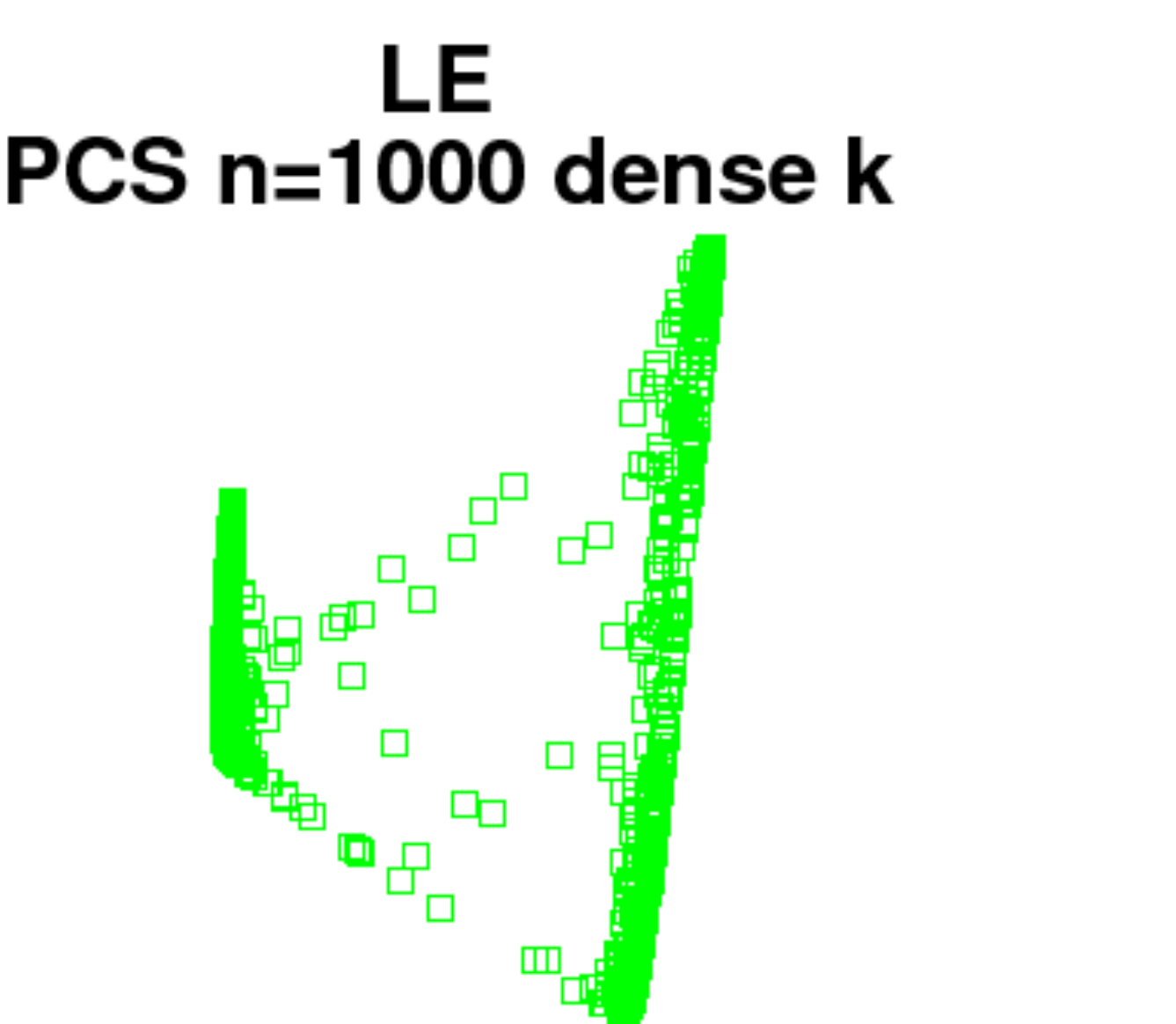}
  \includegraphics[width=\thirdcolwidth]{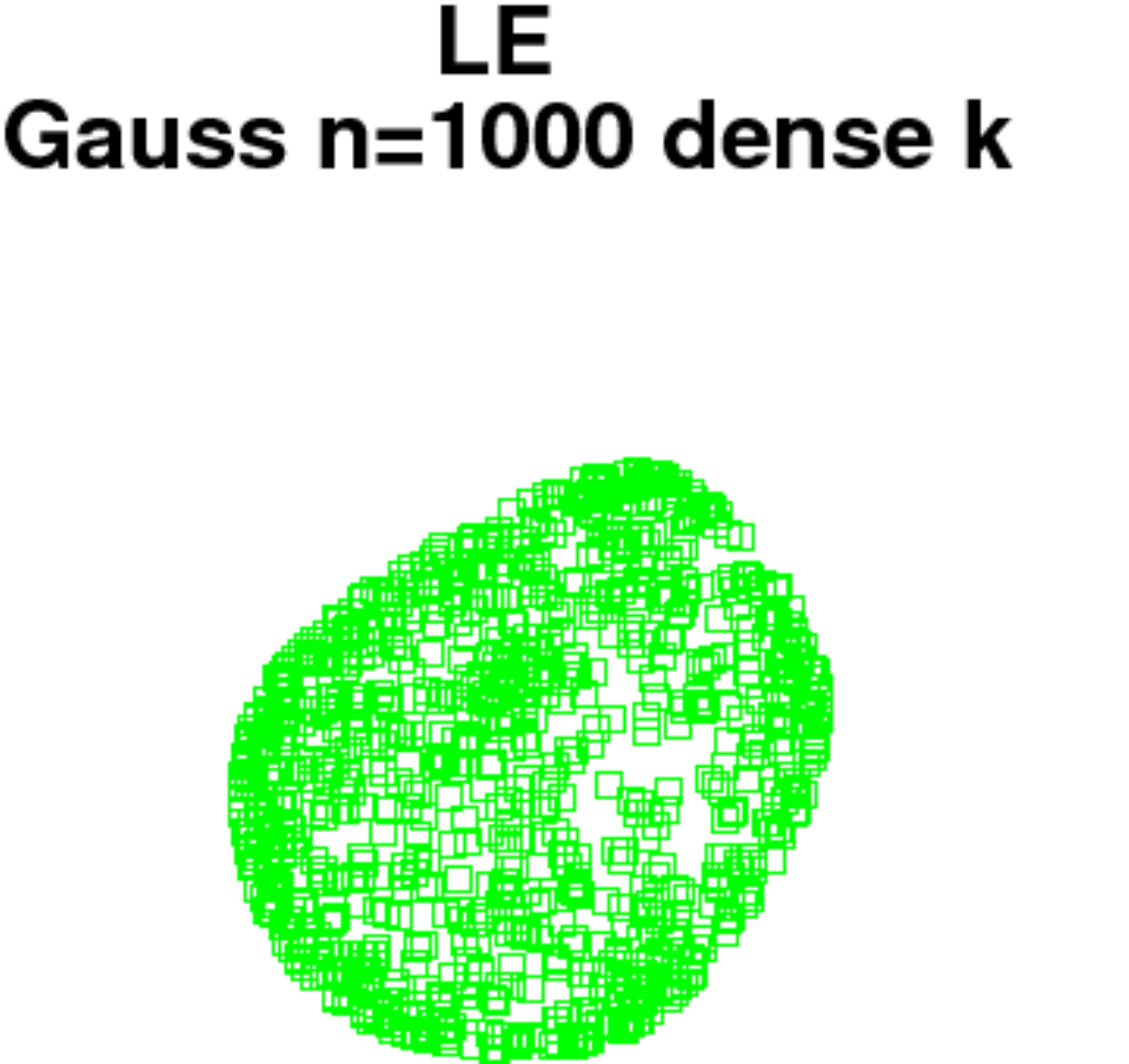}\\
\includegraphics[width=\thirdcolwidth]{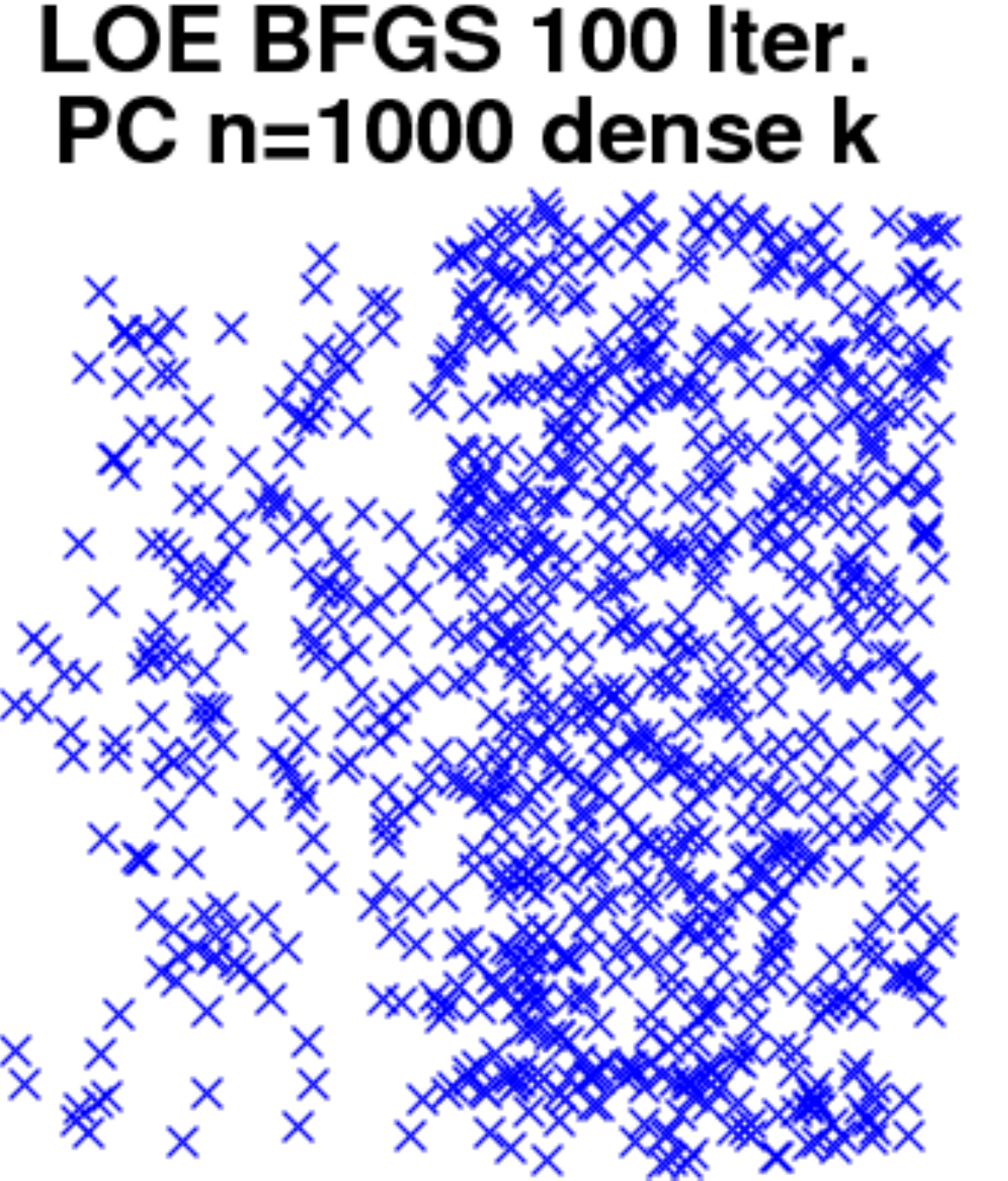}
\includegraphics[width=\thirdcolwidth]{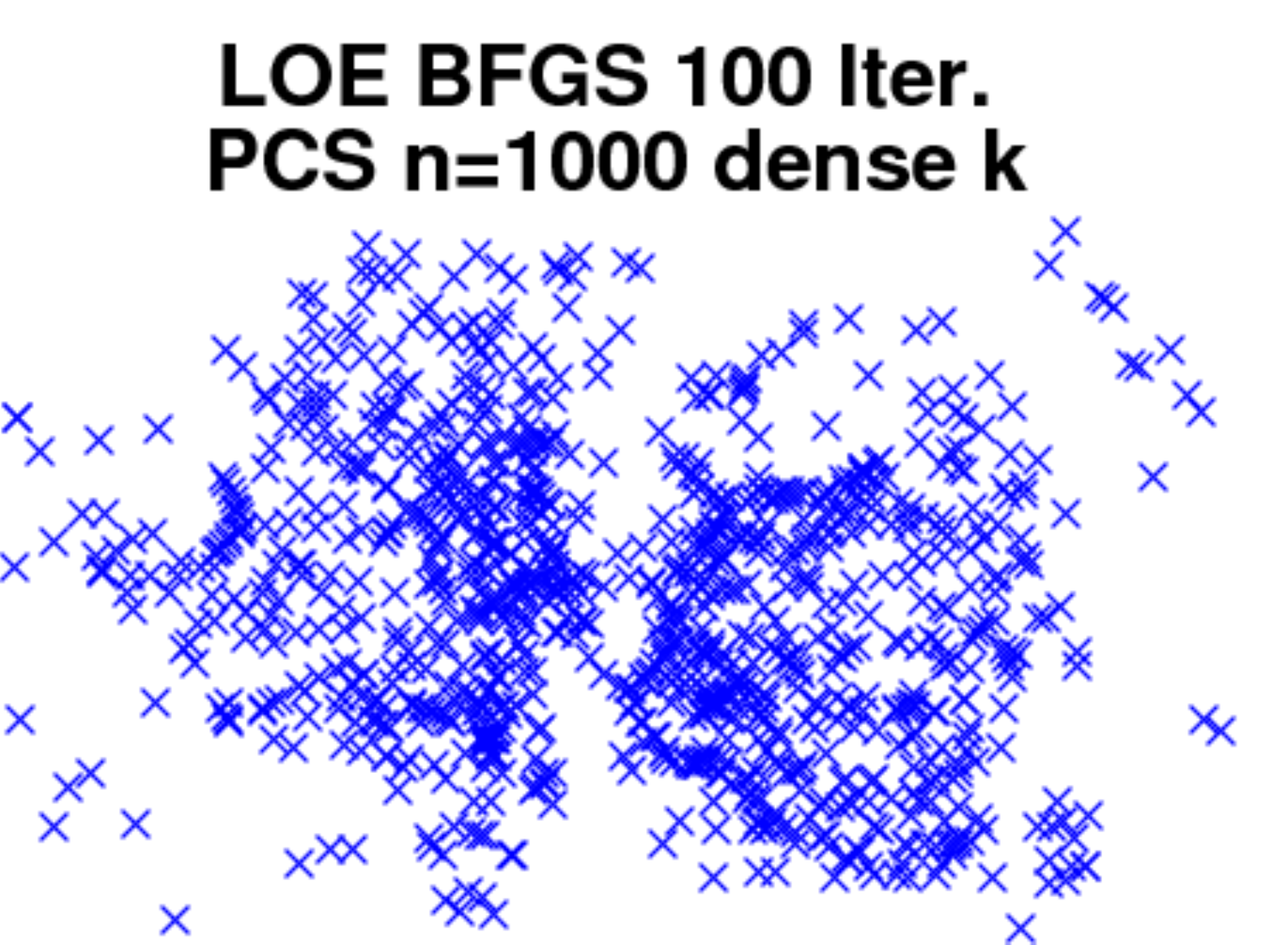}
\includegraphics[width=\thirdcolwidth]{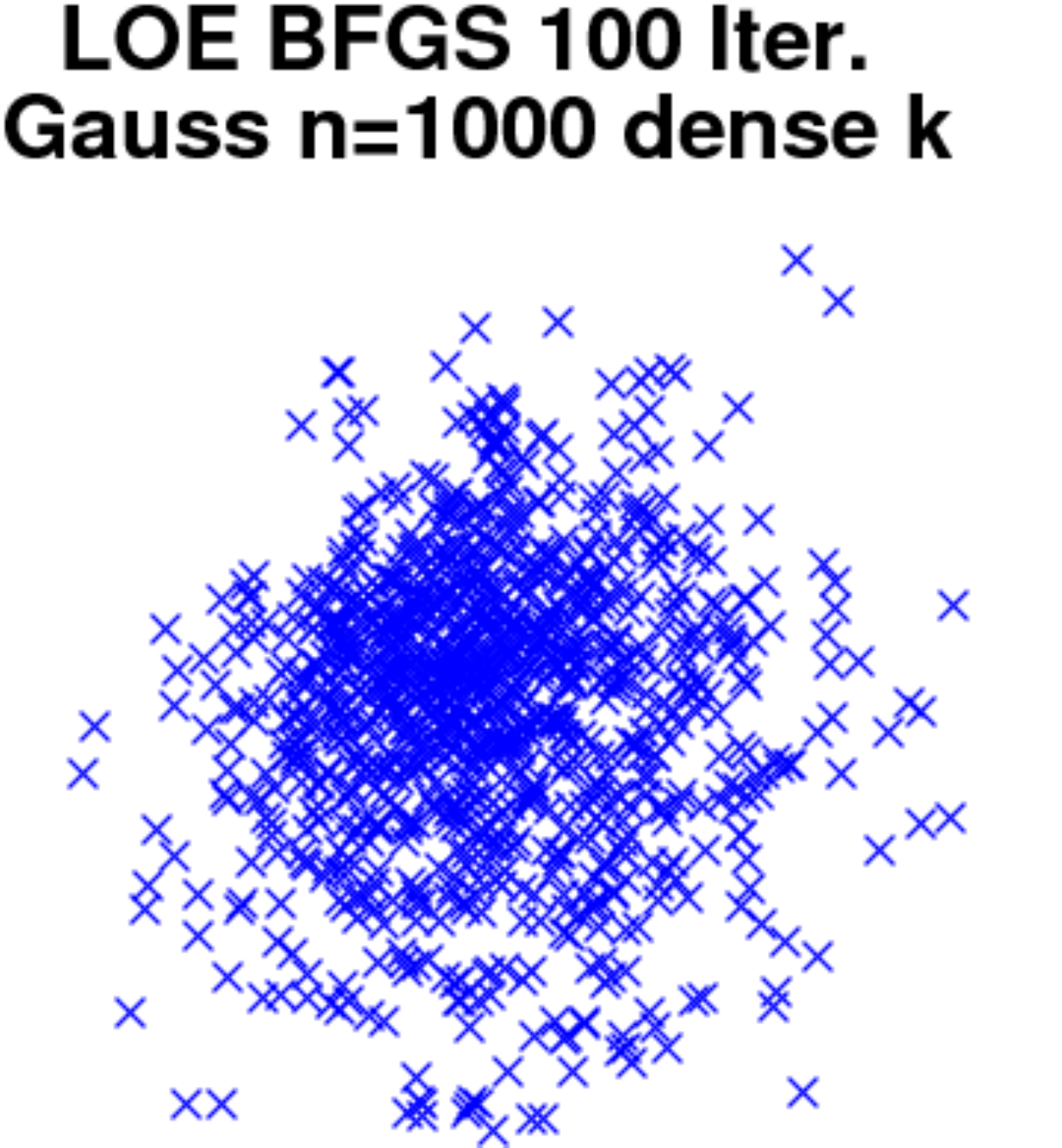}\\
\includegraphics[width=\thirdcolwidth]{ASAPLOEBFGSmp400PCm4n1000denseX.eps}
\includegraphics[width=\thirdcolwidth]{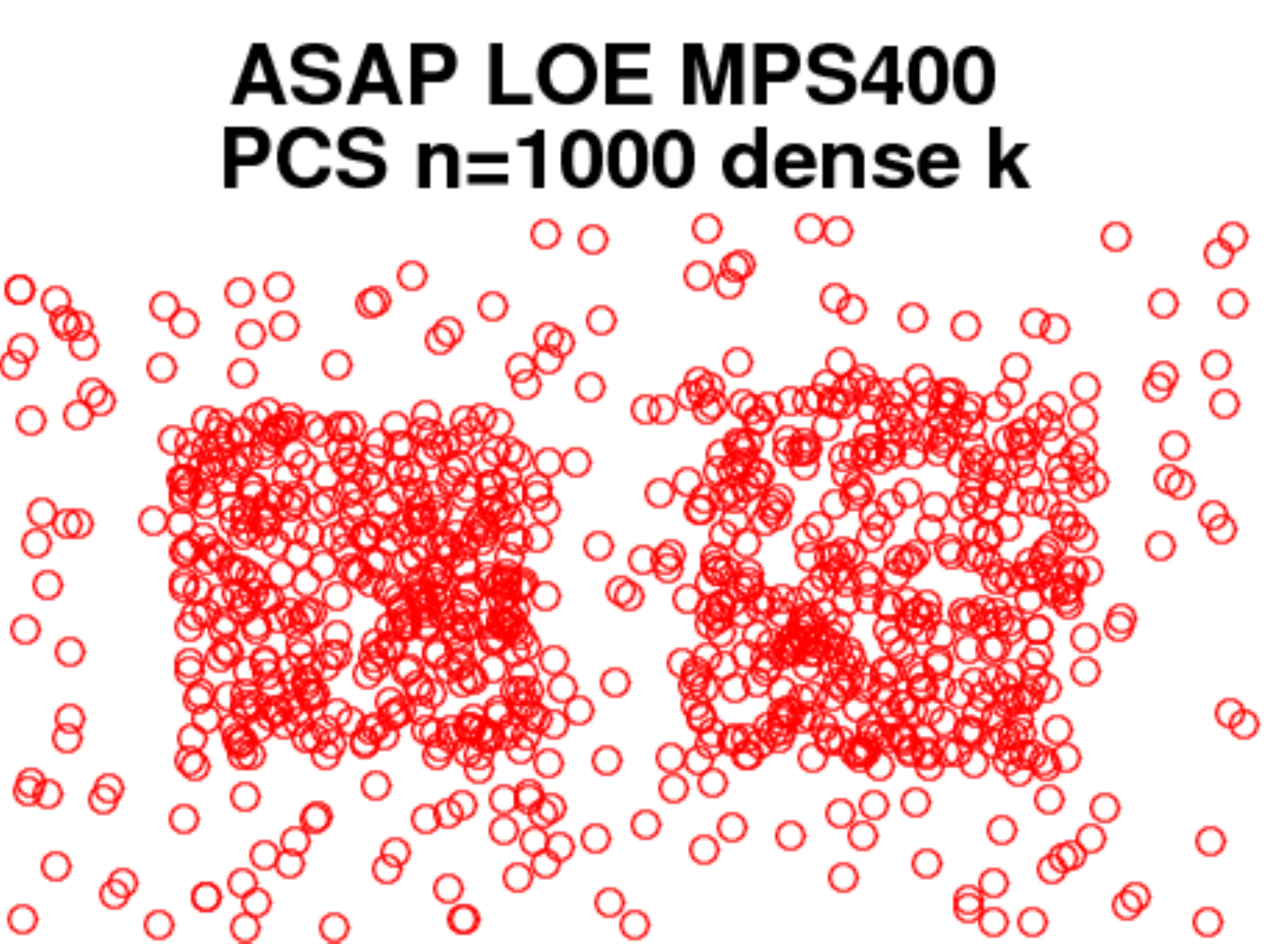}
\includegraphics[width=\thirdcolwidth]{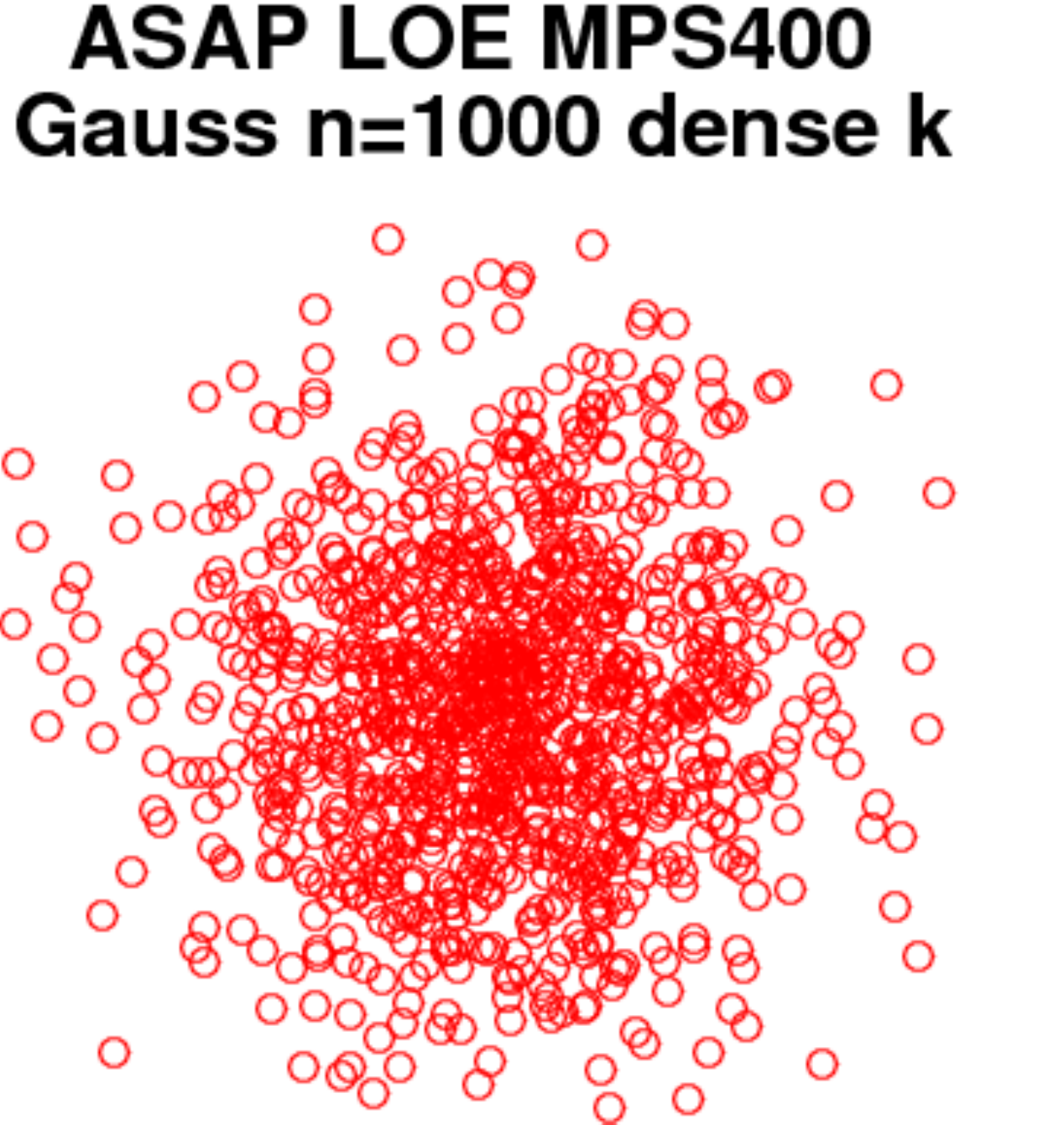}\\
\includegraphics[width=\thirdcolwidth]{PtsPCm4n1000X.eps}
\includegraphics[width=\thirdcolwidth]{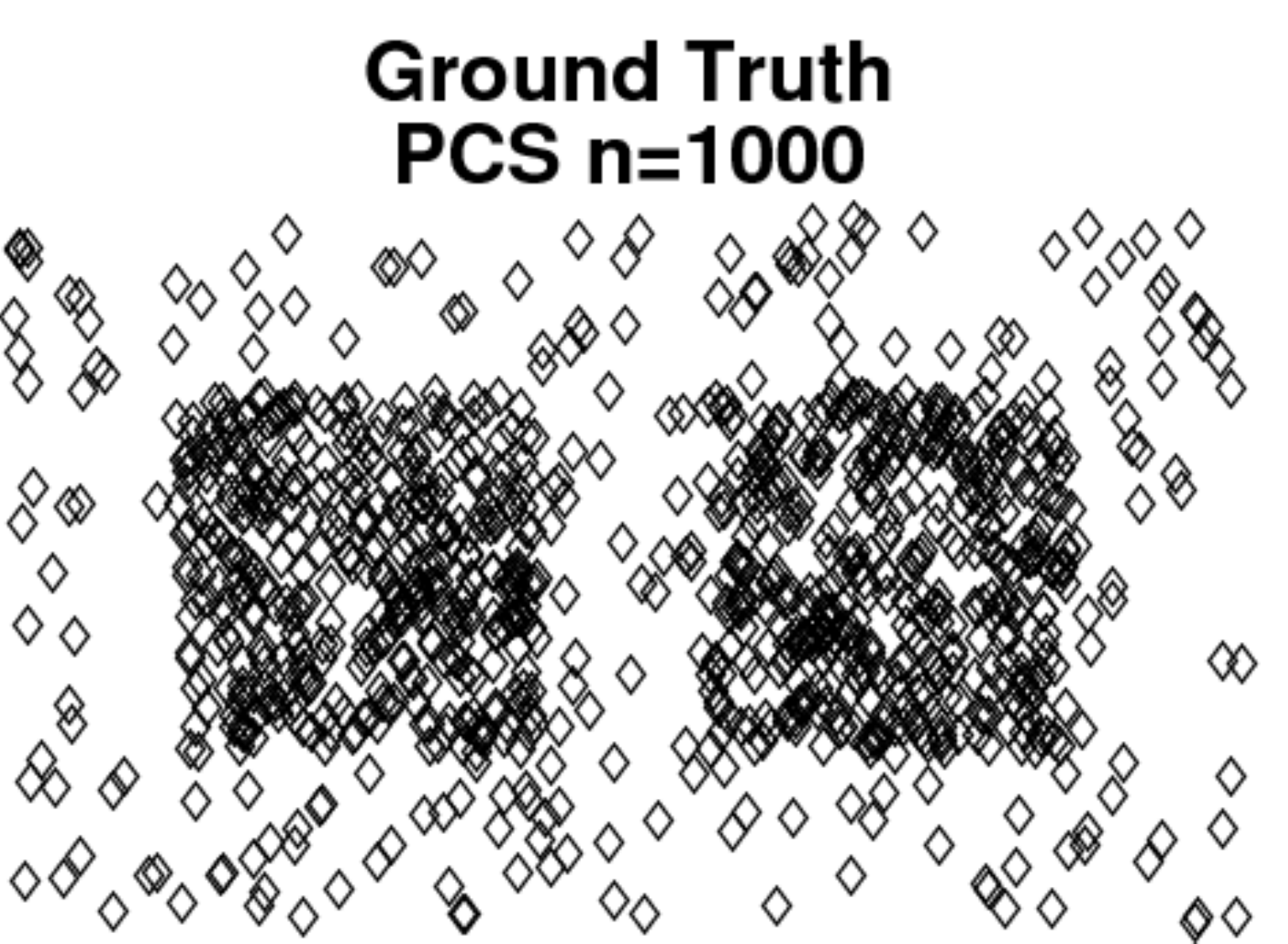}
\includegraphics[width=\thirdcolwidth]{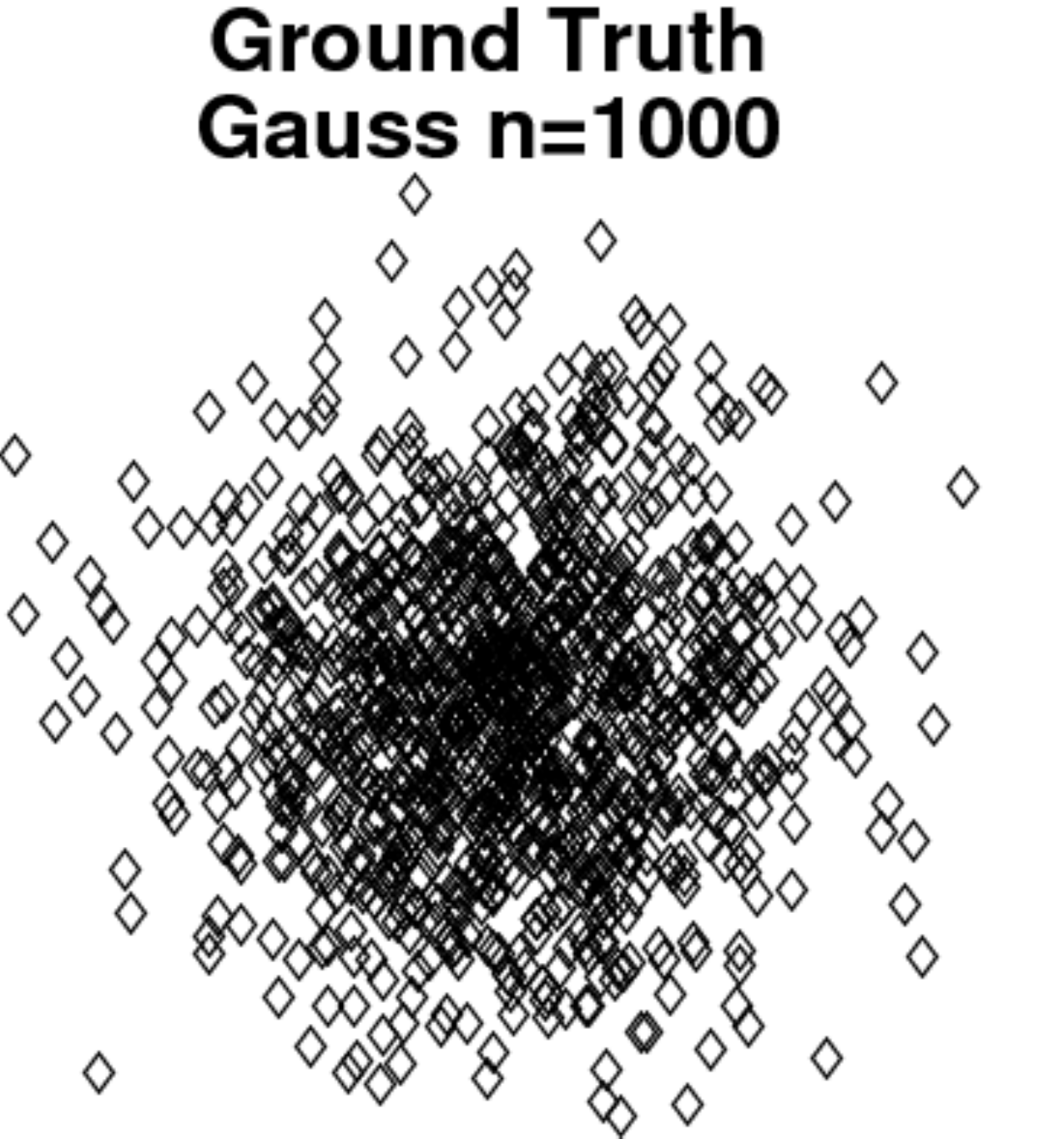}\\
\vspace{-3mm}
\caption{Embeddings for the PC (left), PCS (middle), and Gauss (right) data sets with $n=1000$, and $k$ dense.  
Row 1 : LE.
Row 2: LOE BFGS Iter.=100.
Row 3: ASAP LOE with MPS = 400 (with each ASAP result obtained is less time than the corresponding LOE result).
Row 4: ground truth.}
\vspace{-3mm}
\label{fig:X}
\end{figure}

To further illustrate how the methods perform, we plot the 
embeddings of $n=1000$ point sampled from the 2D densities in Figure~\ref{fig:X}.  In each case, the ASAP LOE with MPS=400 takes less time to run and yields smaller $\mathcal{E}_A$
errors than the LOE BFGS with 100 maximum iterations. We only run LOE MM for $n=500$ because of difficulties we had when trying to get the provided R implementation 
to run on our Linux-based remote computing resource. We ran into no problems with the LOE BFGS implementation. The computers used have 12 CPU cores which are Intel(R) Xeon(R)  X5650  @ 2.67GHz, and have 48GB ram. The R implementation of LOE does not (as far as its authors are aware) take advantage of multiple cores, and runs a single process on a single core. In contrast, our ASAP Matlab implementation uses the Multicore package 
to divide up the local embedding problems among the available cores.

\begin{figure}[h]
  \center
  \includegraphics[width=\thirdcolwidth]{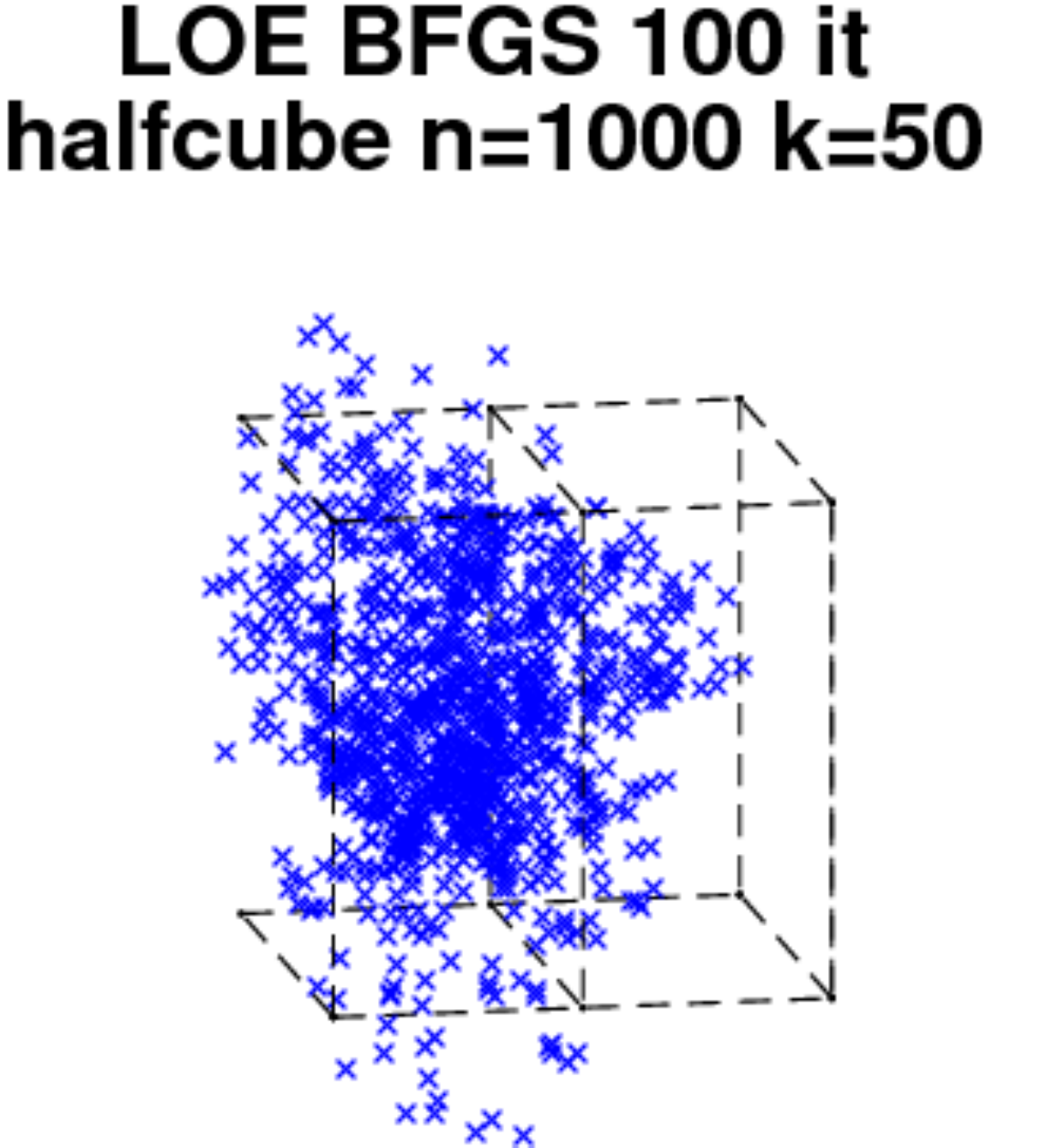}
  \includegraphics[width=\thirdcolwidth]{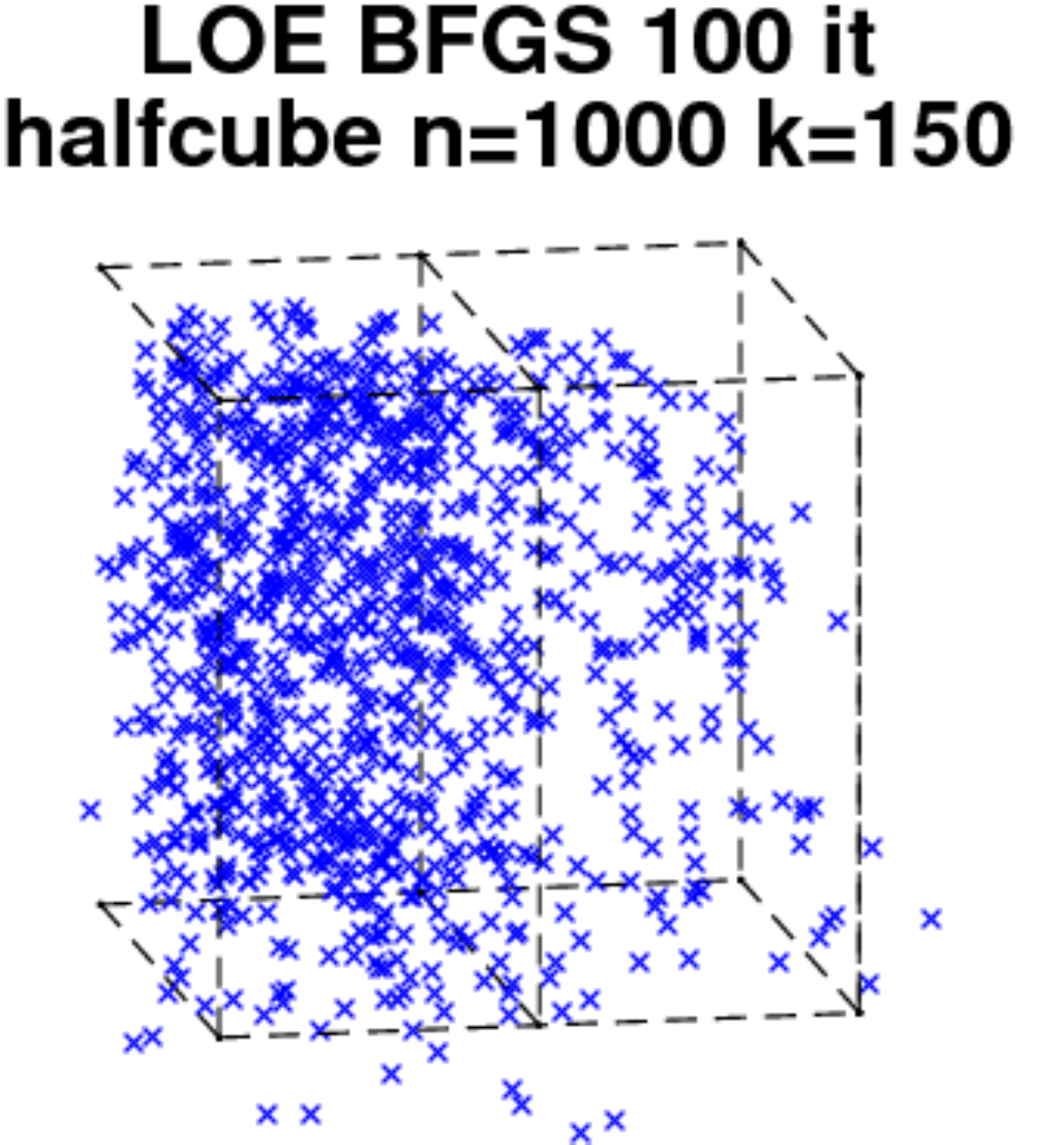}
  \includegraphics[width=\thirdcolwidth]{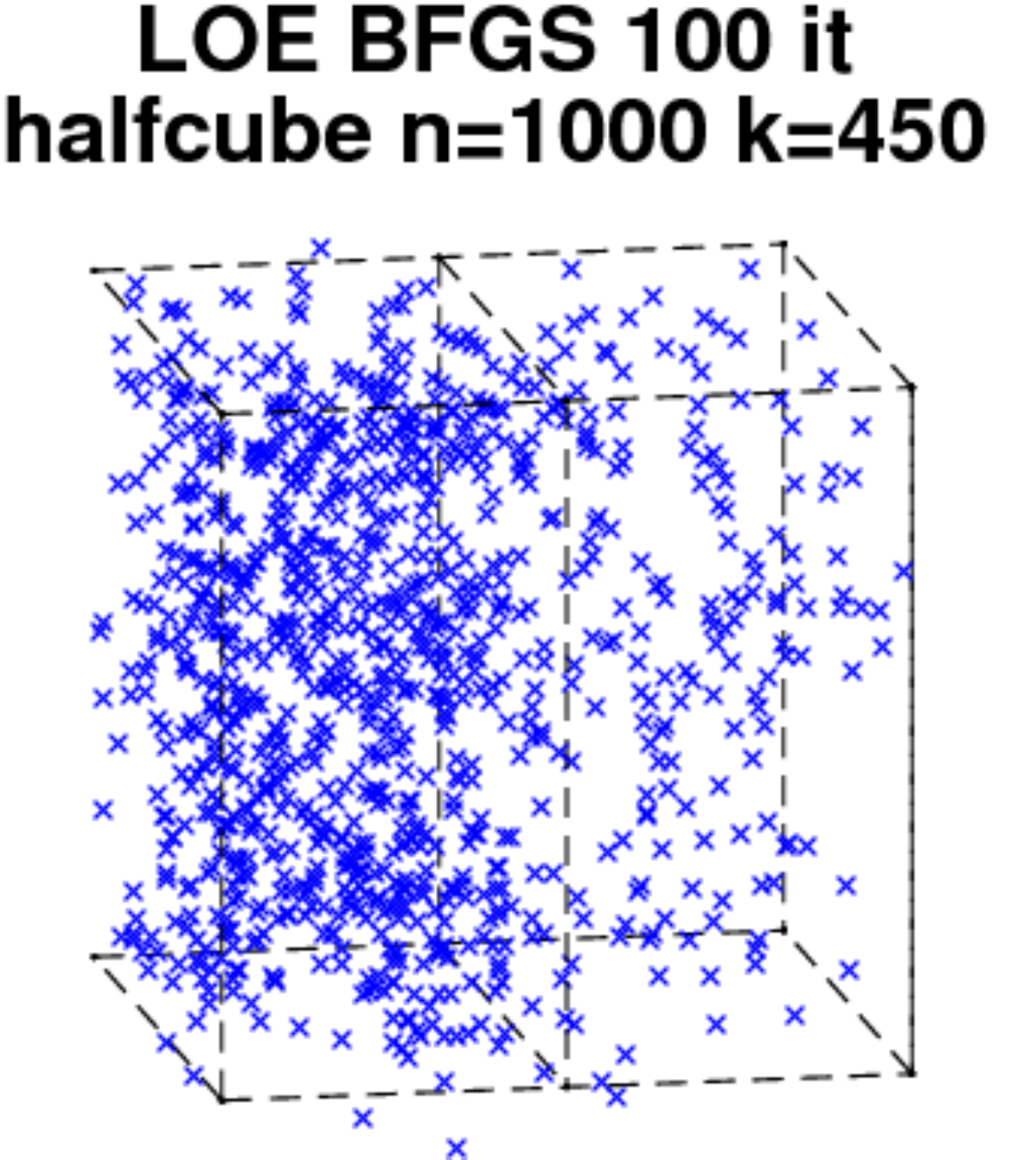}\\
  \includegraphics[width=\thirdcolwidth]{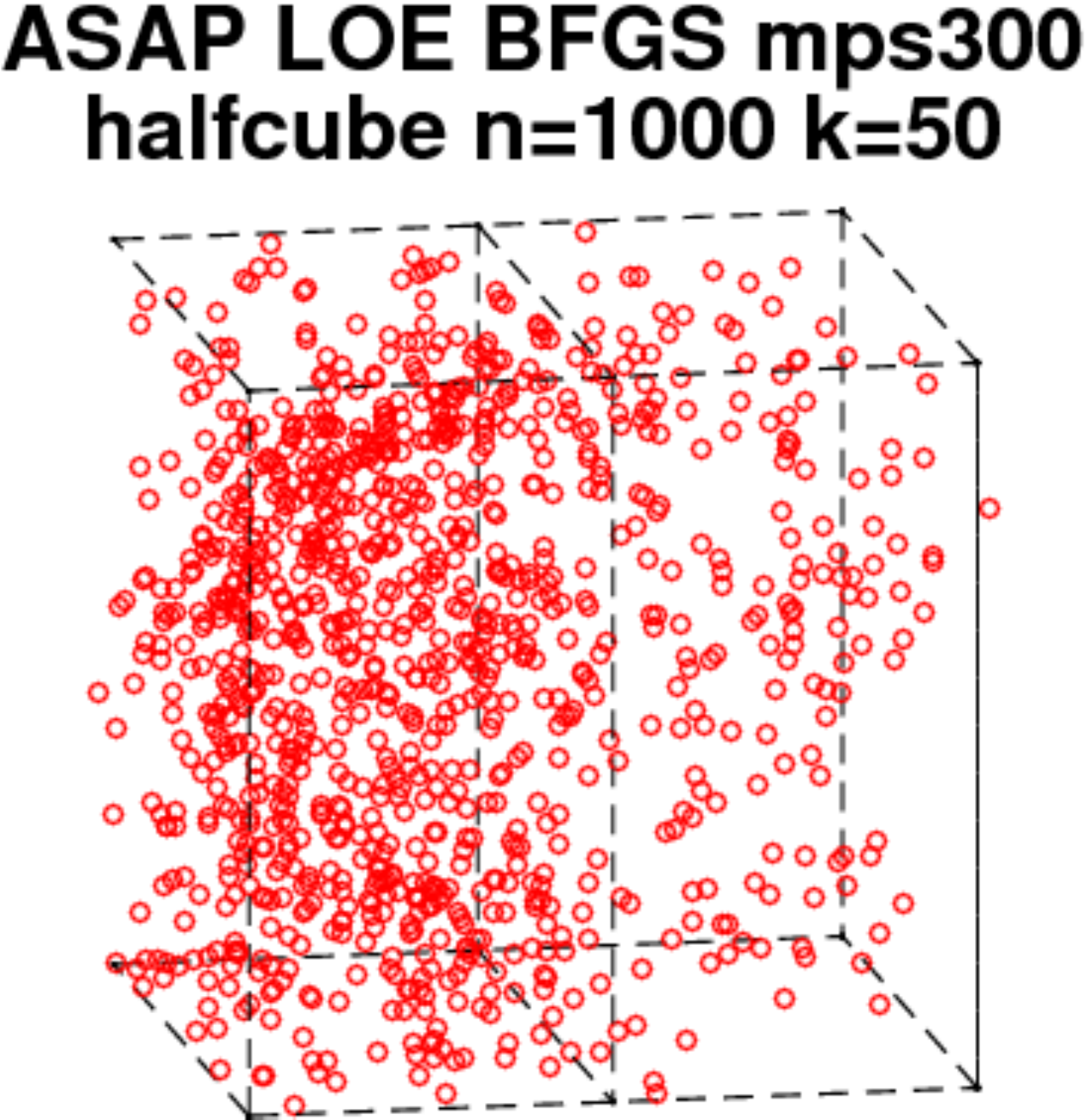}
  \includegraphics[width=\thirdcolwidth]{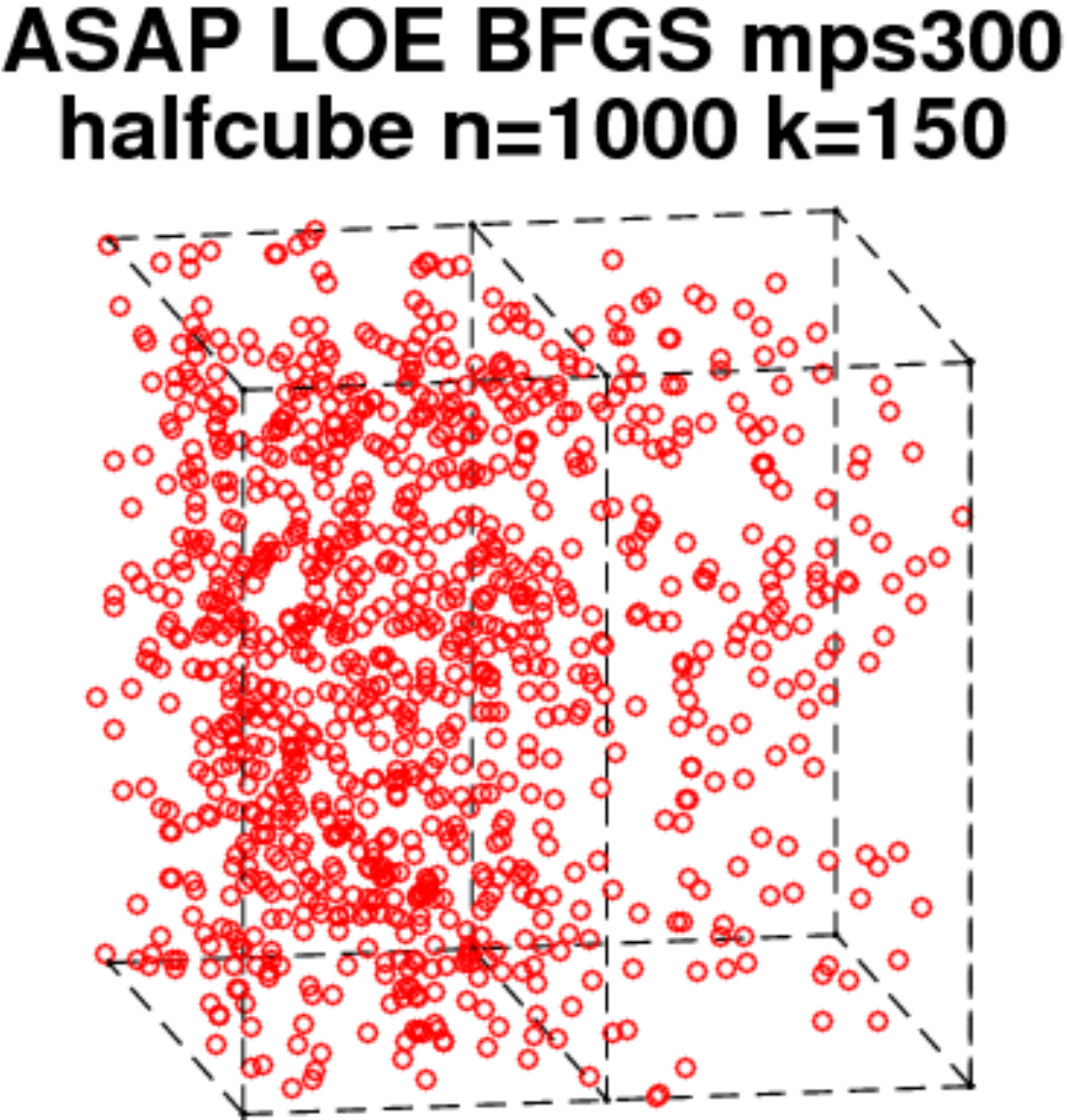}
  \includegraphics[width=\thirdcolwidth]{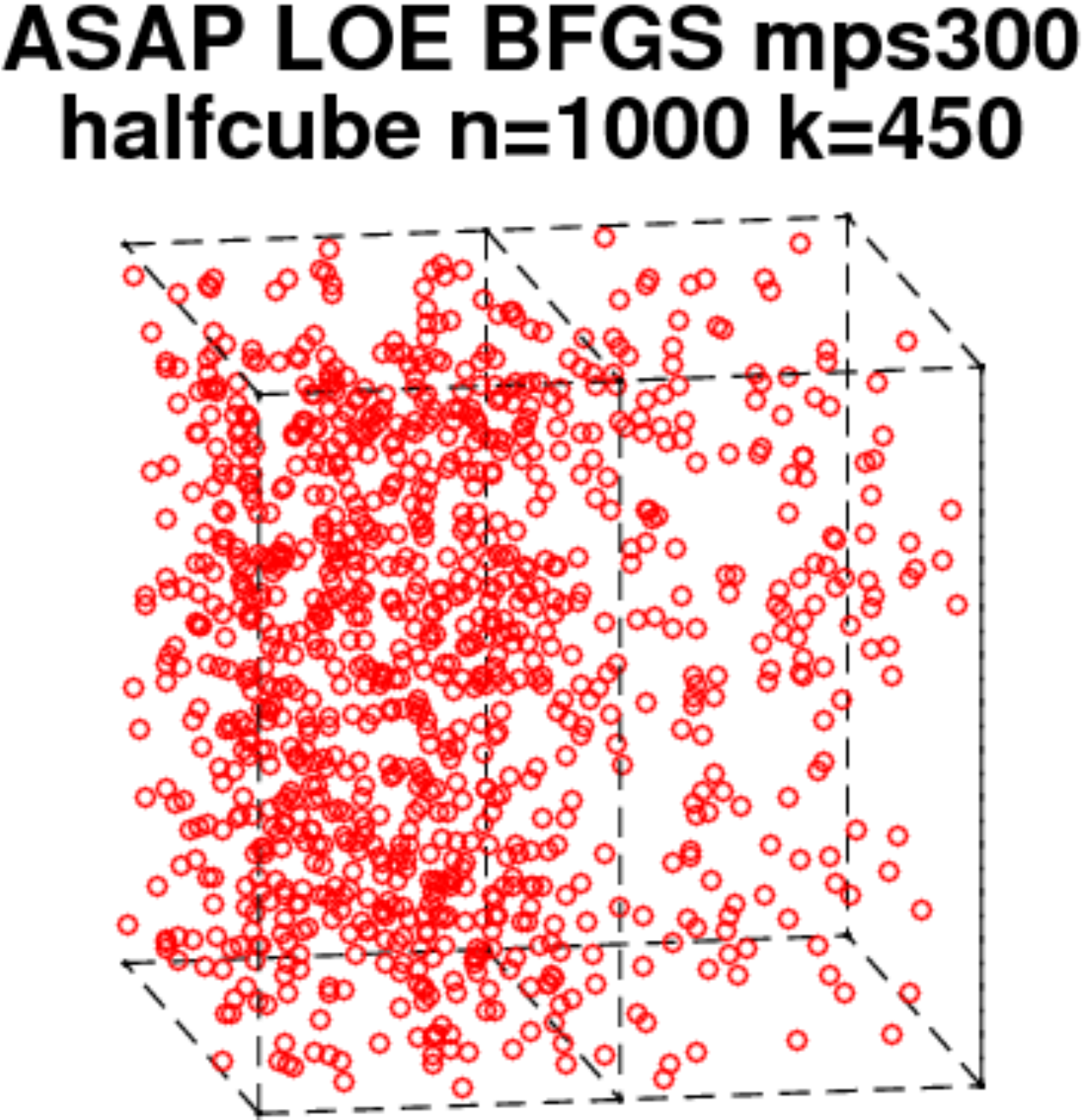}\\
  \includegraphics[width=\thirdcolwidth]{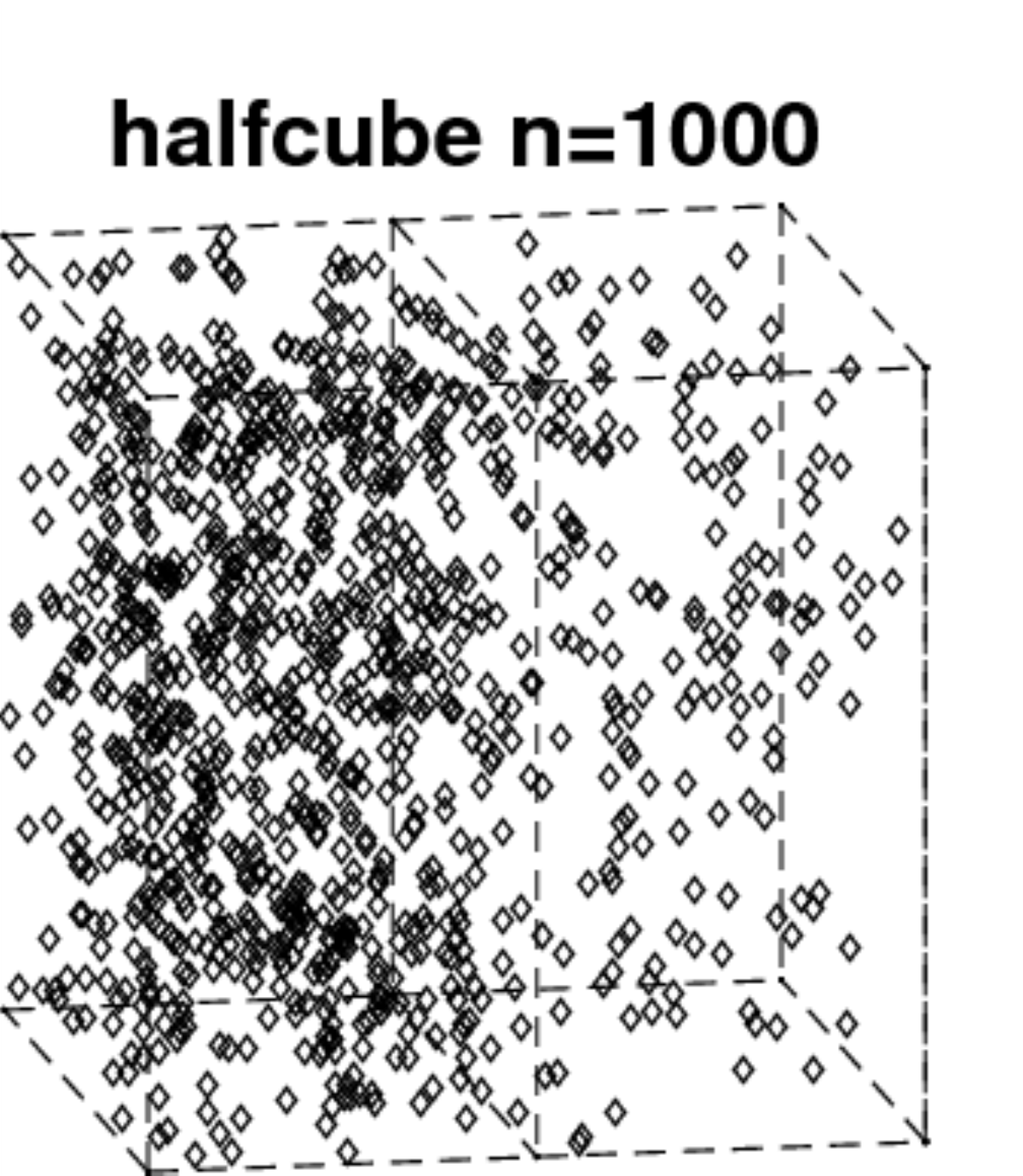}
\vspace{-3mm}
\caption{Embeddings for halfcube data sets with $n=1000$, and $k = 50$ (left), 150 (middle), 450 (right)
Row 2: LOE BFGS Iter.=100.
Row 3: ASAP LOE with MPS = 300 (with each ASAP result obtained is less time than the corresponding LOE result).
Row 4: ground truth.}
\vspace{-3mm}
\label{fig:halfcubeX}
\end{figure}

\begin{figure}[h]
\center
\begin{tabular}{c}
\includegraphics[width=\thirdcolwidth]{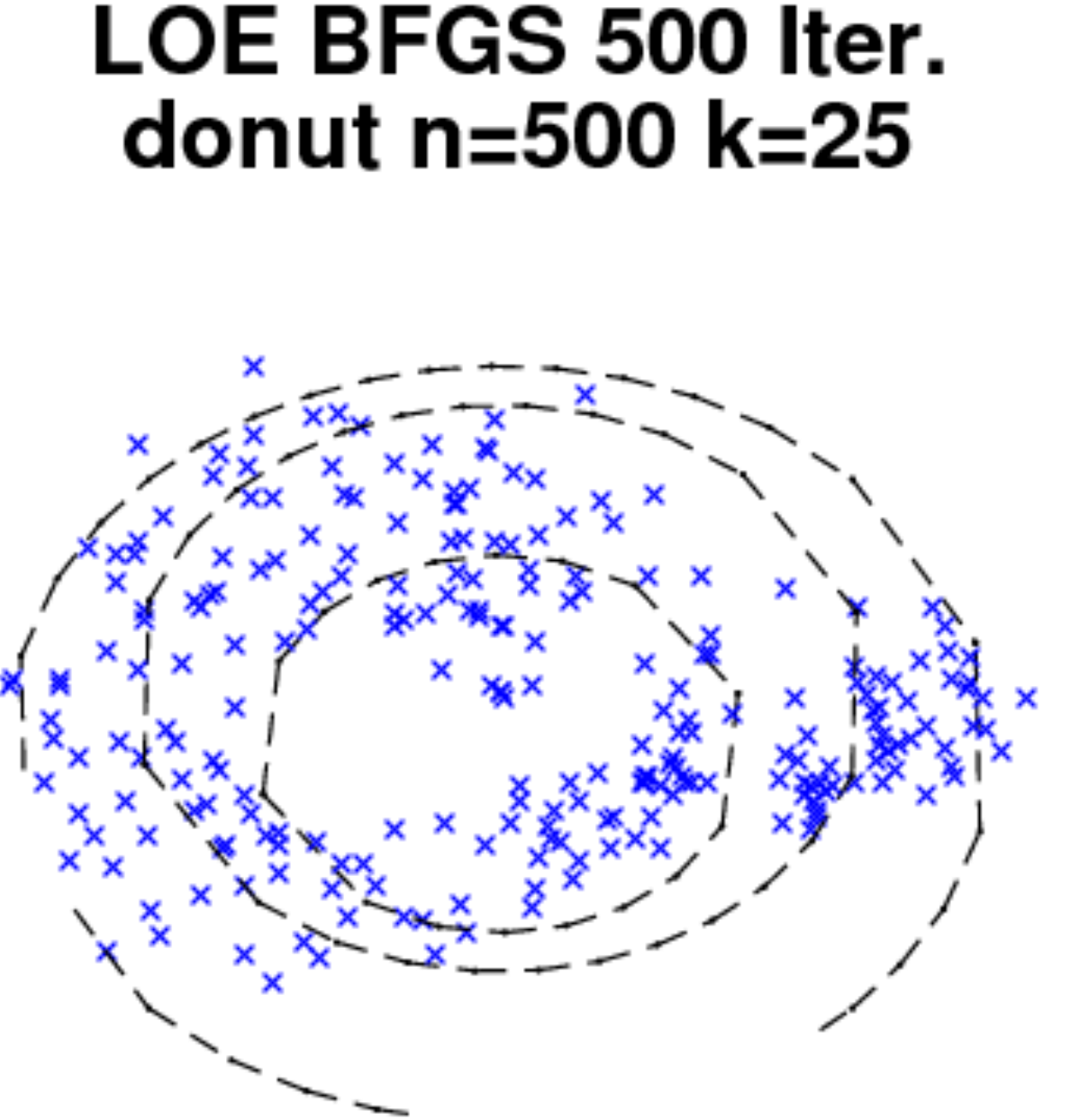}
\includegraphics[width=\thirdcolwidth]{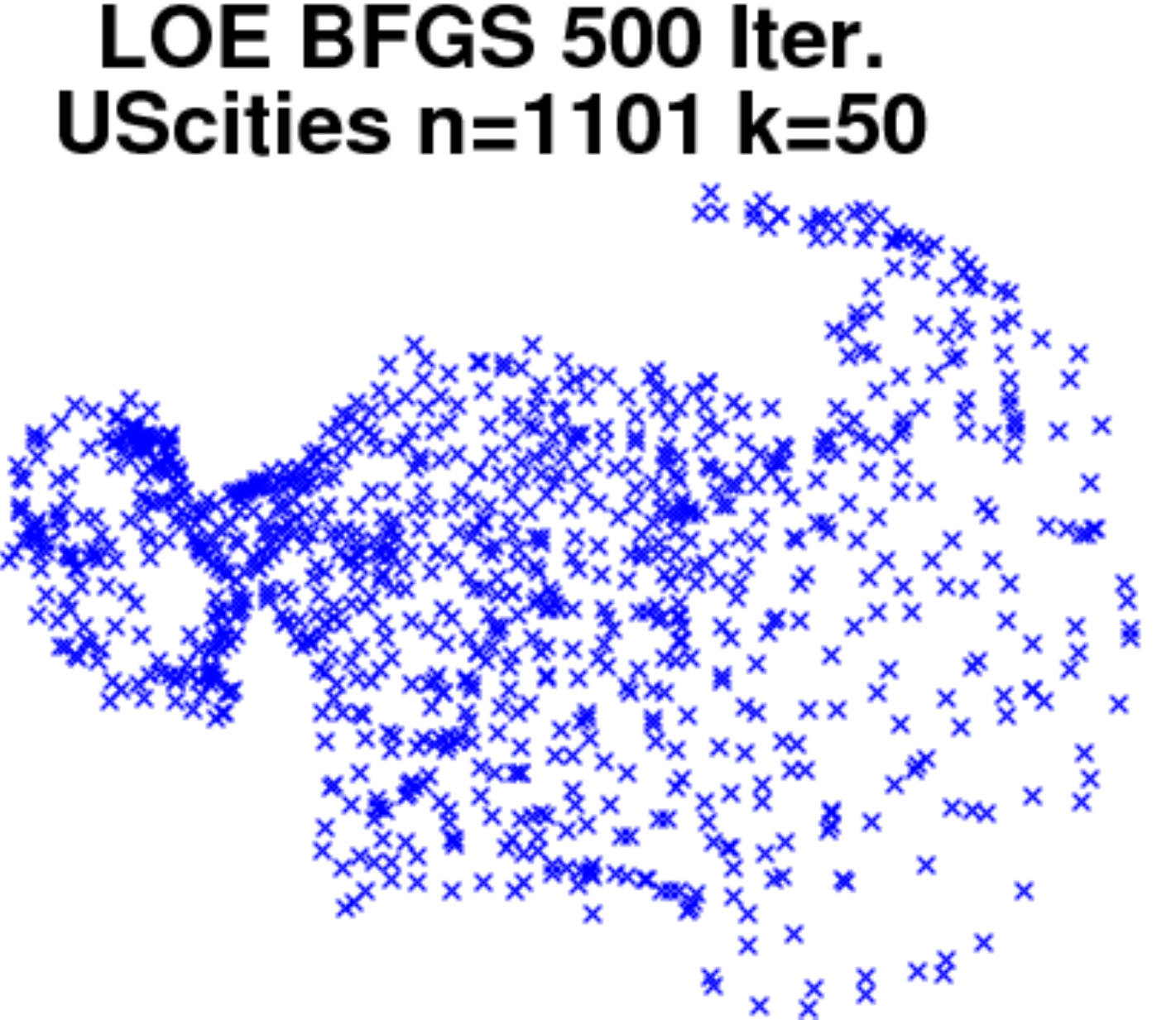}
\includegraphics[width=\thirdcolwidth]{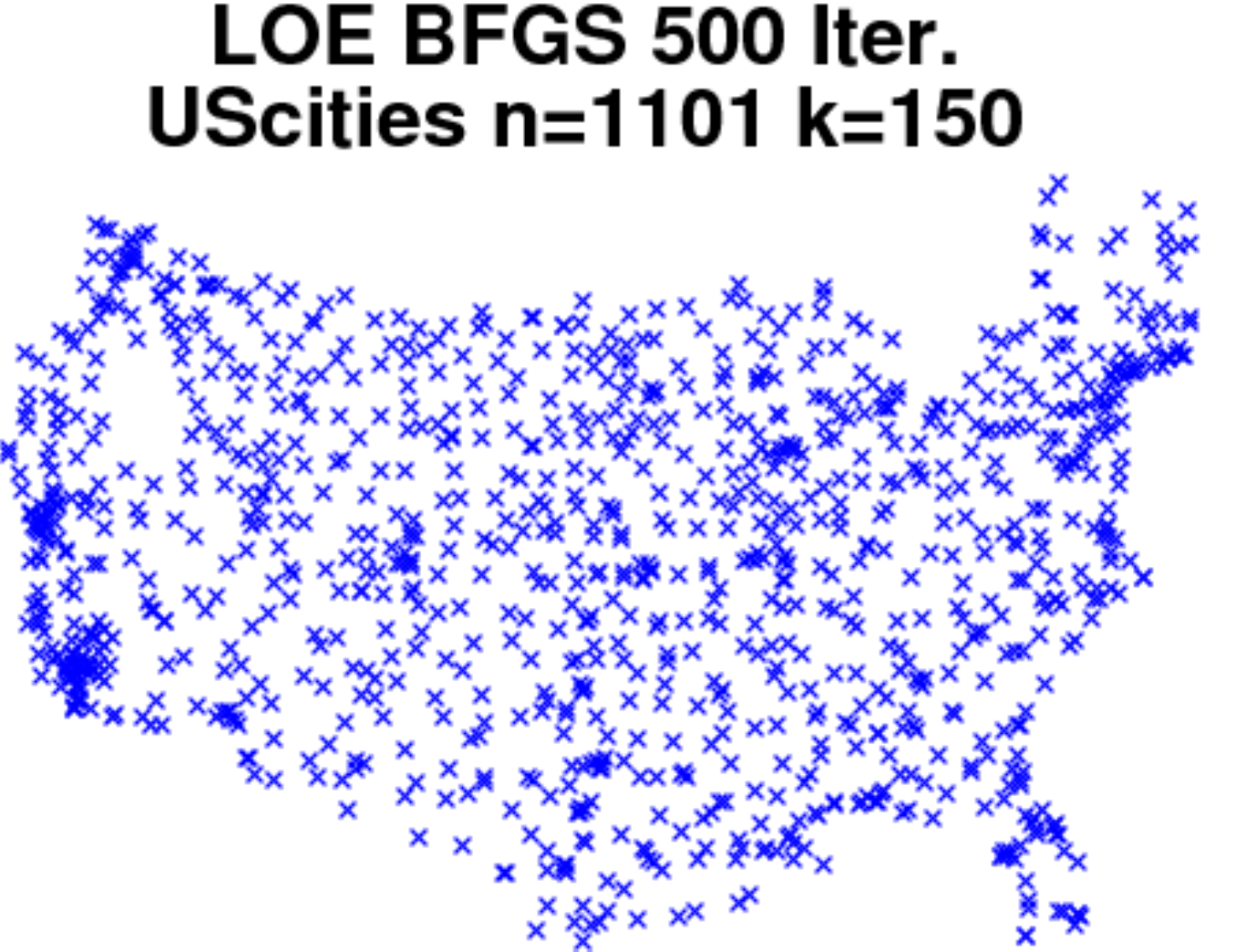}\\
\includegraphics[width=\thirdcolwidth]{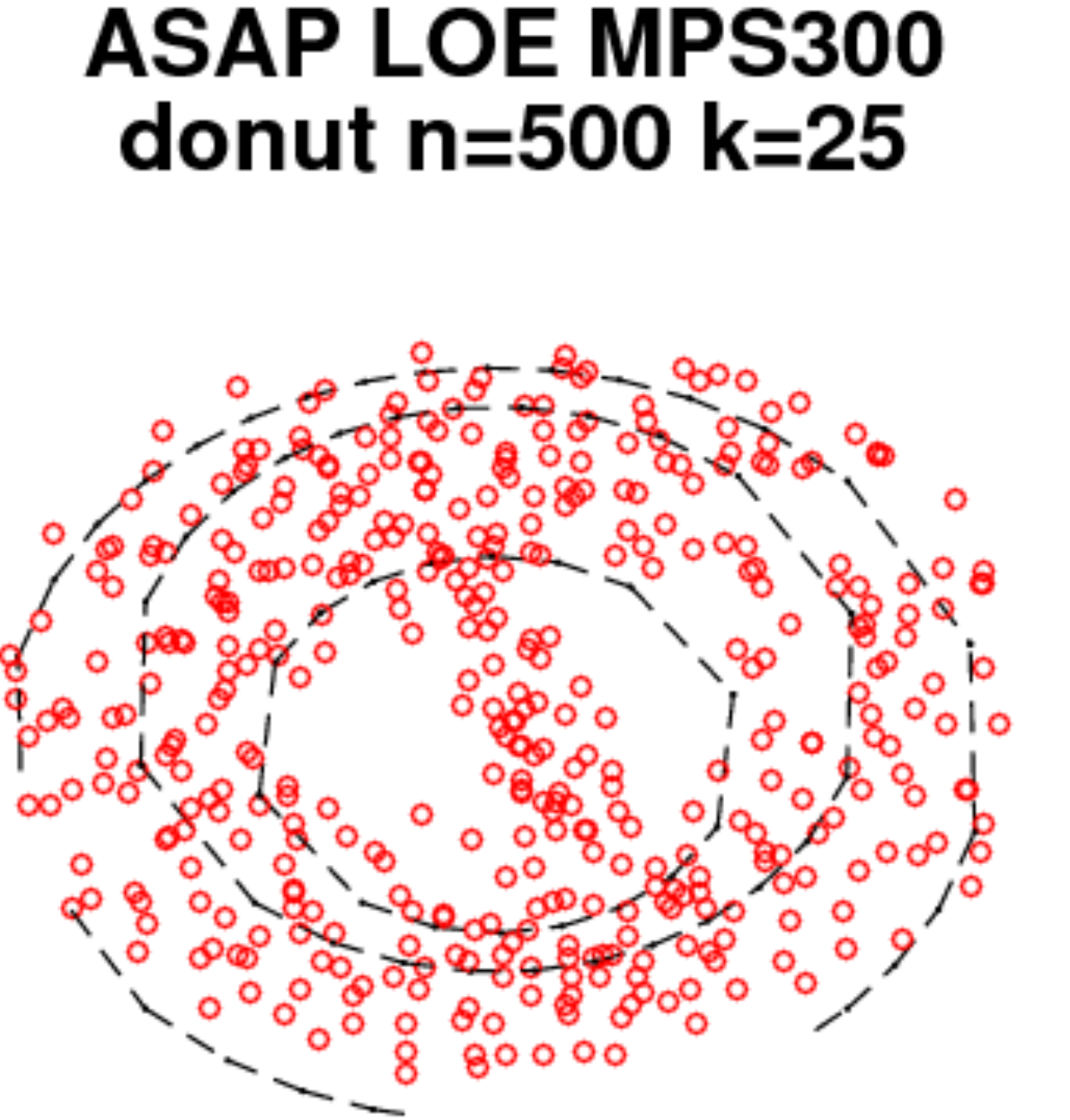}
\includegraphics[width=\thirdcolwidth]{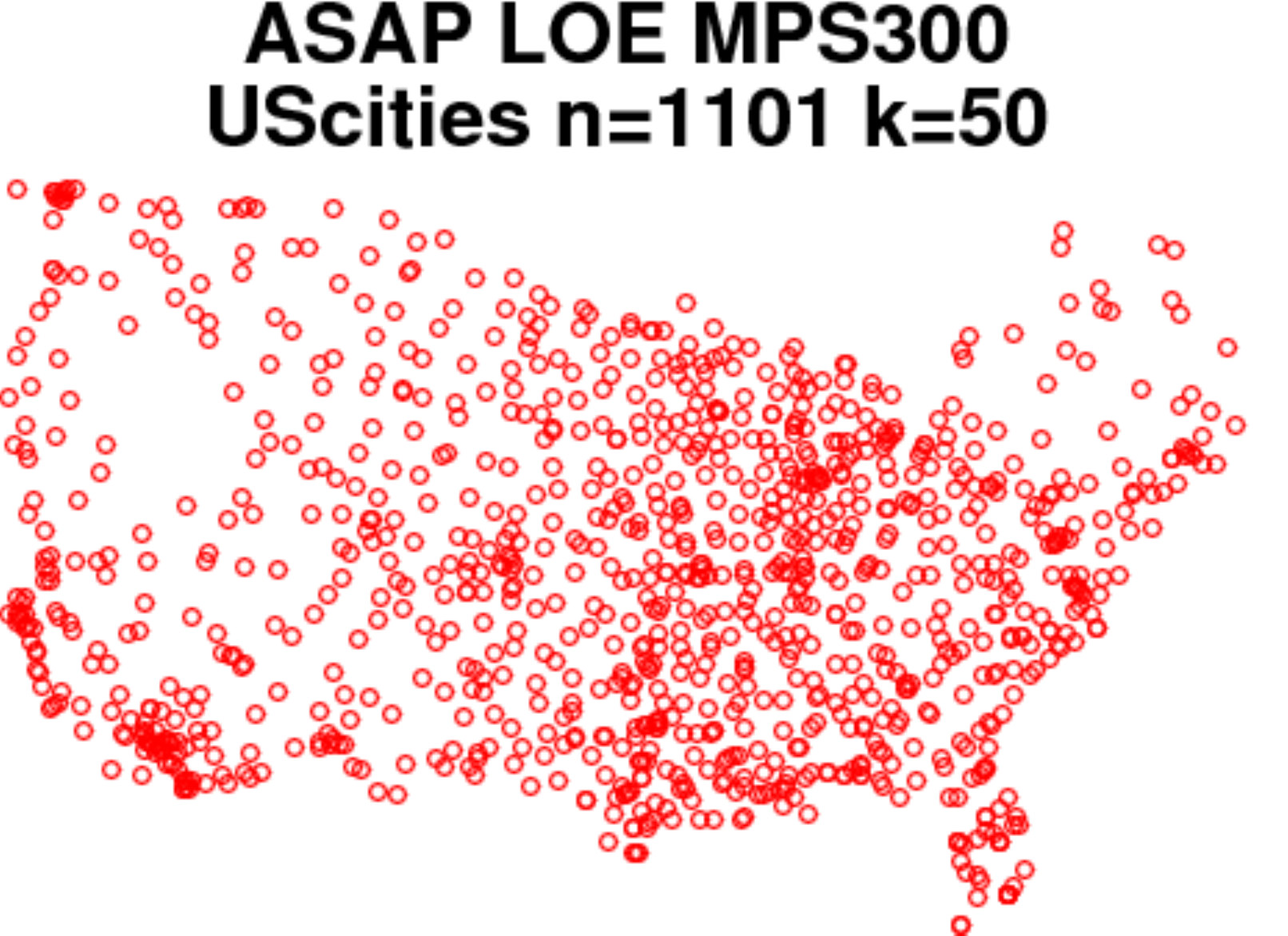}
\includegraphics[width=\thirdcolwidth]{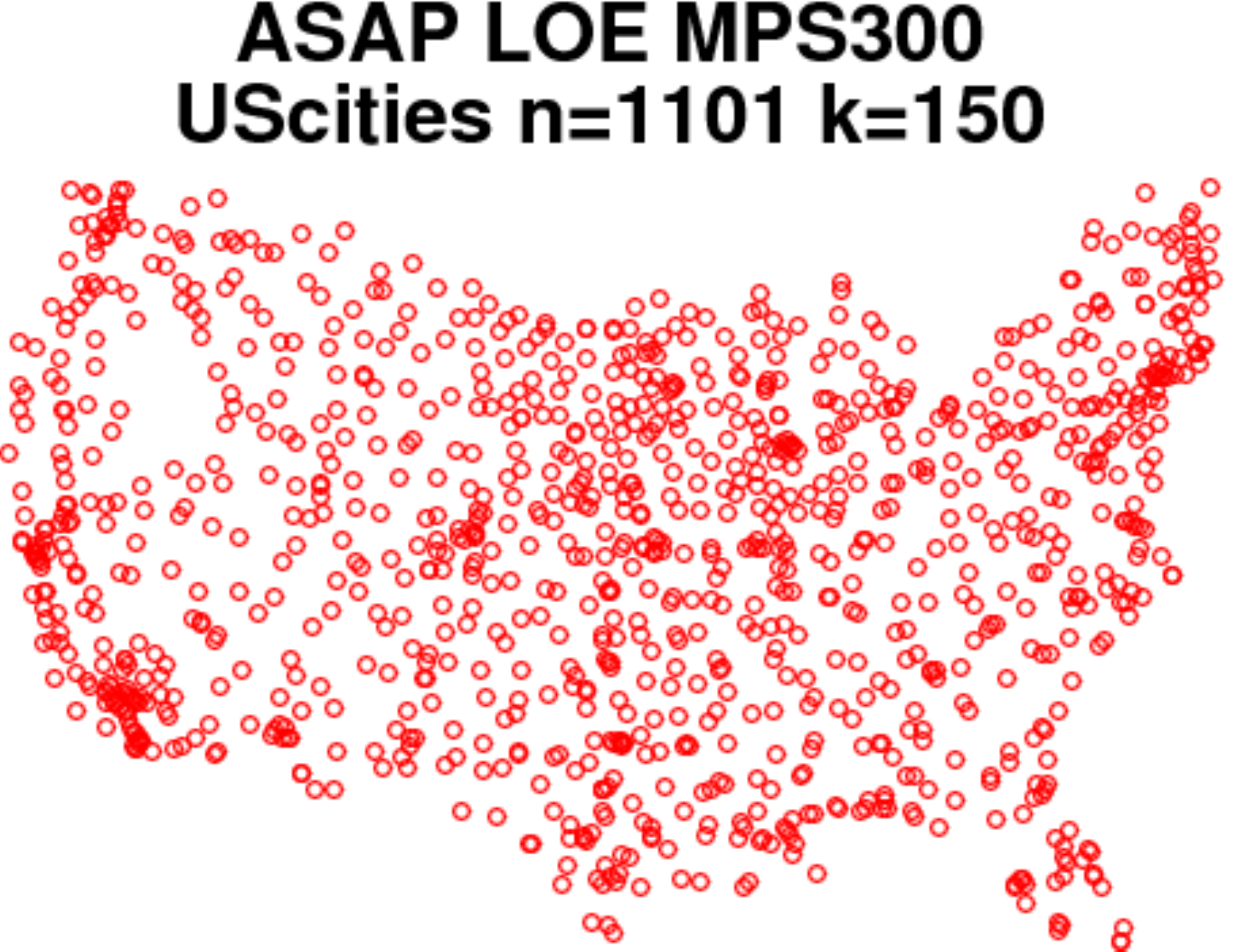}\\
\includegraphics[width=0.3\columnwidth]{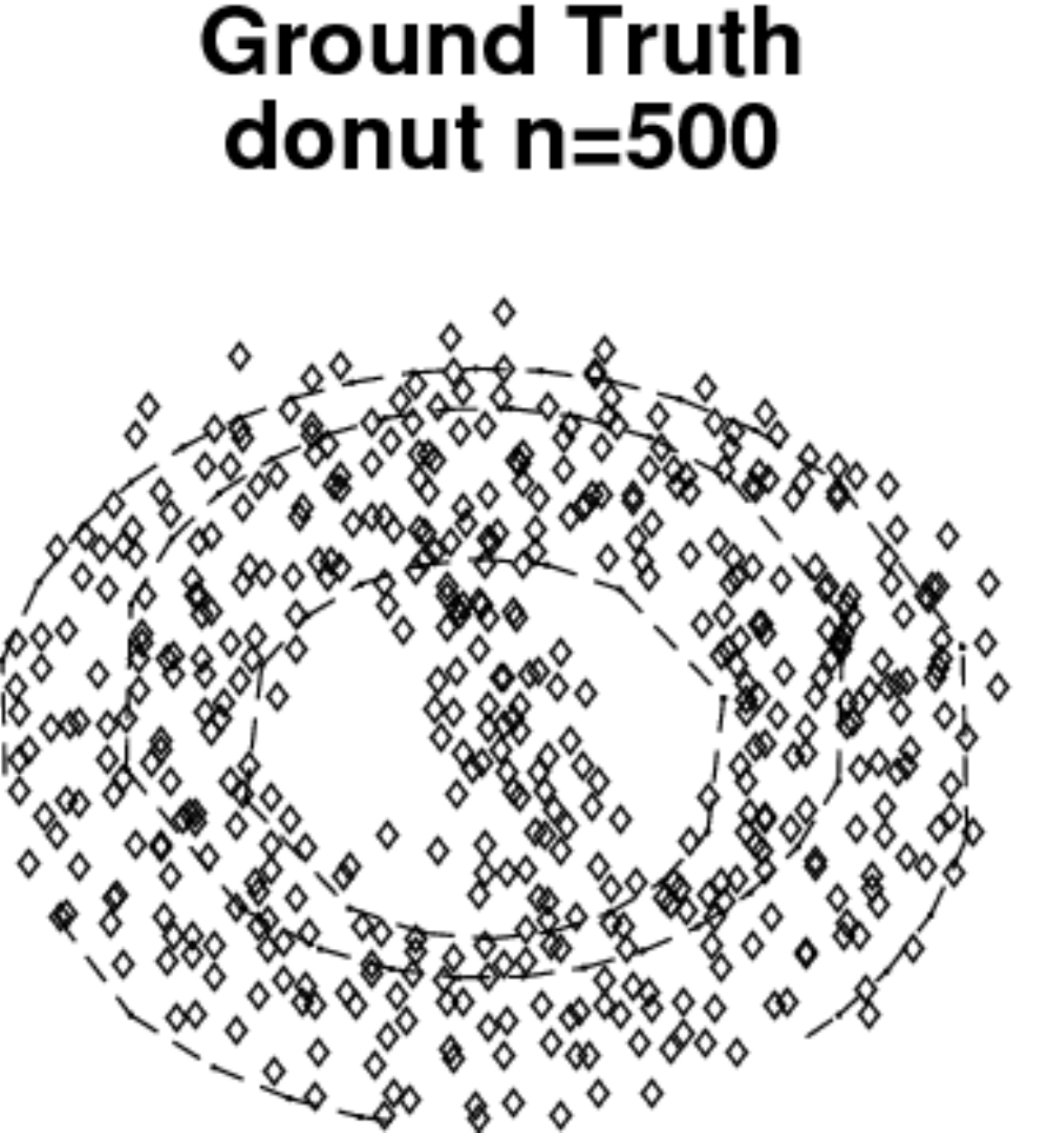}
\includegraphics[width=0.40\columnwidth]{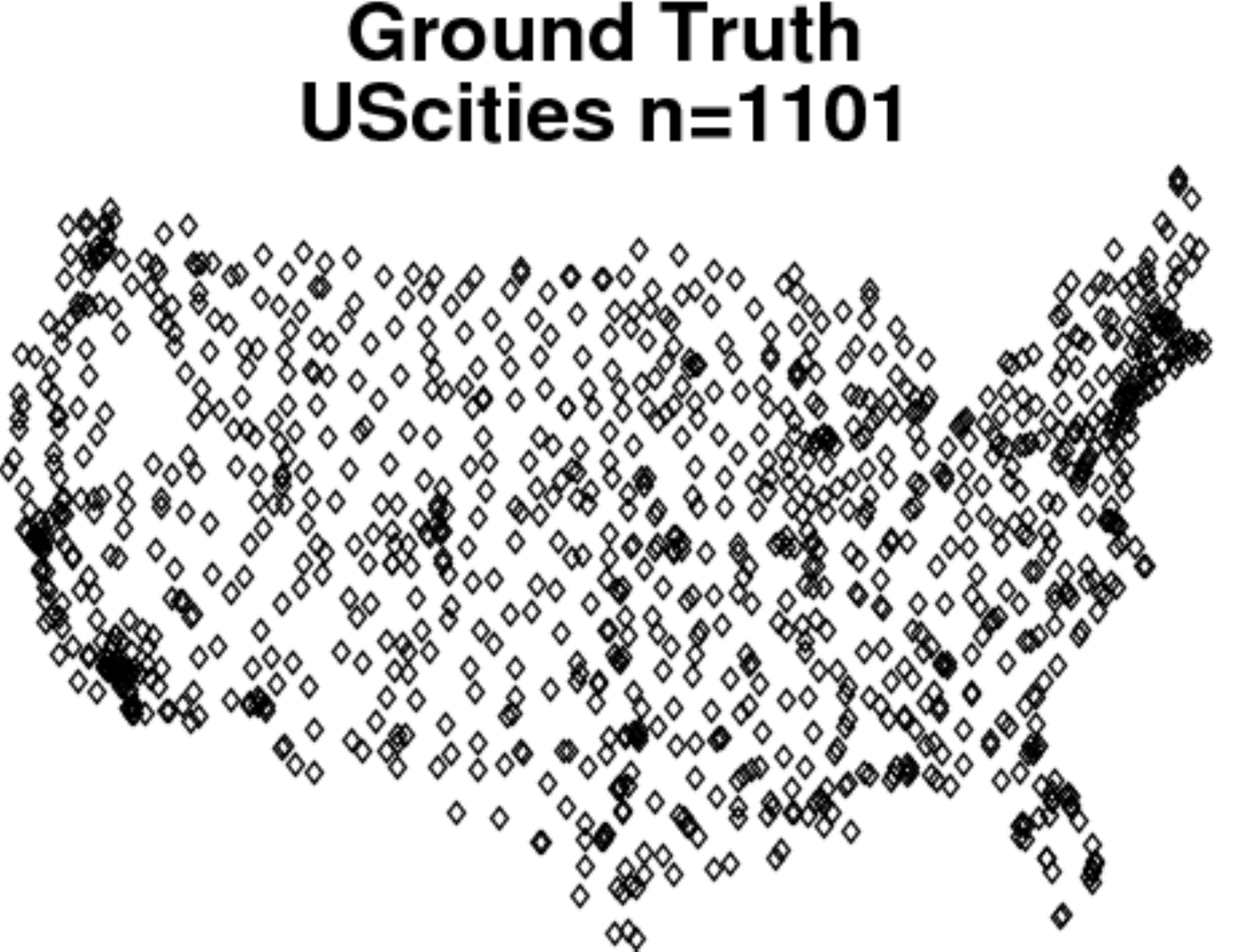}\\
\end{tabular}
\vspace{-3mm}
\caption{Embeddings of Donut (3D) and US Cities (2D) data sets. Row 1: LOE BFGS Iter.=500. Row 2: ASAP LOE MPS=300 (with each ASAP result obtained in less time than the corresponding LOE result). Row 3: Ground truth.}
\vspace{-3mm}
\label{fig:DonutAndUS}
\end{figure}

To demonstrate that this approach is not limited to the 2D case, nor does it only perform well on synthetic data, we plot in Figure~\ref{fig:DonutAndUS}  the  embeddings 
Procrustes aligned with 
points sampled from a 3D donut shape, and actual coordinates of $n=1101$ US cities. In both cases, ASAP LOE with MPS=300 runs faster and yields smaller $\mathcal{E}_A$ 
than LOE BFGS with 500 maximum iterations, the latter of which produces  twisted or folded results.

\subsection{Large $n$ : 50,000}

In Table~\ref{fig:n50000} we show $\mathcal{E}_A$  vs runtime
for ASAP LOE  on a data set of $n=50,000$ points and $k=22$. While this size is completely prohibitive for LOE BFGS, ASAP LOE produces 
good results in less than 4 hours.  The worst possible result would be all edges of original graph misplaced, meaning $\mathcal{E}_A = 2\cdot 50k\cdot 22 / (50k)^2 = 8.8\times10^{-4}$.  $\mathcal{E}_A = 2 \times 10^{-4}$ means we get approximately $3/4$ of the edges correct.  

\begin{table}
\center
\begin{tabular}{|c|c|c|c|}
\hline
MPS & 100  & 300  & 500 \\
\hline
PCS $\mathcal{E}_A $ & $5.1\times10^{-4}$ &  $5.6\times10^{-4}$ &   $1.9\times10^{-4}$ \\
PC $\mathcal{E}_A $ & $5.8\times10^{-4}$ &  $4.7\times10^{-4}$ &   $3.0\times10^{-4}$ \\ 
\hline
\end{tabular}
\vspace{-2mm}
\caption{Recovery results for $n=50,000$ for ASAP LOE.}
\vspace{-2mm}
\label{fig:n50000}
\end{table}

\subsection{Increasing $k$}
\begin{figure}[h]
\center
\begin{tabular}{c}
\includegraphics[width=\halfcolwidth]{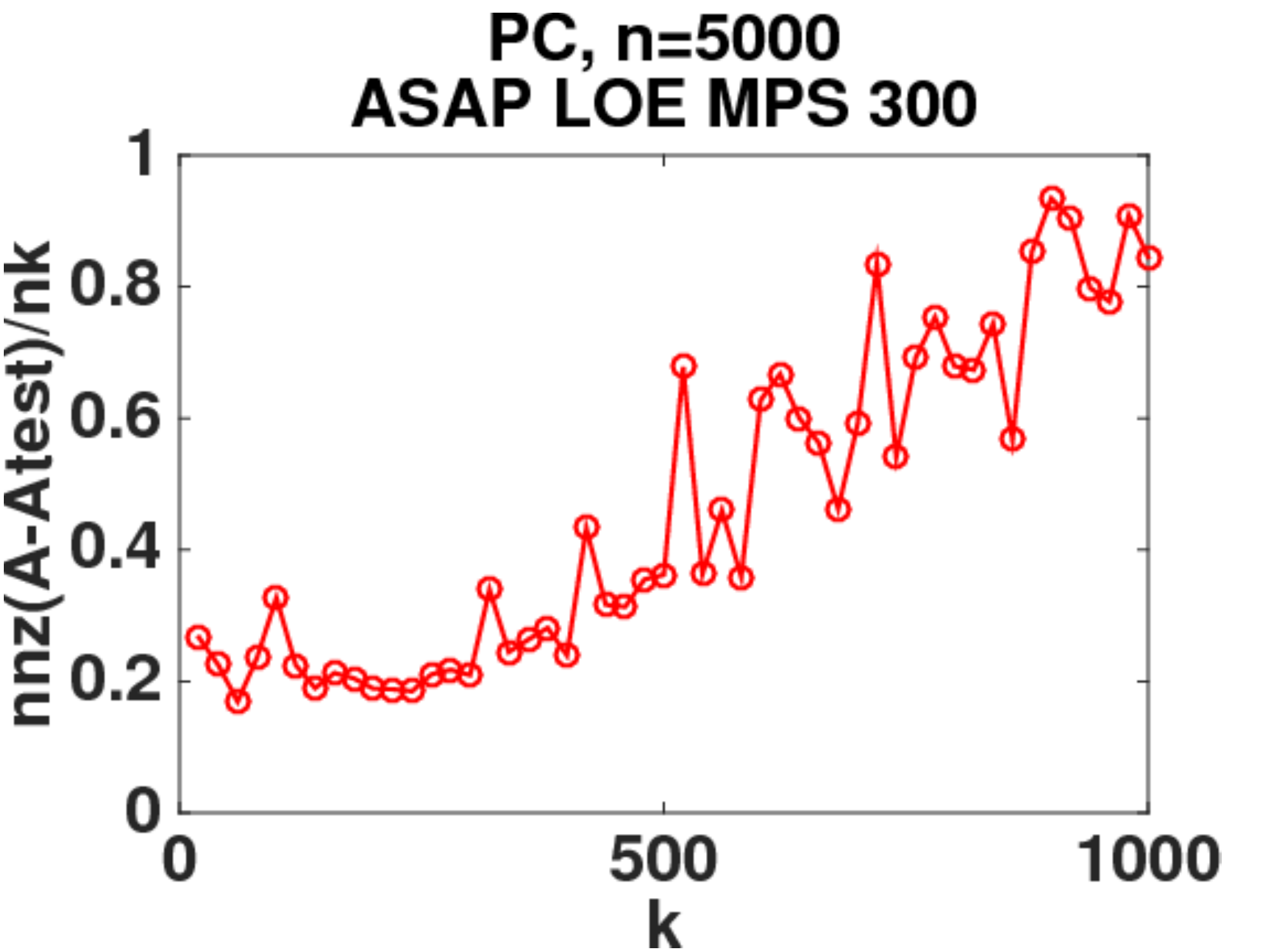}
\includegraphics[width=\halfcolwidth]{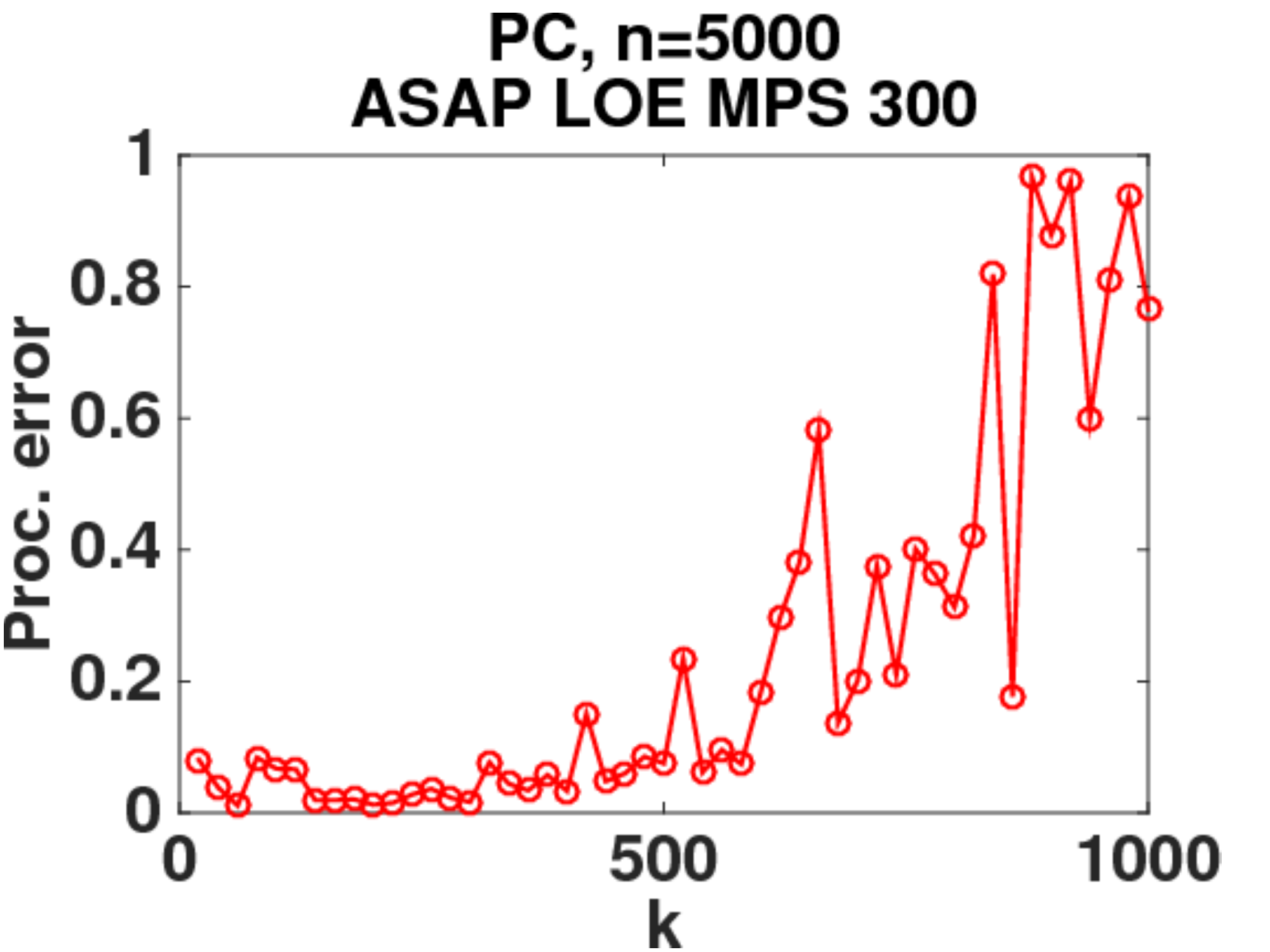}
\end{tabular}
\caption{ASAP LOE MPS=300, $n=5000$, $k$ increasing by 20, Left: number of differences in adjacency matrix divided by number of edges, $nk$,
Right: Procrustes error.}
\label{fig:incrk}
\end{figure}

\begin{figure}[h]
\center
\begin{tabular}{c}
\includegraphics[width=\halfcolwidth]{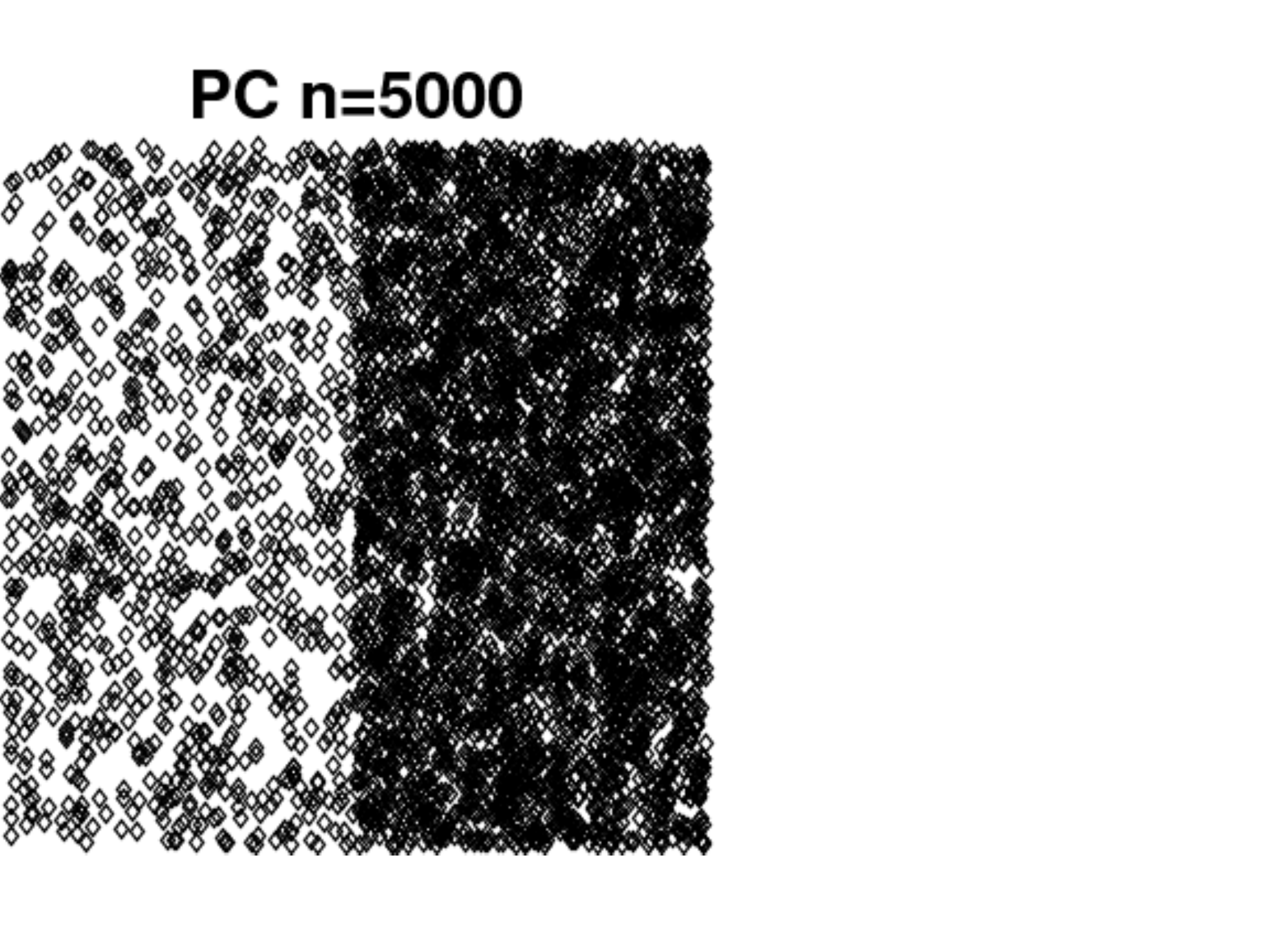}
\includegraphics[width=\halfcolwidth]{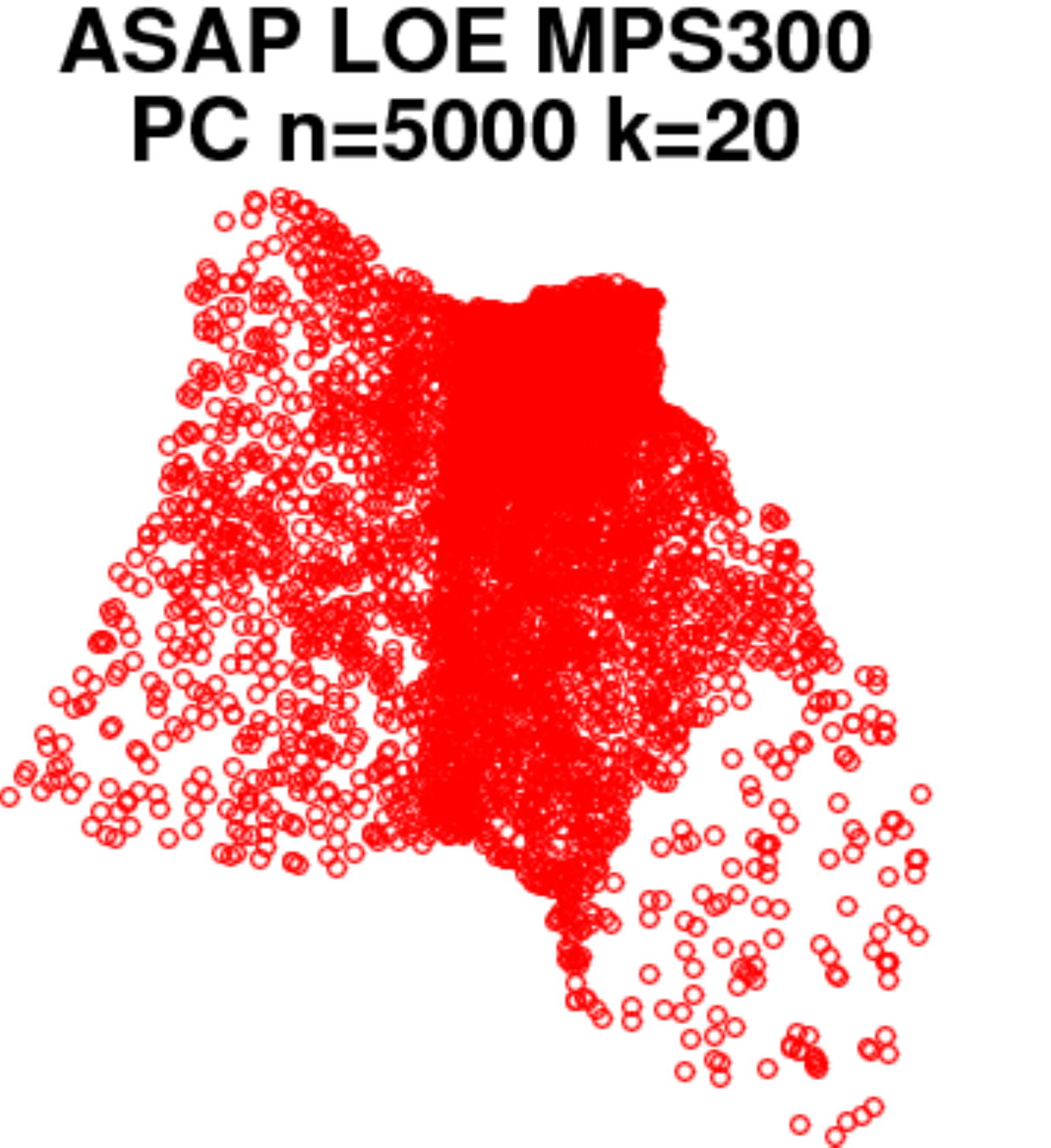}\\
\includegraphics[width=\halfcolwidth]{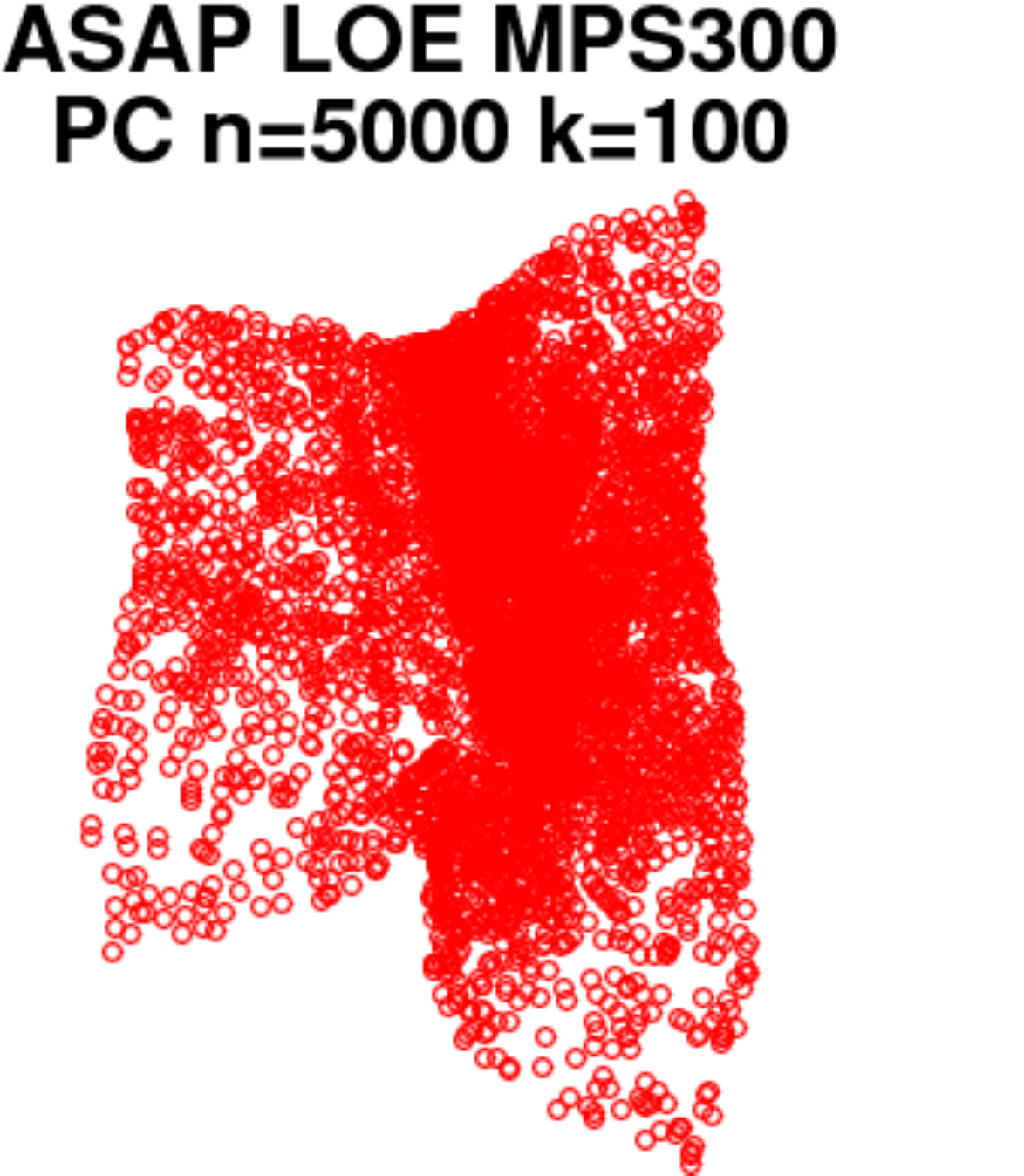}
\includegraphics[width=\halfcolwidth]{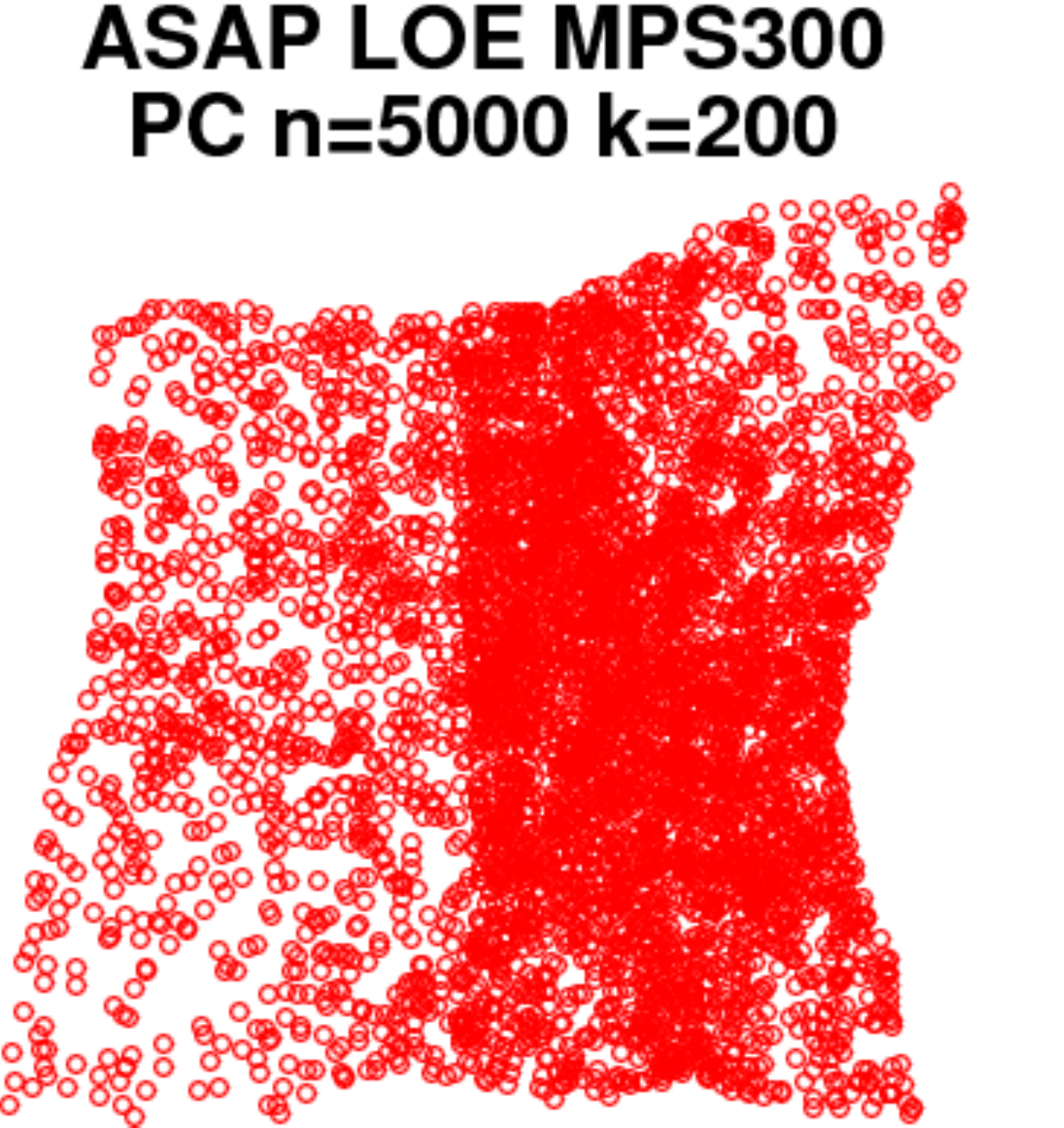}\\
\includegraphics[width=\halfcolwidth]{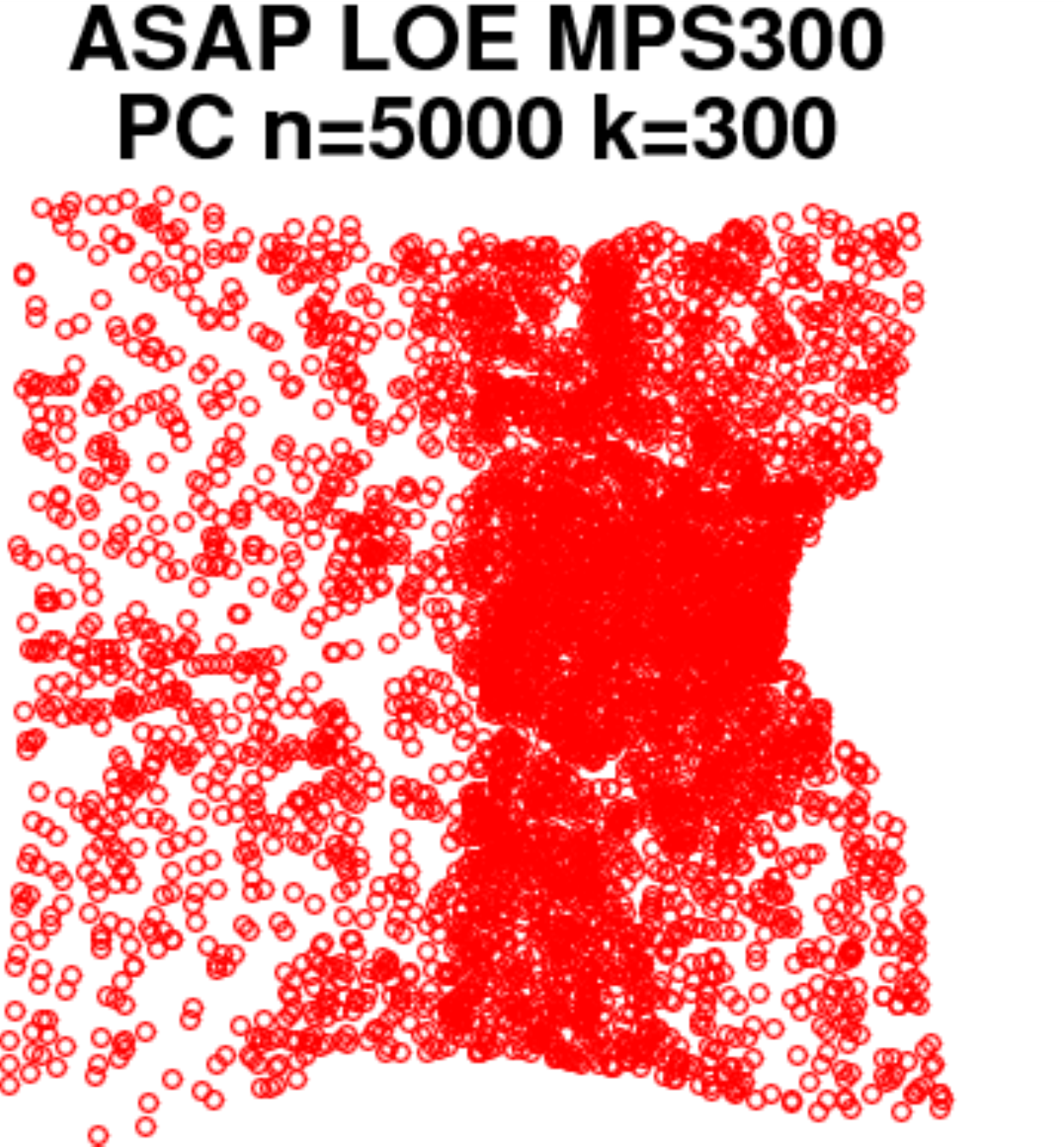}
\includegraphics[width=\halfcolwidth]{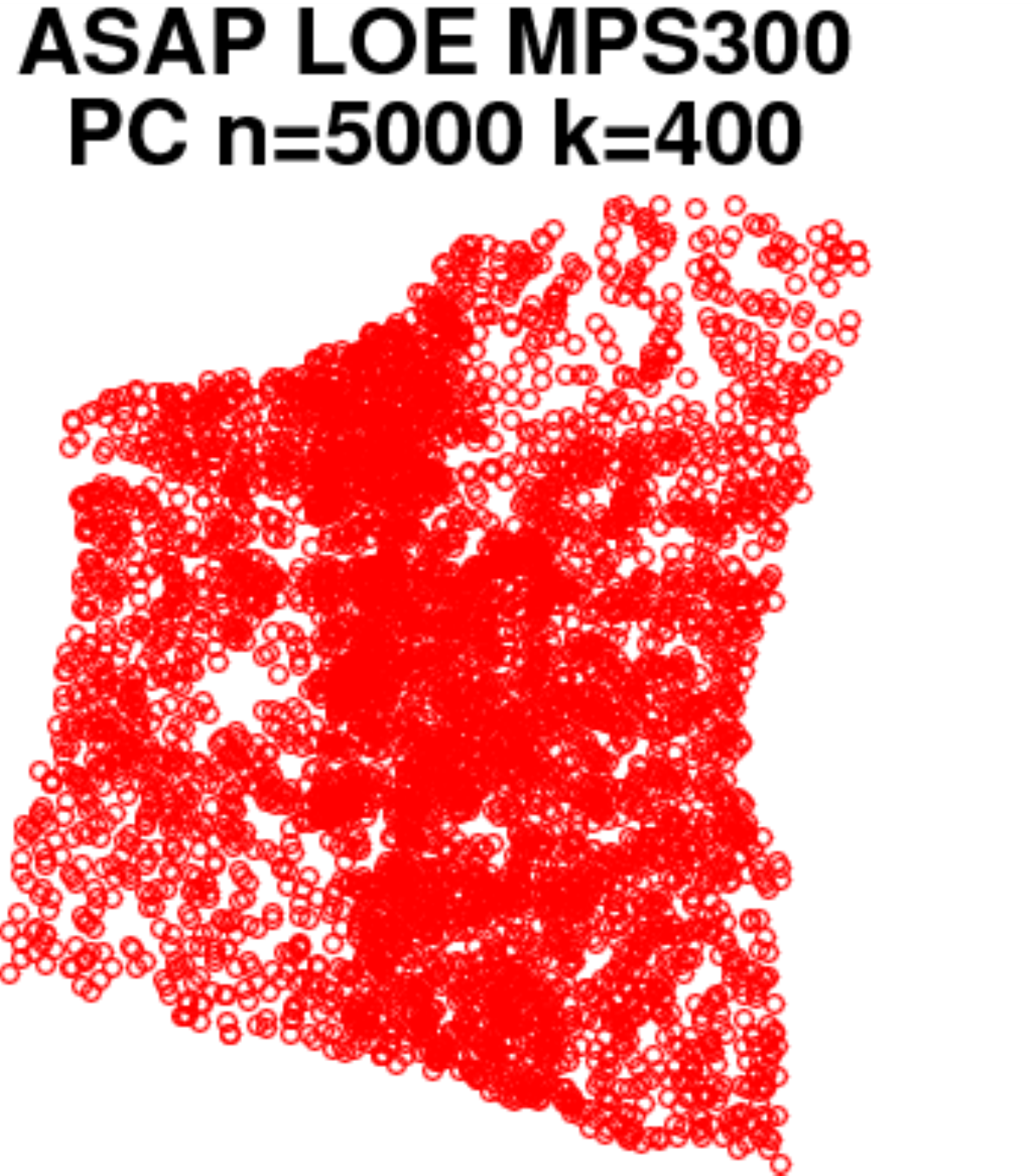}\\
\includegraphics[width=\halfcolwidth]{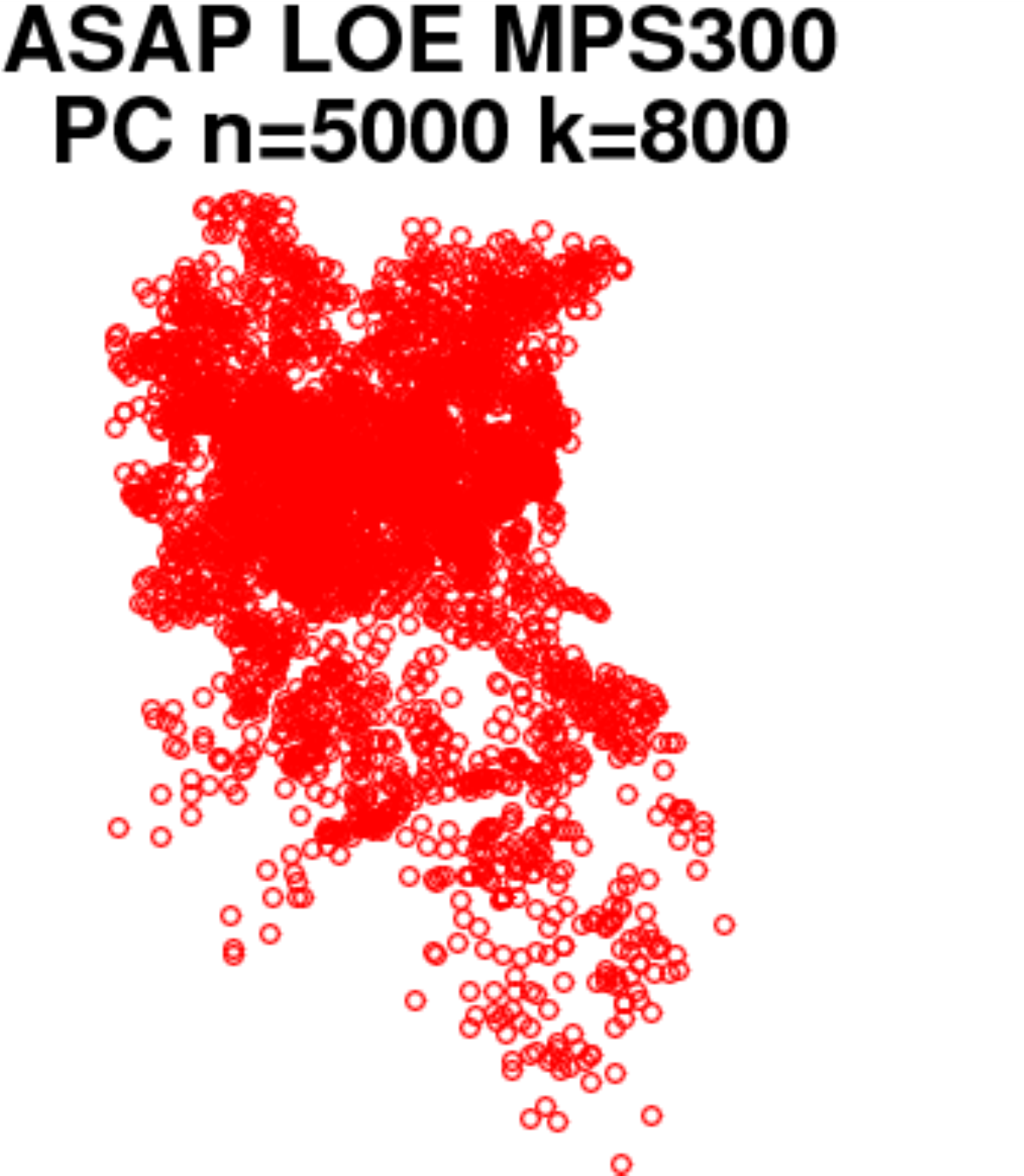}
\includegraphics[width=\halfcolwidth]{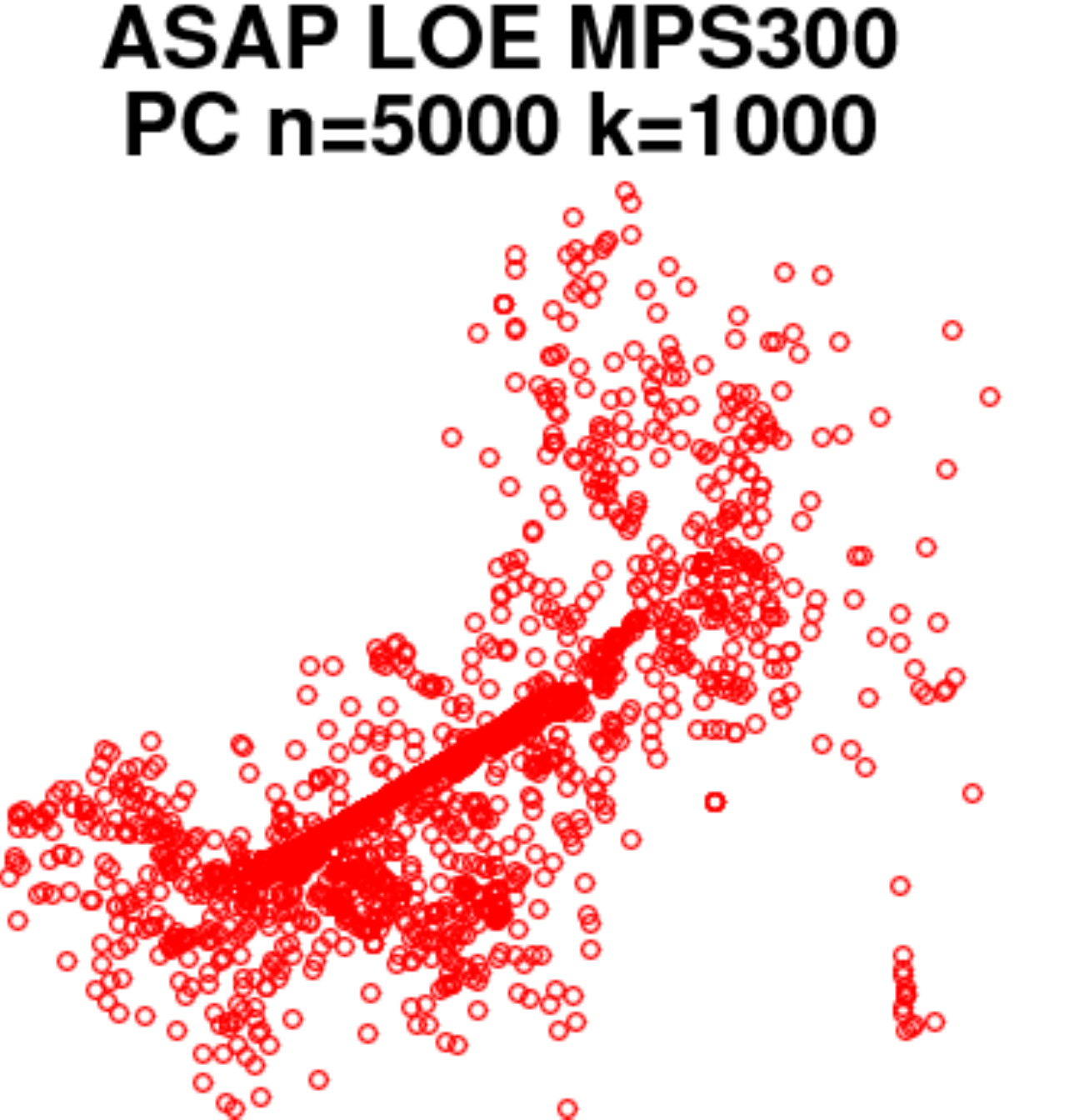}
\end{tabular}
\caption{ASAP LOE BFGS MPS=300, $n=5000$, $k$ increasing by 20, Top left : originally sampled points, Remaining plots : recovered embeddings}
\label{fig:incrkX}
\end{figure}

We show in Figure~\ref{fig:incrk}, scaled $(n/k)\times \text{NumberNonzero}(A-A_0)$ (this is a rescaling of $\mathcal{E}_A $ proportional to number of misplaced edges, more comparable for different values of $k$) and procrustes error versus increasing values of $k$ for $n=\{5000\}$ points drawn from the piecewise constant half-planes distribution using the method ASAP LOE with MPS=300. We see that for large $n$, adjacency matrix error and Procrustes error remain relatively small and stable over a range of small increasing $k$.  Additionally, we show in Figure~\ref{fig:incrkX} some of the embeddings corresponding to these results.  Like the Procrustes error plot, these embeddings suggest that for a range of $k$ small relative to $n$ and not too large relative to MPS, ASAP LOE BFGS returns sensible, although not perfect results.  As $k$ gets too large however the results are quite poor.  We suspect this is a result of $k$ being too large relative to MPS, leading to patches which are overly dense.  When an ordinal graph contains nearly all possible edges, it essentially provides no information.  When such data is of specific interest, one could either increase the mps as computational resources and time allow, or potentially use an alternate method for breaking the graph into overlapping patches which are not too dense.

\subsection{Density Estimation Experiments}

\begin{figure}[H]
\center
\begin{tabular}{c}
\includegraphics[width=\halfcolwidth]{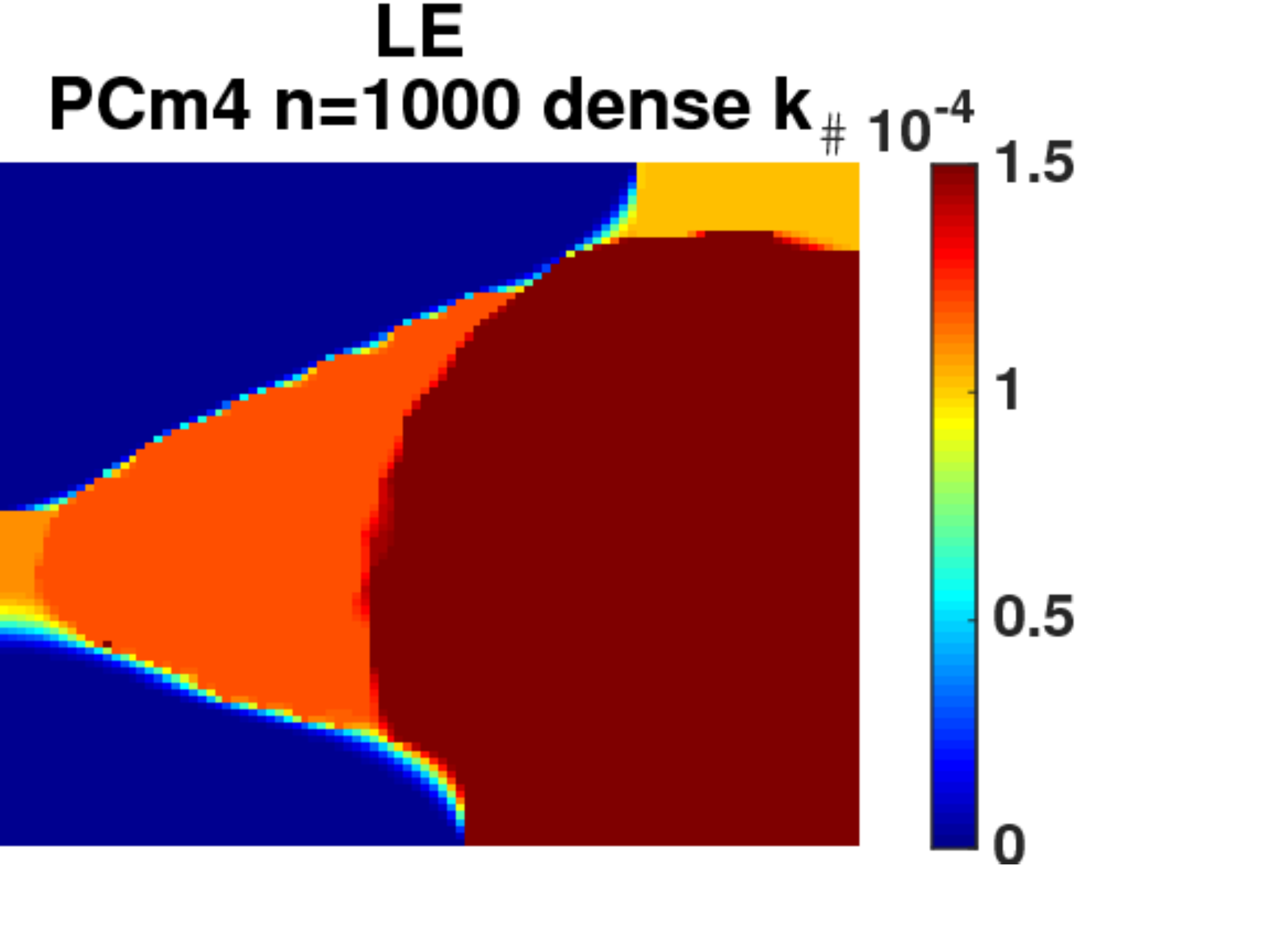} 
\includegraphics[width=\halfcolwidth]{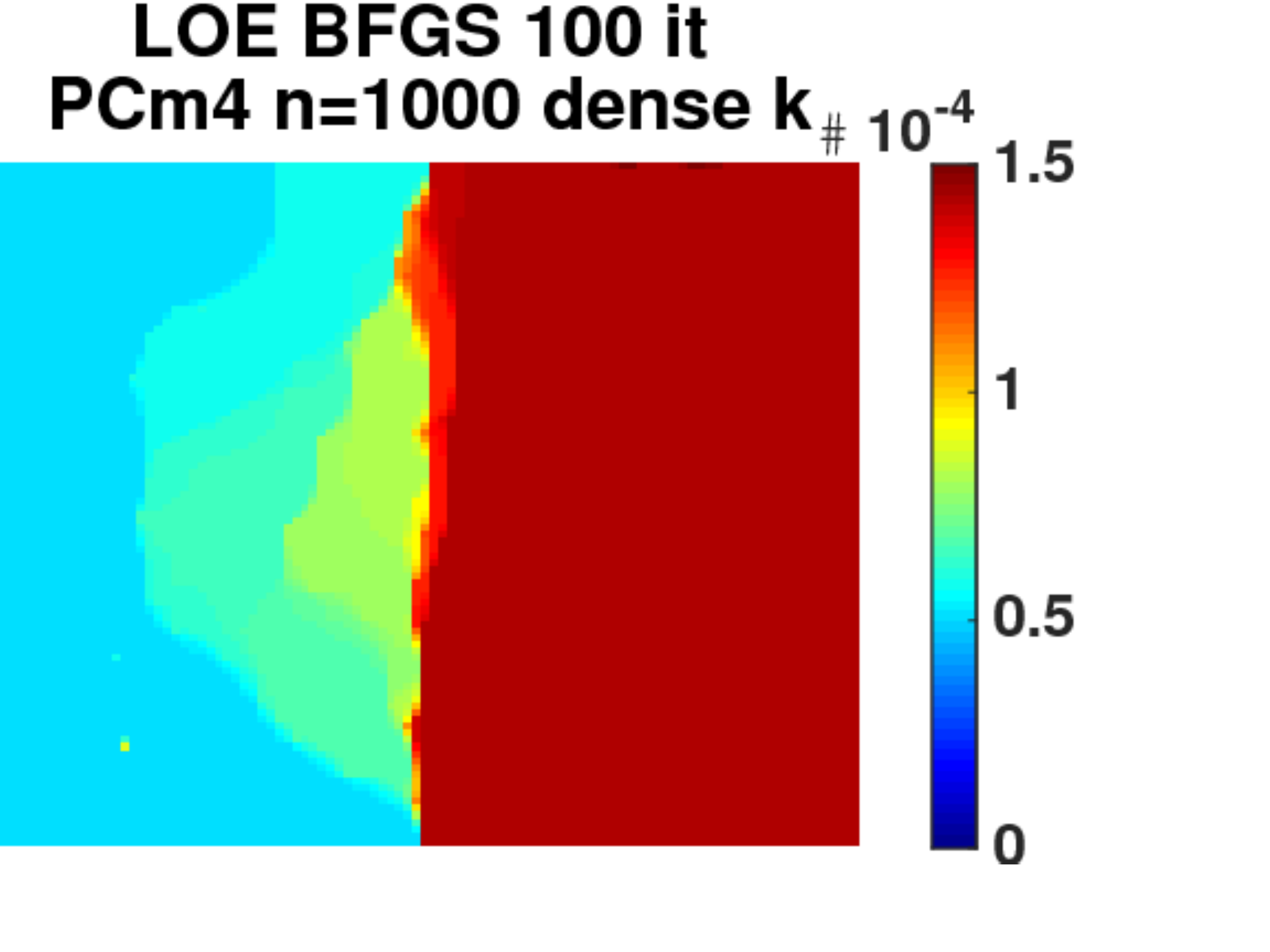}\\
\includegraphics[width=\halfcolwidth]{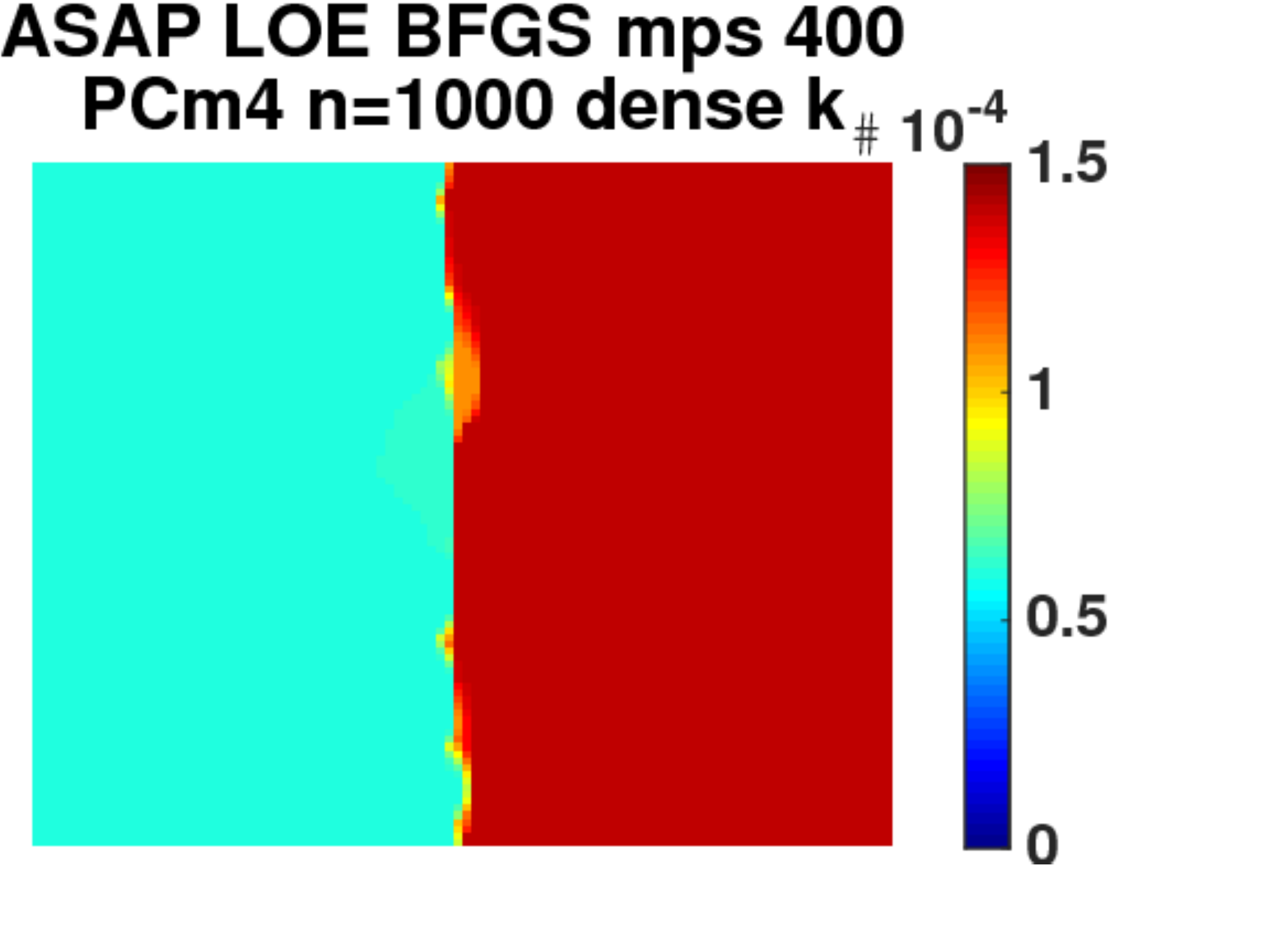}
\includegraphics[width=\halfcolwidth]{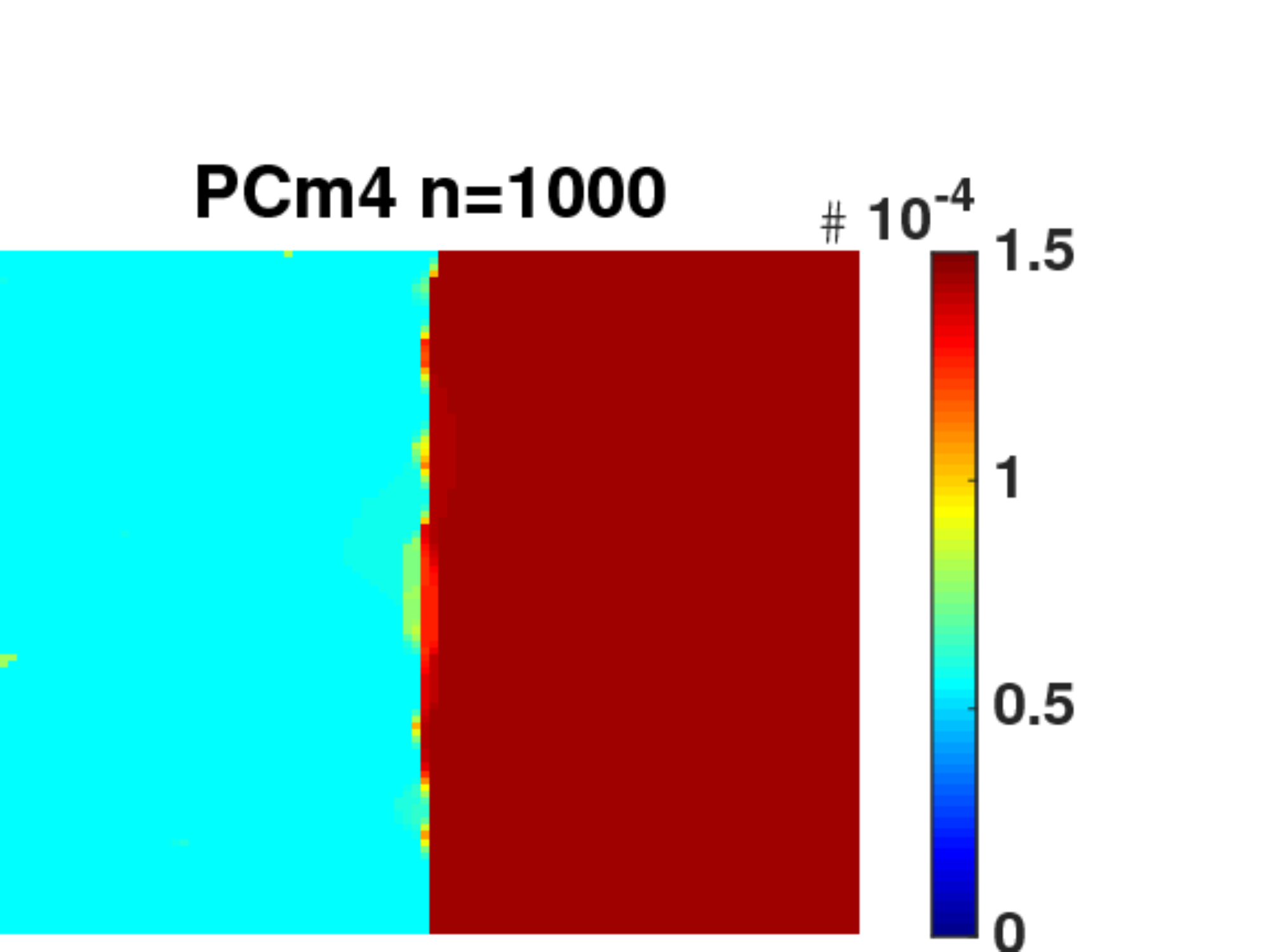}
\end{tabular}
\caption{TV MPLE applied to example embeddings of PC $n=1000$, $k$ dense, and top left : LE, top right : LOE BFGS maxIt=100, bottom left : ASAP LOE BFGS max patch size 400, bottom right : estimated density from ground truth points, see column 1 of Figure~\ref{fig:X}}
\label{fig:TVPC}
\end{figure}

In Figure~\ref{fig:TVPC} we show the results of applying TV MPLE to some of the embeddings shown in Figure~\ref{fig:X}.  The regularization parameter used is .0001 .  This is not obtained by cross-validation, but it simply seems to perform well on the originally sampled points.  The densities of the approximate embeddings are as expected, with ASAP LOE BFGS recovering the density best, with LOE BFGS behind, and LE doing the worst.  This altogether  suggests that better embedding results do lead to better density estimation, if that is the end goal.

\begin{figure}[H]
\center
\begin{tabular}{c}
\includegraphics[width=\halfcolwidth]{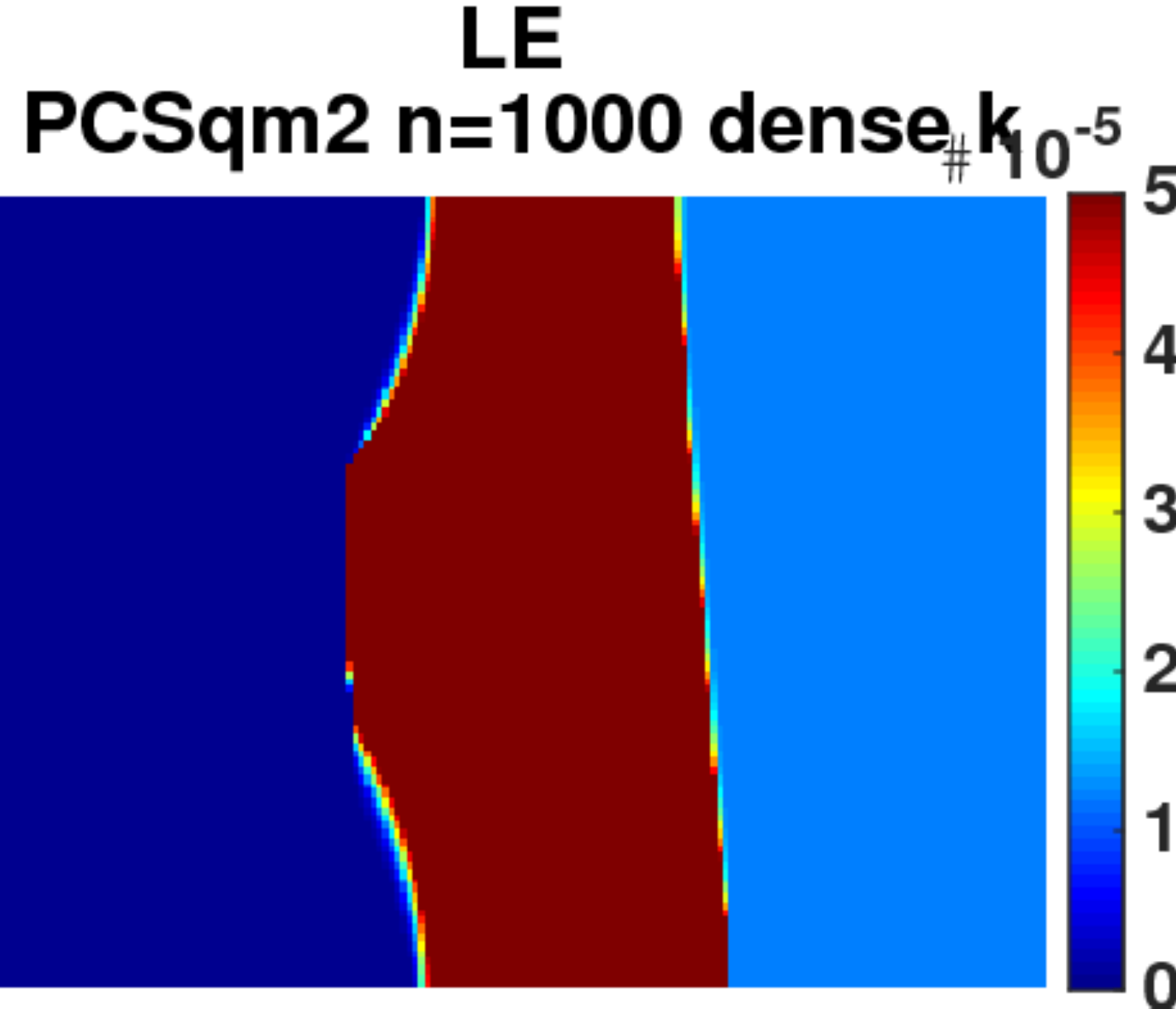} 
\includegraphics[width=\halfcolwidth]{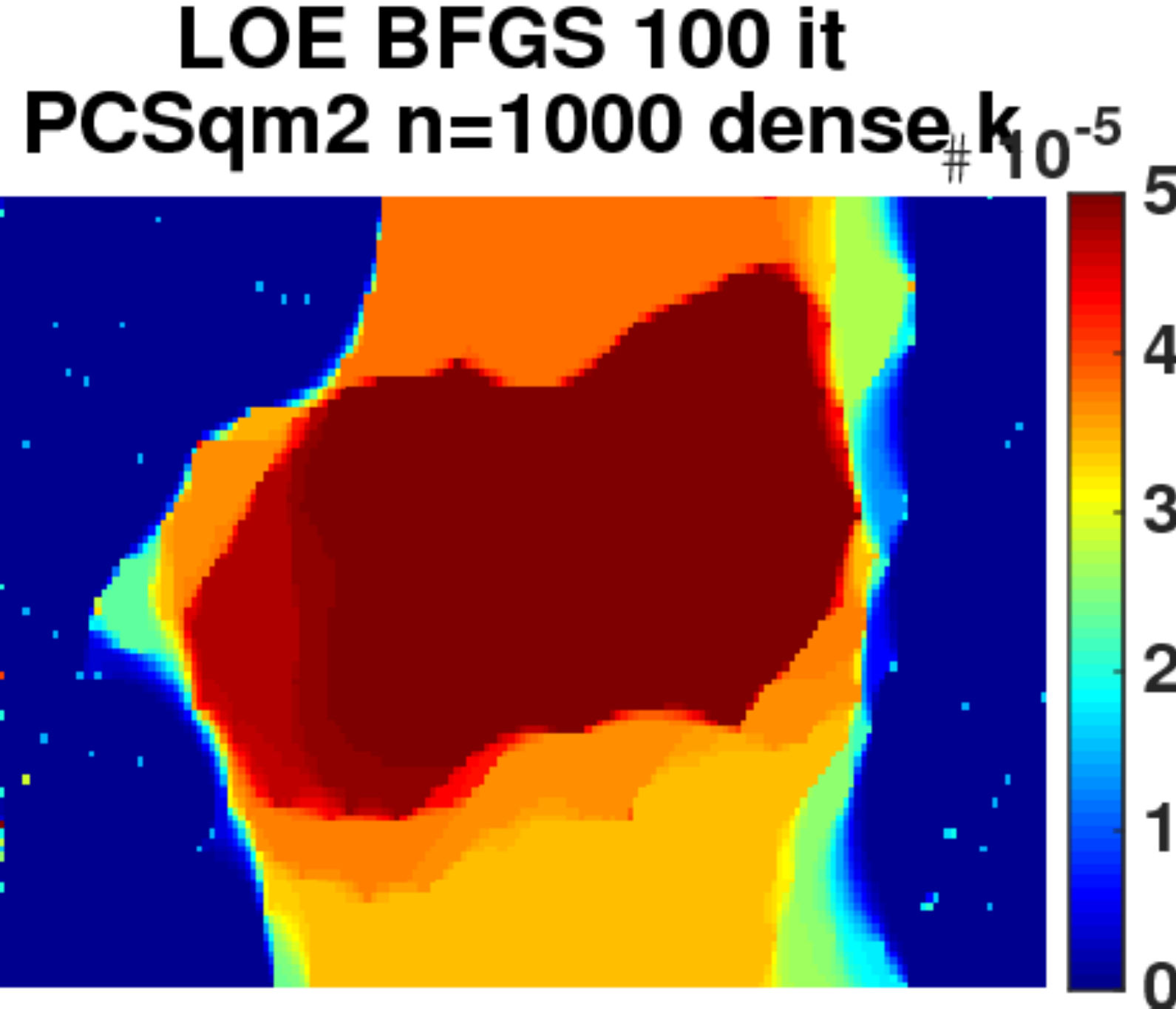}\\
\includegraphics[width=\halfcolwidth]{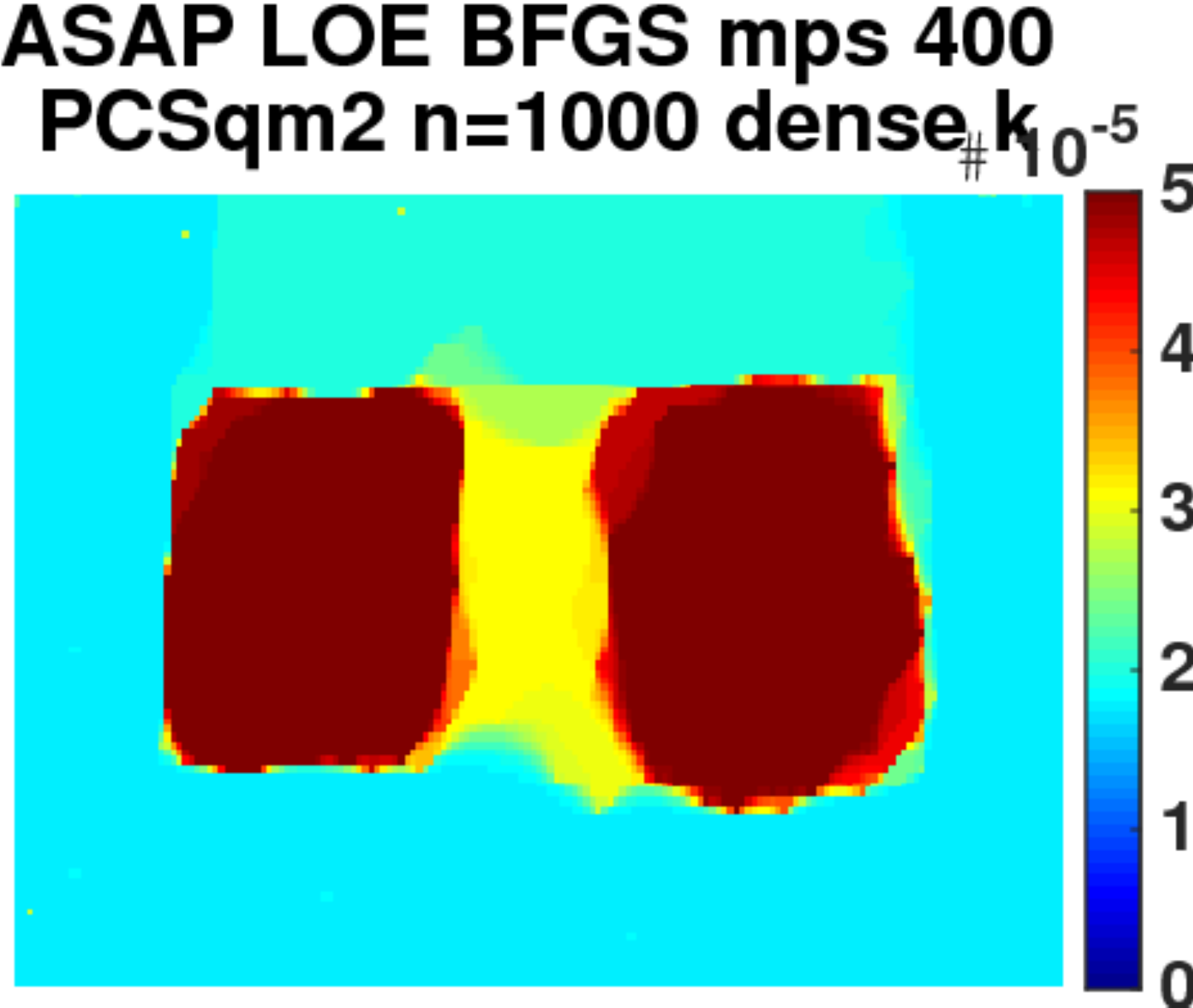}
\includegraphics[width=\halfcolwidth]{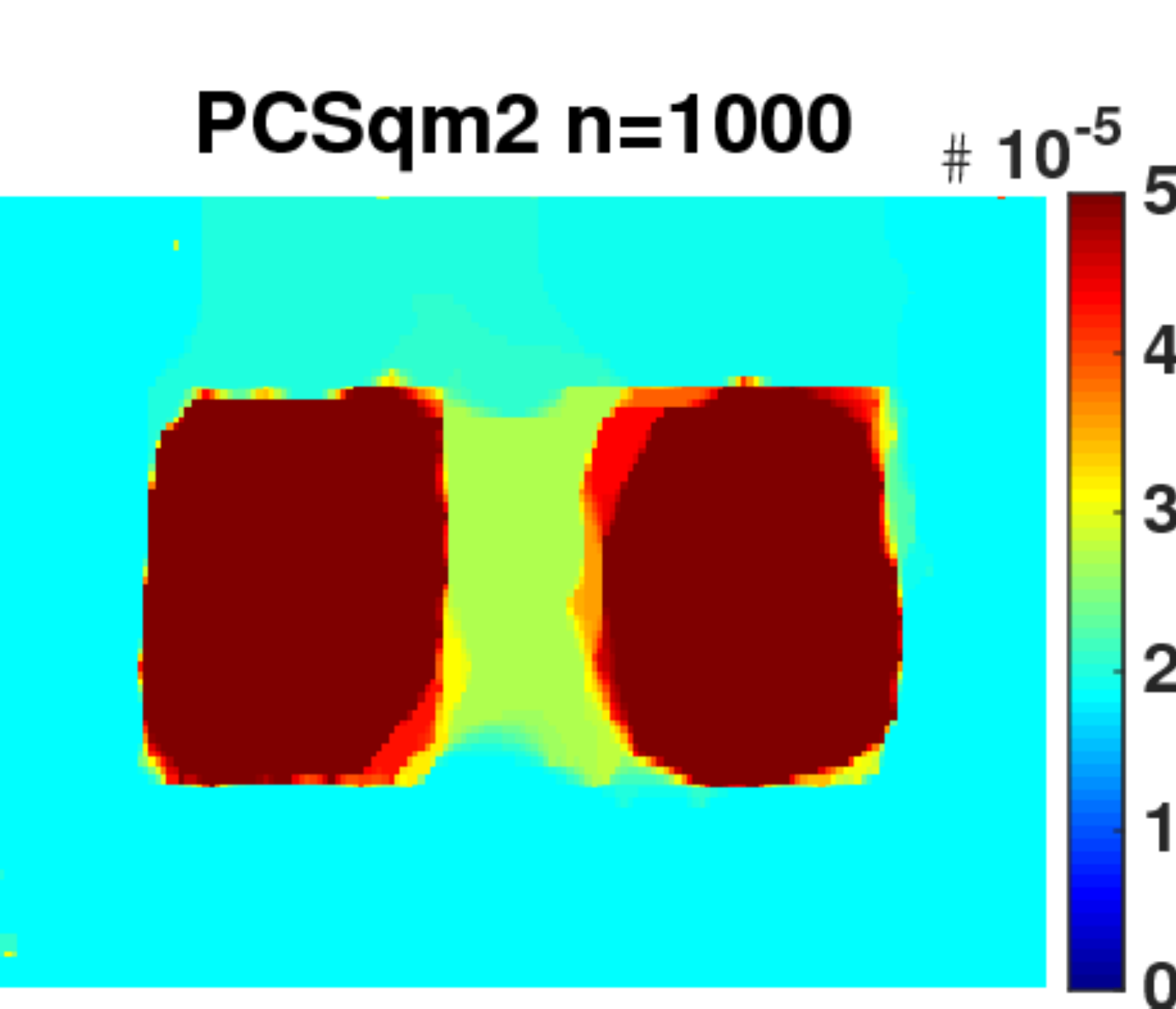}
\end{tabular}
\caption{TV MPLE applied to example embeddings of PCS $n=1000$, $k$ dense, and top left : LE, top right : LOE BFGS maxIt=100, bottom left : ASAP LOE BFGS max patch size 400, bottom right : estimated density from ground truth points, see column 2 of Figure~\ref{fig:X}}
\label{fig:TVPCS}
\end{figure}

\subsection{Network of network scientists embedding}
\label{sec:netsci2010}
To further illustrate potential of ordinal embedding to broad categories of data, we present here an experiment embedding data that does not have an apparent ground-truth geometry.  We use data from a co-authorship network of network scientists \cite{mapequation} from 2010, which was studied in \cite{rombach2014core} to evaluate methods of computing core-periphery structure.  The network contains nonegatively weighted undirected edges where the weights are based on the number of papers they have co-authored.  The network has 552 nodes and 1318 edges, with the number of edges attached to each node ranging from 1 to 38.  The mean number of edges attached to each node is 4.7754 and the median is 4.

To embed the data, we treat the co-authorship links as nearest neighbor relationships.  In other words, if  X and  Y have authored papers together, but X has not authored any papers with  Z, we impose that the distance between X and Y should be smaller than the distance between X and Z.  We used LOE BFGS and ASAP LOE BFGS to perform these embeddings in 2D and 3D.  In this case, the LOE results were ultimately best with the 2D LOE BFGS 500 iteration embedding misplacing 754 of the 1318 nearest neighbor edges and the 3D LOE BFGS 500 Iteration embedding misplacing 272 of the nearest neighbor edges, while the 2D ASAP LOE BFGS mps 500 misplaced 910 edges, and the 3D ASAP LOE BFGS mps 500 misplaced 467 edges.  That being said, several of the runtimes for the ASAP LOE results beat the LOE results.  We speculate that the reasons LOE outperforms ASAP LOE in accuracy in this case are twofold : 1) the number of nodes, $n=552$, is too small to make the LOE method applied to the full data sufficiently intensive, and 2) the wide distribution of degrees of the nodes in the network perhaps does not go well with our approach of breaking up the network via spectral clustering.  Perhaps other methods for braking up the network should be considered when the degree distribution is highly varied.

\begin{figure}[H]
  \center
  \includegraphics[width=0.9\columnwidth]{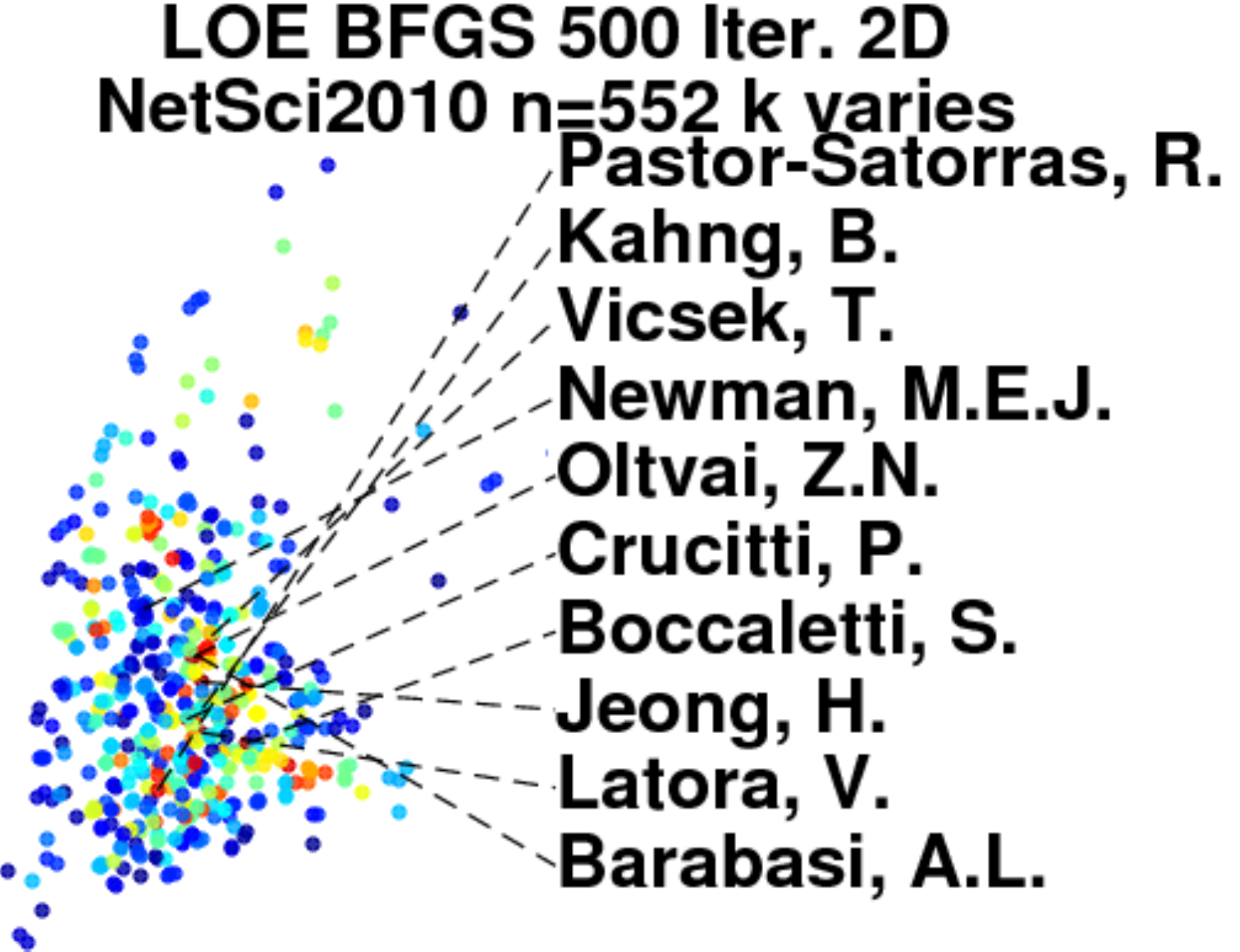}
\vspace{-3mm}
\caption{LOE BFGS 2d embeddings of data from NetSci2010 data set, $n = 552$, where co-authorship imposes that authors should be close}
\vspace{-3mm}
\label{fig:netsci2010_2d}
\end{figure}

\begin{figure}[H]
  \center
  \includegraphics[width=0.9\columnwidth]{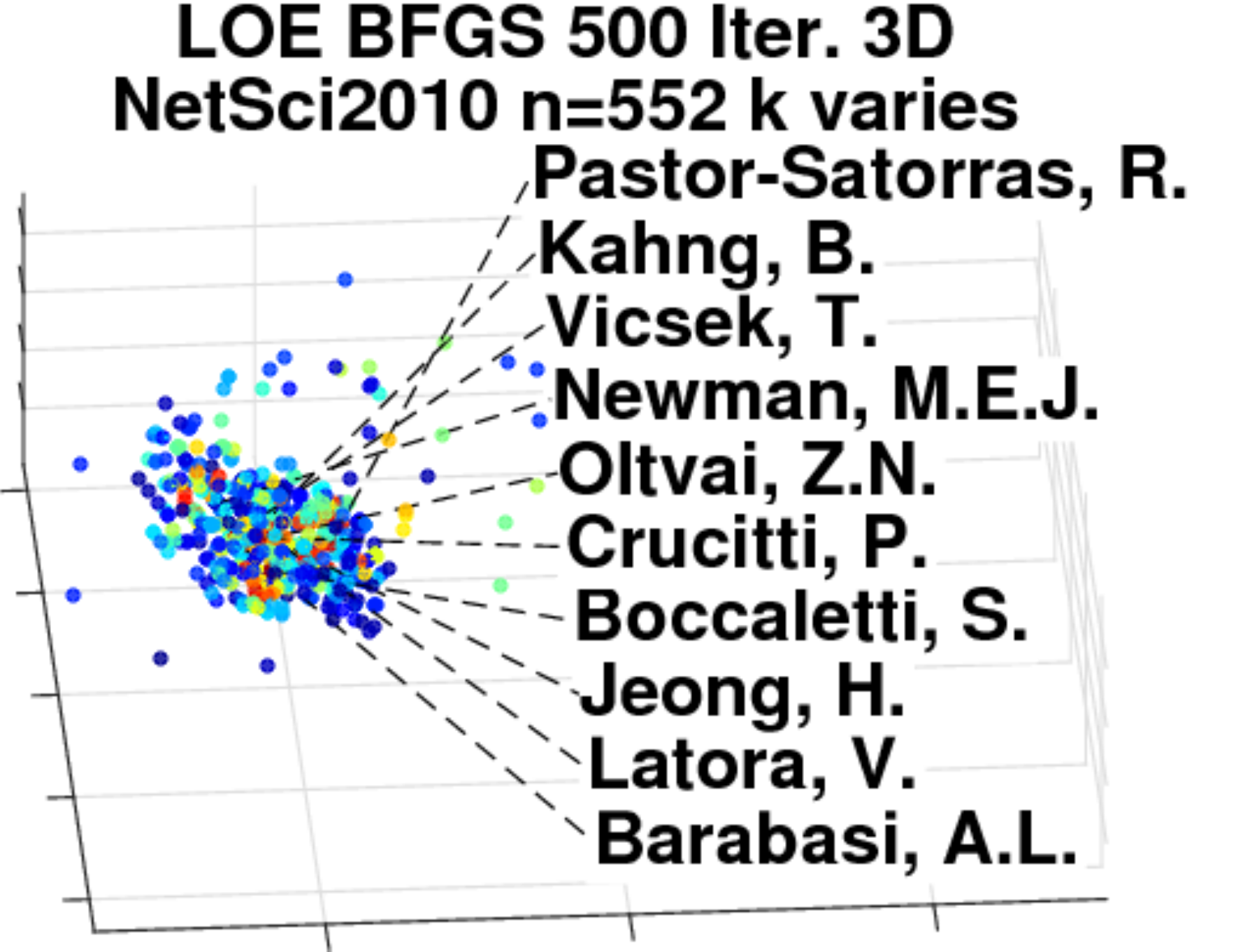}
\vspace{-3mm}
\caption{LOE BFGS 3d embeddings of data from NetSci2010 data set, $n = 552$, where co-authorship imposes that authors should be close}
\vspace{-3mm}
\label{fig:netsci2010_3d}
\end{figure}

Independent of the comparison of the two methods, we look at the best 2D and 3D embeddings from LOE (shown in Figure~\ref{fig:netsci2010_2d} and Figure~\ref{fig:netsci2010_3d} respectively), to see if the embeddings preserve any interesting structure in the network.  Since the network was previously studied for core-periphery detection, we color the nodes based on the corescore computed by the method proposed in \cite{rombach2014core} (mapping low values to blue and high values to red), and label the names of the authors with the top 10 corescores.  These red, core authors appear primarily central to the embeddings, suggesting that these embeddings preserve important structural properties in the original network.

\section{A Linear Program Alternative to SDP embedding}
\label{sec:lp}

In this section we present the algorithm and a few results for a Linear Program Embedding approach using metric MDS (LPEm) for ordinal embedding.  Though the results are ultimately not competitive with Local Ordinal Embedding, the approach is different enough so that the ideas may be of independent interest.  
In contrast to the SDP methods which cast embedding problems in terms of the Gram matrix $K$ our LPEm approach for kNN embedding optimizes over the variables $D$ (the distance matrix), $R$ (the radius at each node), and the slack variables.  The radius at each node $i$, denoted by $R_i$ is defined to be the distance between node $i$ and its $k$-th closest neighbor.  Thus $R_i$ is the radius of the neighborhood at node $i$.  In kNN embedding, the objective and constraints can be written as linear constraints in $D,R$ and the slack variables, 
altogether leading to a linear program which is computationally cheaper to solve than an SDP. Although SDP-based methods can encompass a larger class of problems, they currently do not approach the scalability or numerical maturity of LP and SOCP solvers.

After the LP returns a candidate distance matrix $D$ and radii $R$, we pass $D$ into a standard mdscale, here using metric multidimensional scaling (see Algorithm~\ref{alg:lp}),
\begin{algorithm}
\begin{equation}
	\begin{aligned}
	(D^*,R^*)=& \underset{\alpha, \beta,  R , D}{\argmin}  
	    & & \sum_{ij \in E(G)} \alpha_{ij} + \sum_{ij \notin E(G)} \beta_{ij} \\
	  & \mbox{subject to}
    & & \alpha,\beta\in \mathbb{R}_{+}^{n\times n}, R\in\mathbb{R}_{+}^{n}, D\in \mathbb{R}_{+,sym}^{n\times n}\\
	& & & D_{ij} \leq  R_i + \alpha_{ij}, \mbox{ if } ij \in E(G) \\
	& & &  D_{ij}  > R_i - \beta_{ij}, \mbox{ if } ij \notin E(G)\nonumber \\
    & & & \sum_{i=1}^{n} R_i = V \\
     & & & D_{ij} + D_{ik} \leq D_{kl}, (i,j,k) \in \mathcal{T}   \\	
X=&\mbox{mds}\left(D^*,d\right)
\end{aligned}
\end{equation}
\caption{LP approach}
\label{alg:lp}
\end{algorithm}
where by $\mathcal{T}$ we mean the set of triangle inequalities we considered (ordered set $(i,j,k)$). If $(i,j,k) \in \mathcal{T}$, the same holds true for the two other permutations.  The full set of triangle inequalities are necessary, though not sufficient, for the matrix $D$ to correspond to an Euclidean distance matrix.
If one omits slack variables, there are $n(n-1)/2$ distance values to solve for along with $n$ radii, and thus $n(n+1)/2$ unknowns in total.  Considering the ordinal constraints, for the upper bounds on the entries $D_{ij}$, there are $n$ ways to choose $i$, and for each $i$ there are $k$ ways to choose $j$, thus $nk/2$ constraints (accounting for symmetric distances).  For the lower bounds on the entries $D_{ij}$ there are $n$ ways to choose $i$ and for each $i$ there are $n-k-1$ ways to choose $j$, giving $n(n-k-1)/2$ constraints.  So there are $n(n-1)/2$ ordinal constraints on relating the $n(n-1)/2$ distances and $n$ radii.  In other words, the intuition behind the added triangle inequalities is that they help to better constrain the system.  There are on the order of $n^3$ triangle inequalities (choose any three points), so for large $n$, there are many more constraints than unknowns.  

To avoid the added complexity from imposing all triangle inequalities, one could consider models that impose only a fraction of such constraints via either imposing them locally, for $k$-hop neighboring triples of points,  or globally, such as picking edges via an Erd\H{o}s-R\'{e}nyi model, or mixing the two approaches.  

We remark that dropping triangle inequalities altogether could certainly speed up the embedding process.  The resulting non-metric $D$ may correspond to an increasing function of distance (e.g., distance squared), which suggests that non-metric MDS would be appropriate.

In general, even if the recovered distance metric corresponds to a metric distance, this is not a guarantee that the distance is realizable in a low-dimensional space.  That requires a rank constraint on $D$, which is non-convex and is computationally intractable for an LP or SDP.  The ultimate embedding into a low-dimensional space thus potentially gives up some structure in both the LP and SDP formulation, and it can be argued that this effect is lessened  via the local to global approach.

\begin{figure}[H]
  \center
  \includegraphics[width=\thirdcolwidth]{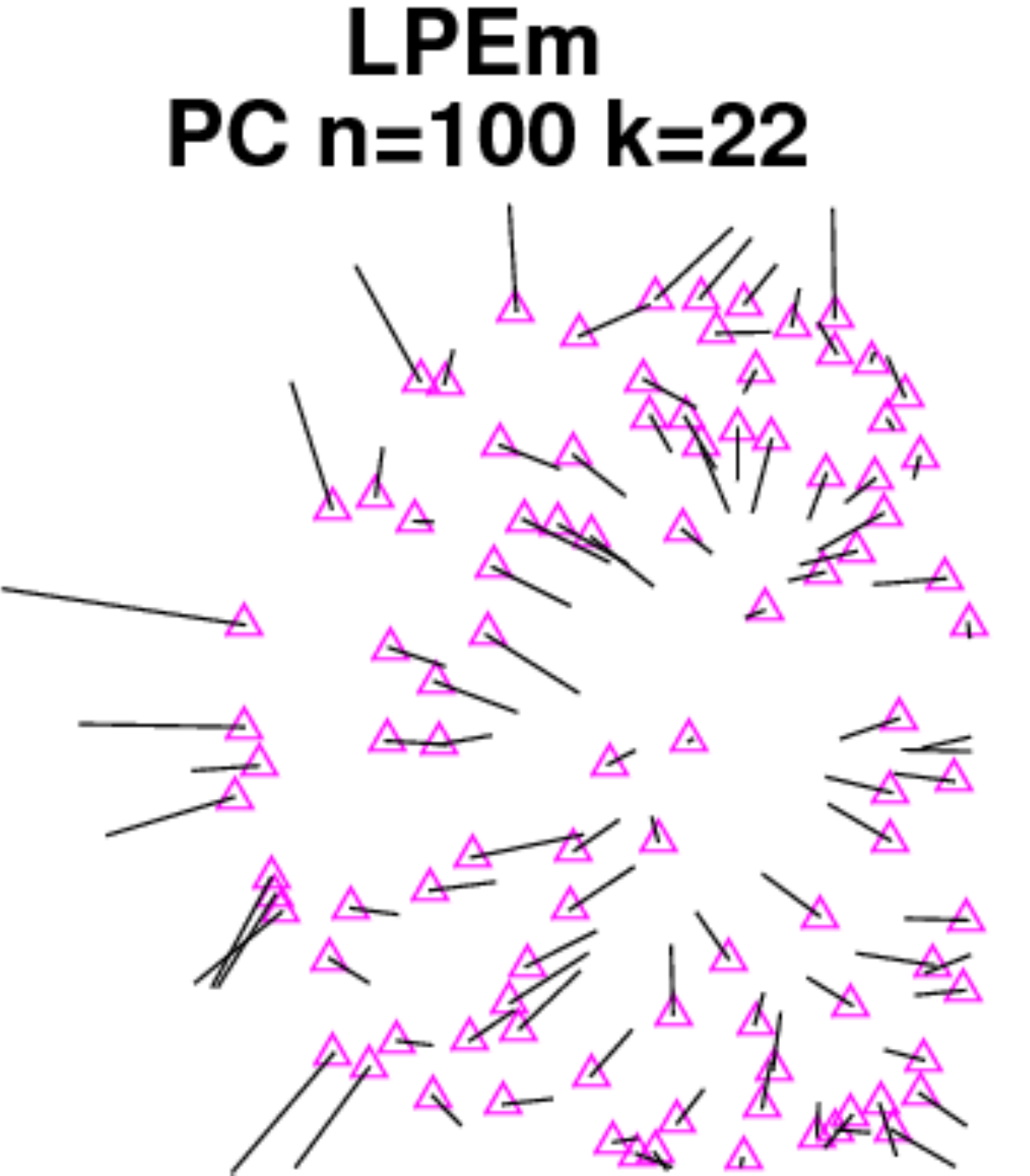}
  \includegraphics[width=\thirdcolwidth]{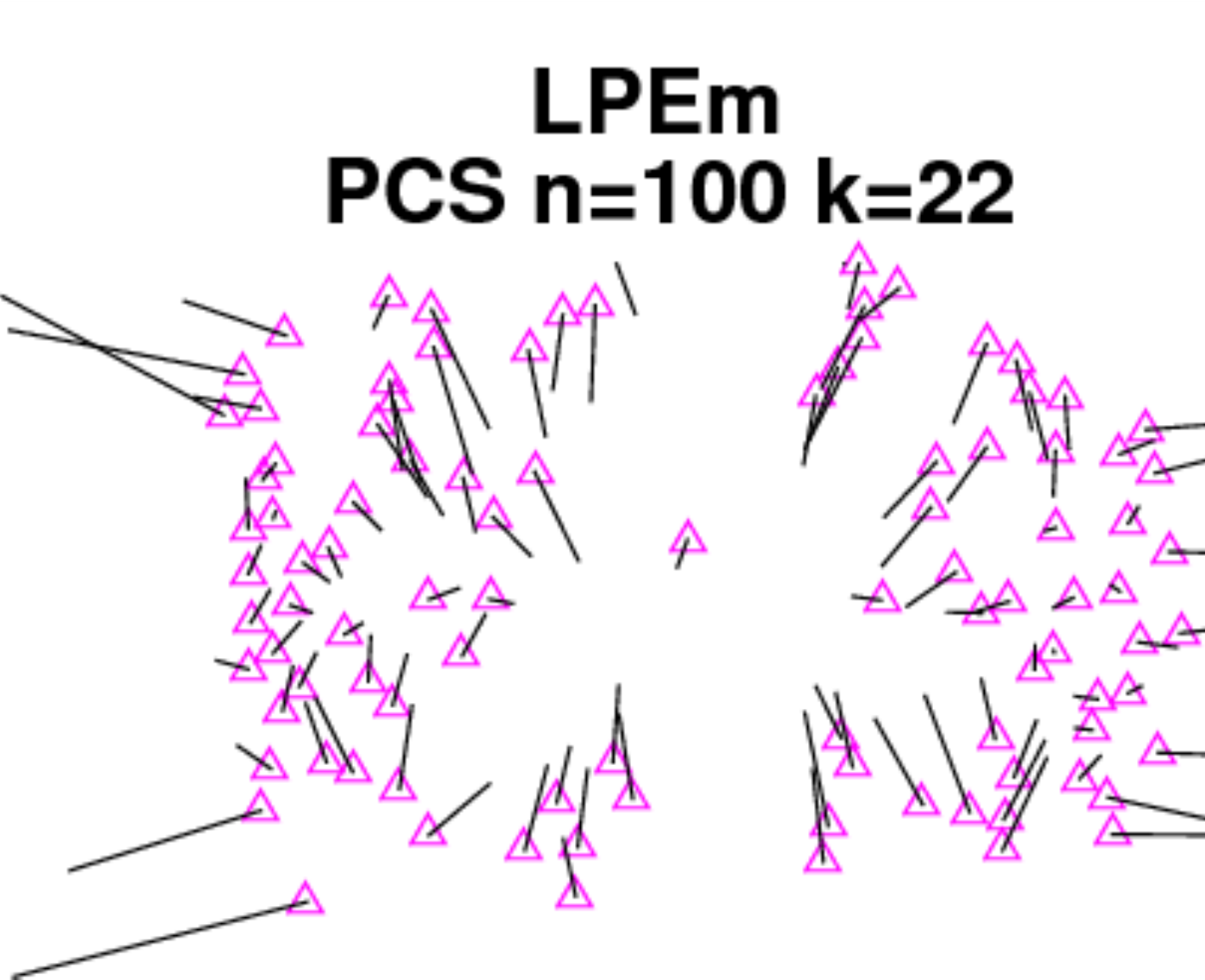}
  \includegraphics[width=\thirdcolwidth]{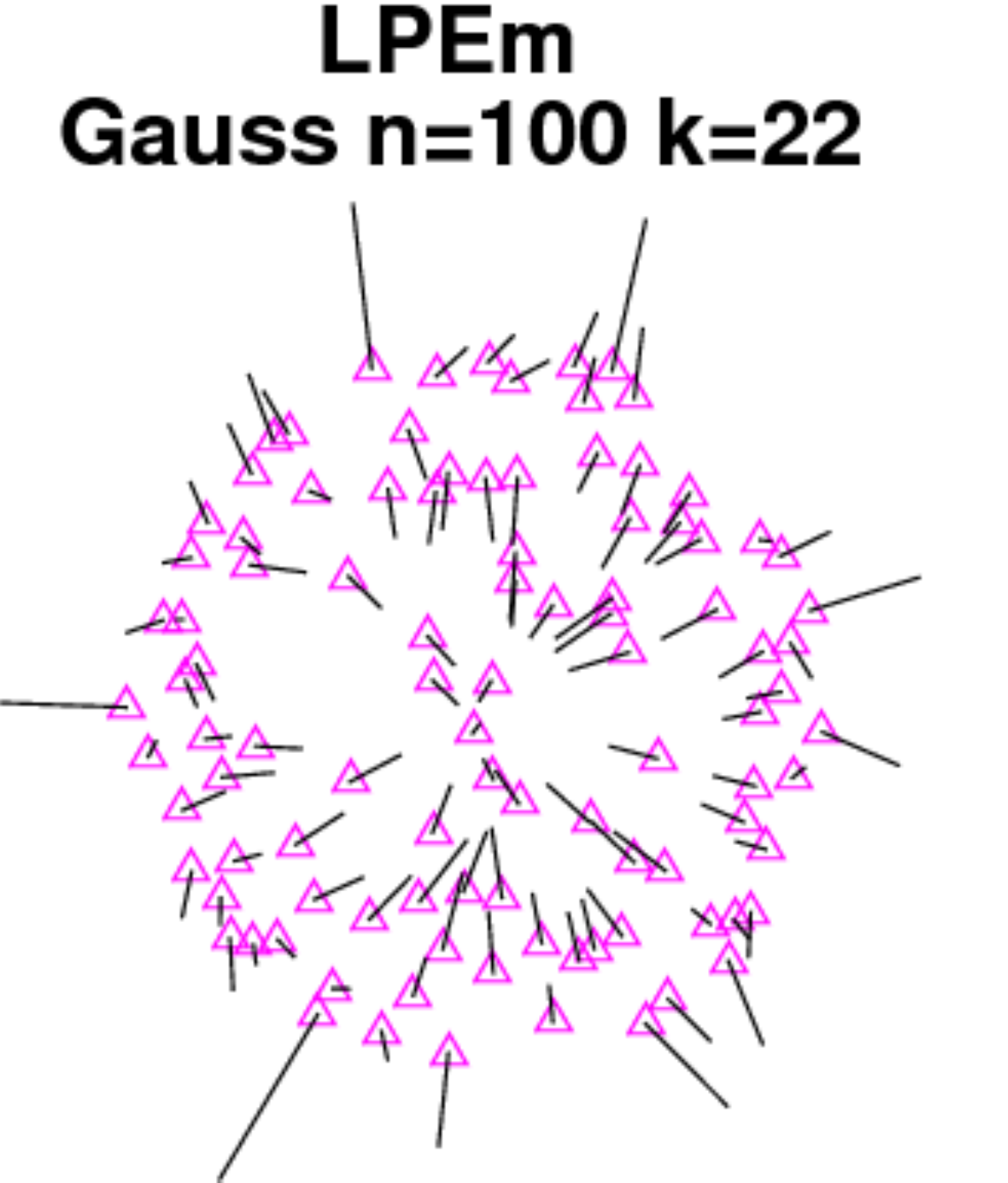}\\
  \includegraphics[width=\thirdcolwidth]{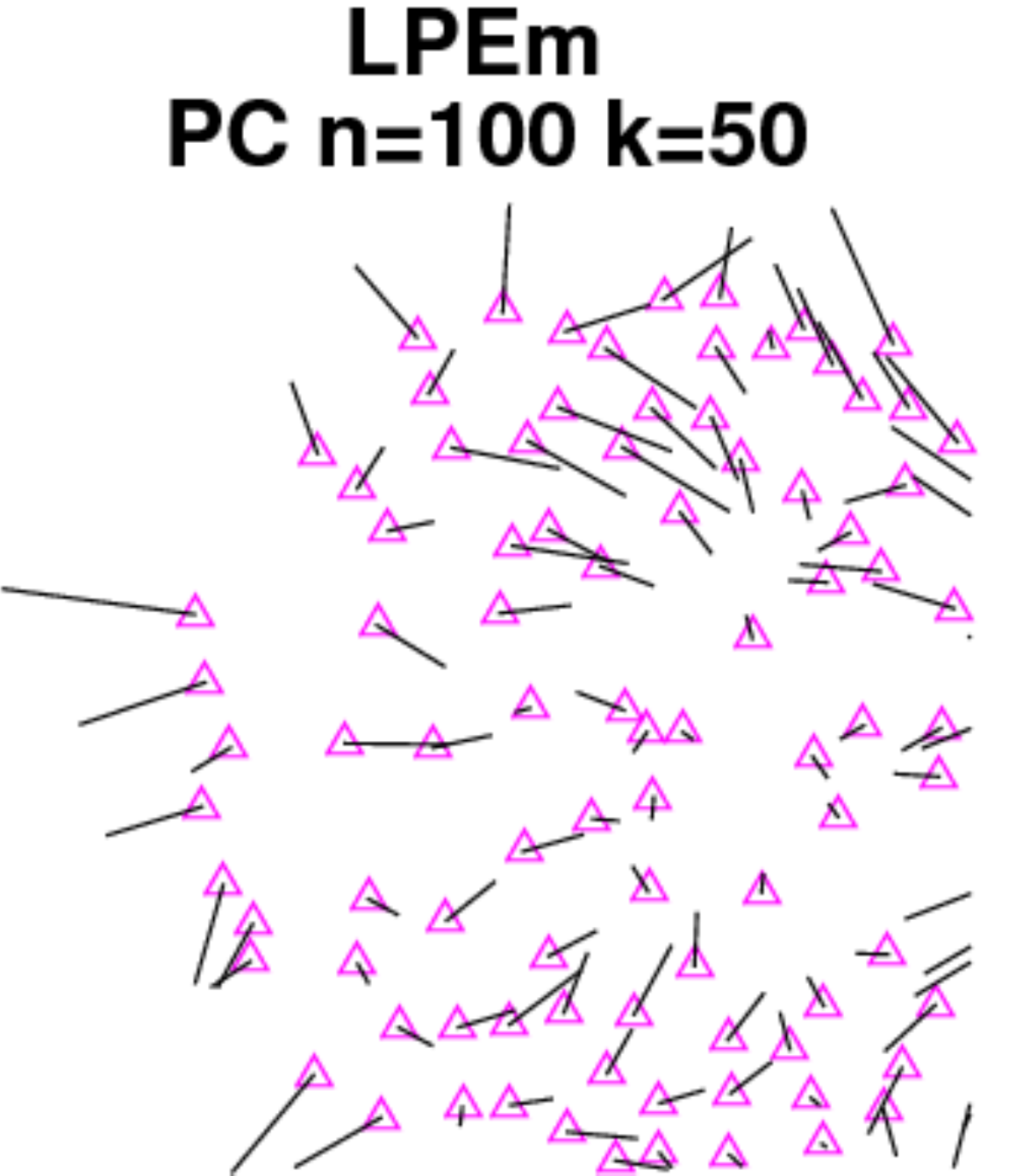}
  \includegraphics[width=\thirdcolwidth]{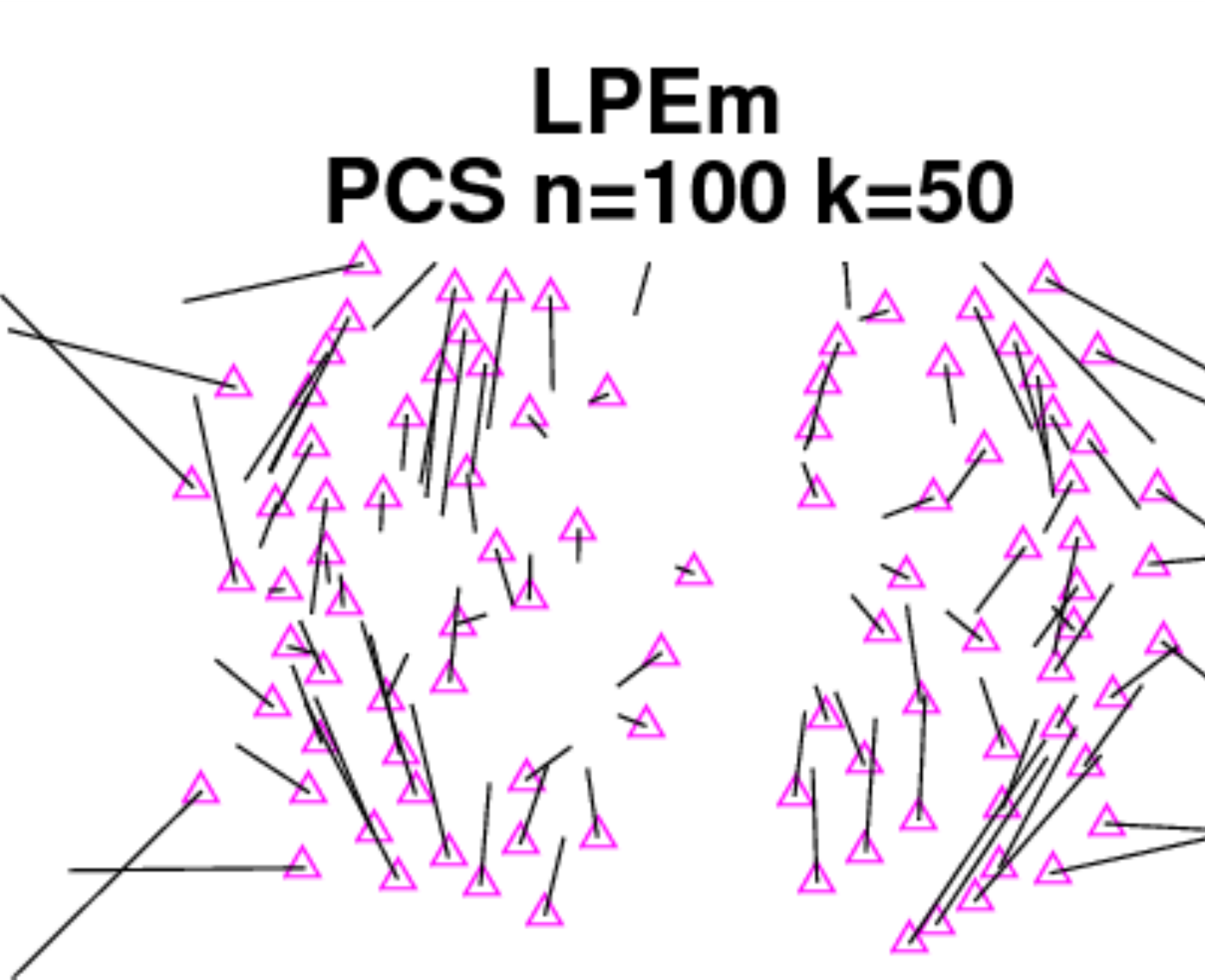}
  \includegraphics[width=\thirdcolwidth]{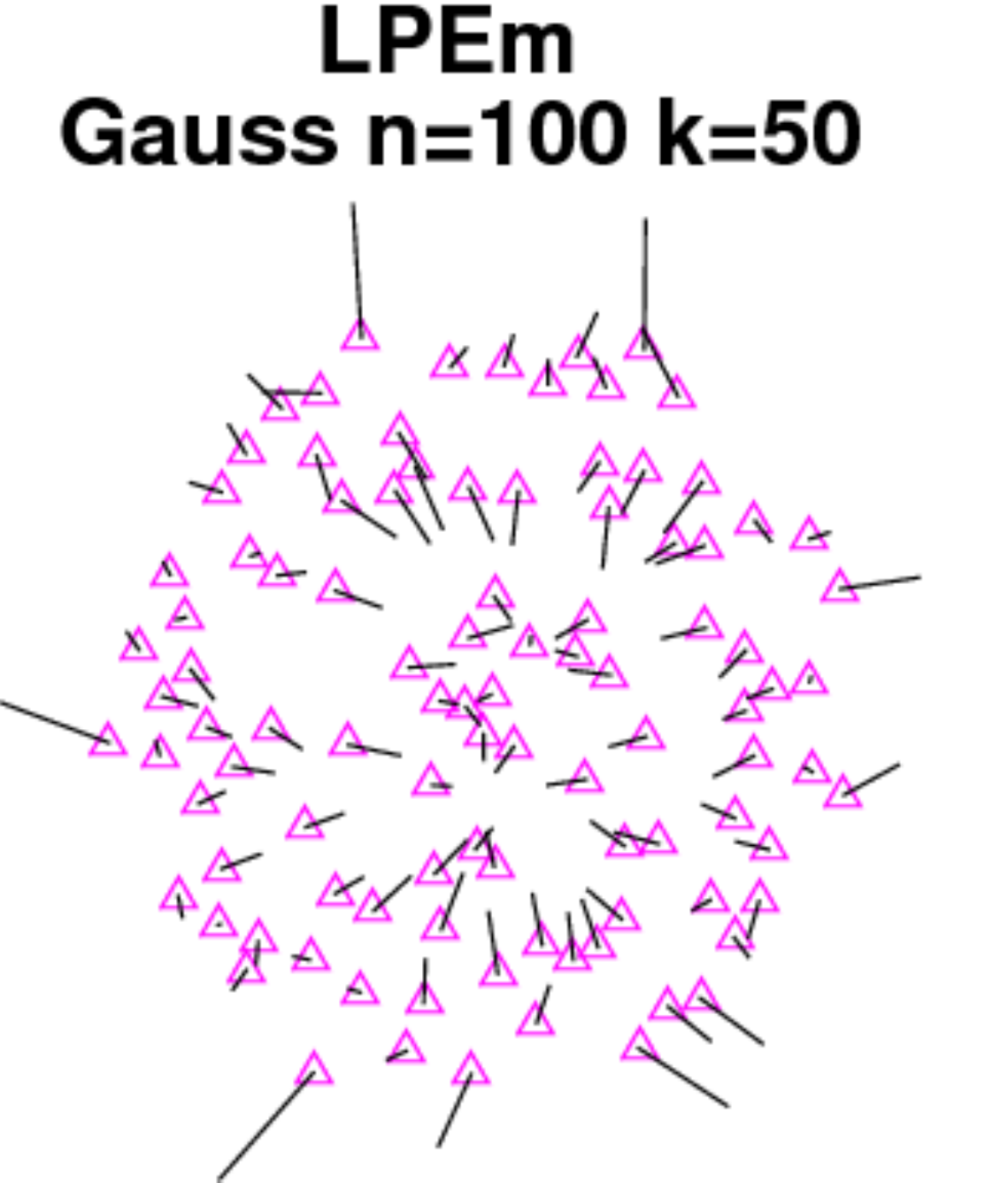}\\
  \includegraphics[width=\thirdcolwidth]{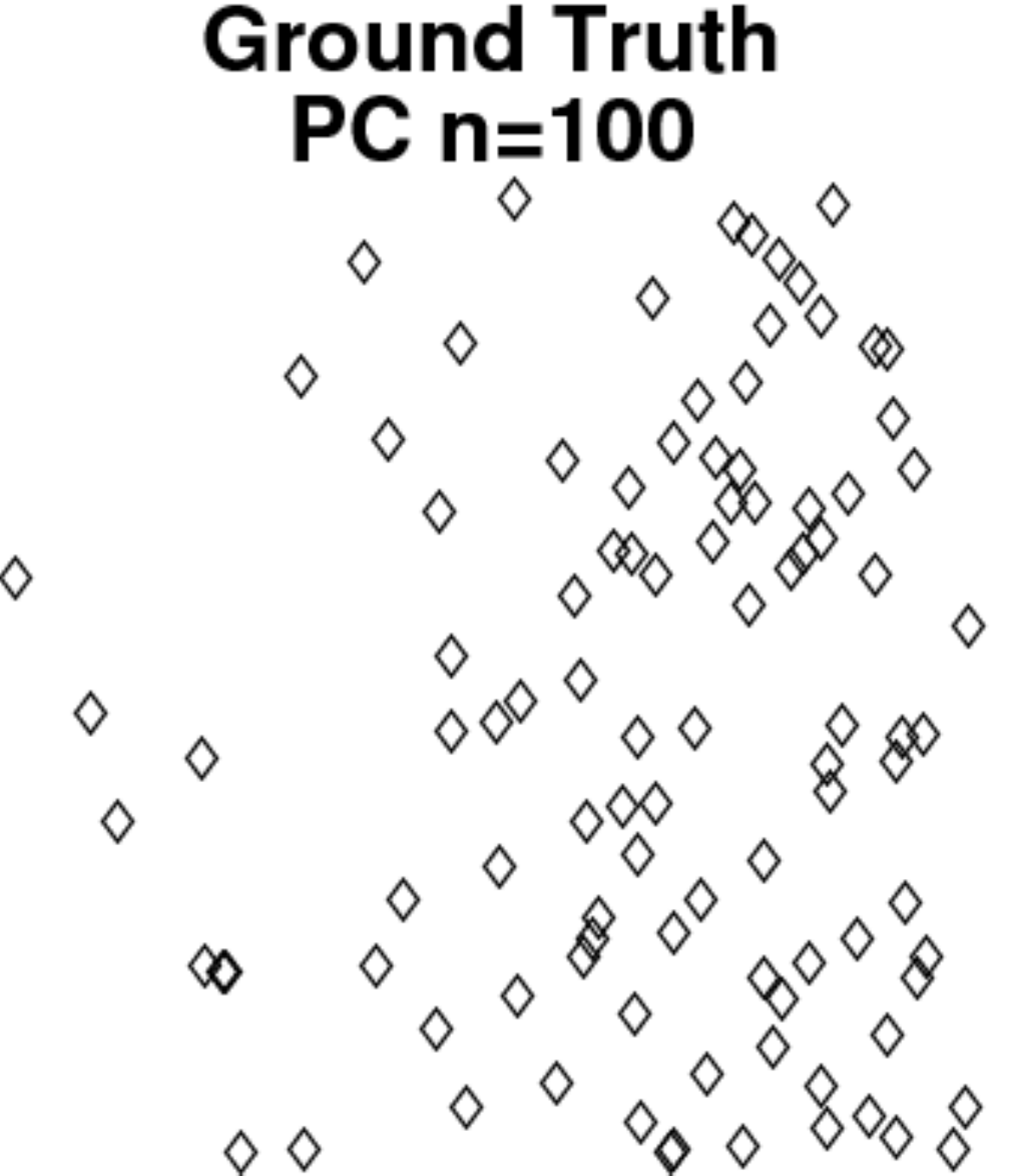}
  \includegraphics[width=\thirdcolwidth]{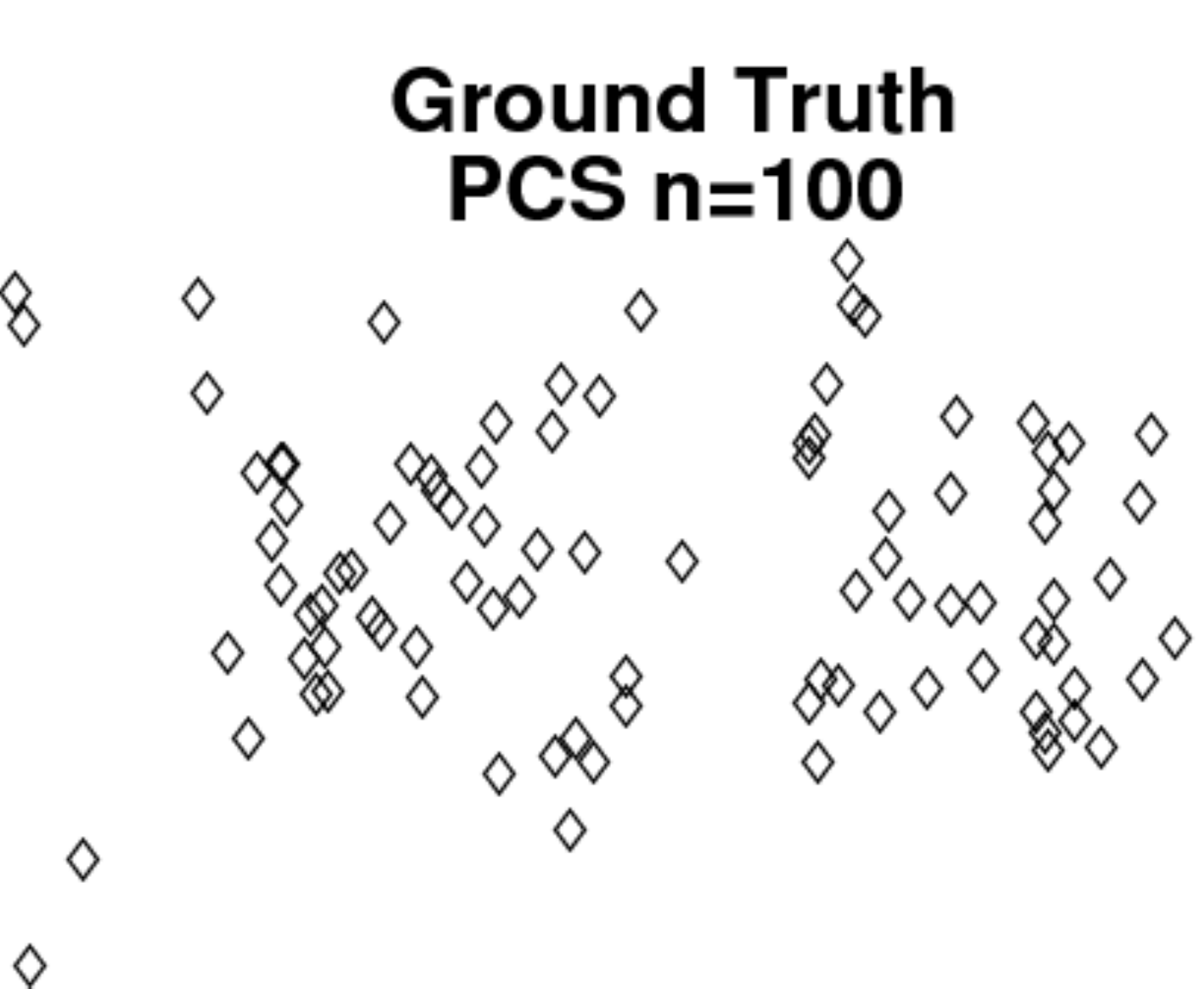}
  \includegraphics[width=\thirdcolwidth]{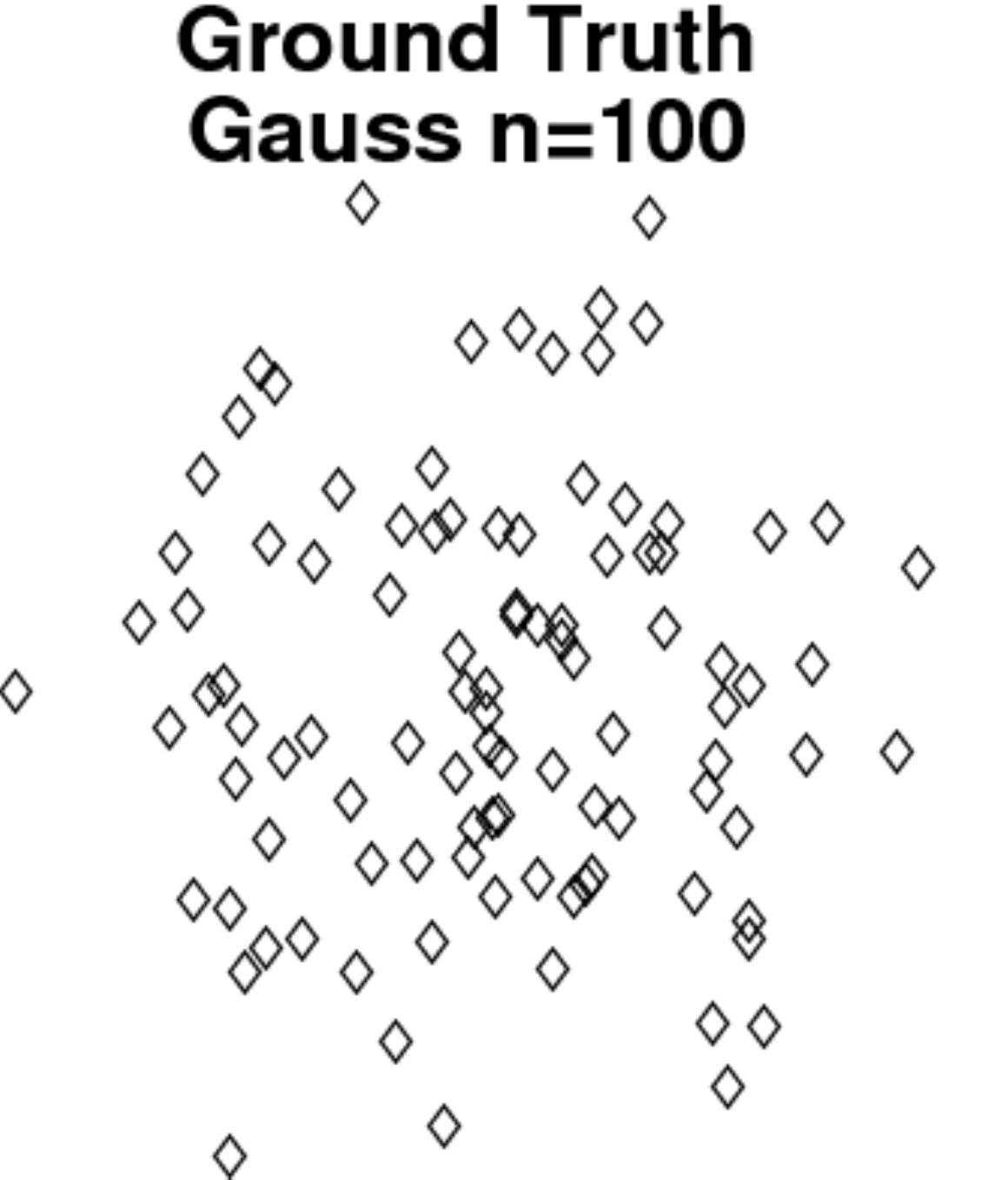}
\vspace{-3mm}
\caption{Linear Program Embeddings for the PC (left), PCS (middle), and Gauss (right) data sets with $n=100$,
Row 1 : $k = 22$
Row 2: $k = 50$
Row 3: ground truth.
Line segments highlight the displacement of each point.}
\vspace{-3mm}
\label{fig:LPEmX}
\end{figure}

In Figure~\ref{fig:LPEmX} we show an example with points drawn from the densities discussed in the previous section along with points embedded using the LPEm approach.  In these experiments we use a very dense value of $k$, $k = n/2 = 50$, which is where the approach seemed to work the best.  The recovery of the piecewise constant half-planes is the best, but the preliminary results led us to decide not to experiment with this method further for the time being.  The method was implemented using the CVX library, a package for specifying and solving convex programs  (\cite{cvx, gb08}).  Overall, we find the LP formulation appealing due to its simplicity.  It would be interesting if a similarly simple approach could obtain competitive results on the problem of ordinal embedding, especially since until the work of von Luxburg and Alamgir \cite{von2013density},  it was unknown to the community whether the problem was practically solvable at all.

\FloatBarrier

\vspace{-4mm}
\section{Summary and discussion}
\label{sec:conclusion}
\vspace{-3mm}
We have demonstrated that the computational efficiency of LOE for the kNN embedding problem can be significantly improved, while maintaining and often improving spatial and ordinal
accuracy in a distributed setting.  Our application of the 
divide-and-conquer ASAP method renders the problem of kNN embedding 
significantly more tractable, distributing the embedding steps, and using fast spectral methods to combine them.  We expect that such improvements will make it possible to use kNN embeddings in a broader range of 
settings, and that the ASAP framework will be of independent interest to the machine learning community for tackling large geometric embedding problems.

\section*{Acknowledgements}
The authors would like to thank Andrea Bertozzi for her support throughout this project. 
M.C. acknowledges support from AFOSR MURI grant FA9550-10-1-0569.  J.W. acknowledges support from NSF grant no. DGE-1144087 and the UC Lab Fees Research grant 12-LR-236660.  
Part of this work was undertaken while M.C. and J.W. were attending the Semester Program on Network Science and Graph Algorithms at the Institute for Computational and  Experimental Research in Mathematics (ICERM) at Brown University - we thank our hosts for their warm hospitality. We would also like to thank Andrea Bertozzi for organizing the research cluster at ICERM, 
Ulrike von Luxburg and Mauro Maggioni for helpful conversations, and Yoshikazu Terada for providing the LOE code.

\subsection{Rigidity Theory Appendix}
\label{sec:rigidity}

One of the main questions in the field of rigidity theory asks whether one can uniquely determine (up to rigid transformations, such as translations, rotations, reflections) the coordinates of a set of points $p_1, \ldots, p_n$ given a partial set of distances $d_{ij}=||p_i-p_j||$ between $n$ points in $\mathbb{R}^d$.
To make our paper self-contained, this short appendix if a very brief summary of the main definitions and results related to local and global rigidity from the literature (e.g., \cite[and references therein]{coning,Harvard1,conditions_UGR,Roth1981}). Readers who are unfamiliar with rigidity theory may use this short Appendix as a glossary. As previously discussed in Section~\ref{sec:breakintopatches}, one of the steps of the divide-and-conquer approach proposed for the kNN-recovery problem relies to testing whether the underlying resulting patches are globally rigid. As observed in our numerical simulations detailed in Figures~{\ref{fig:AerrPC},\ref{fig:AerrPCS}, \ref{fig:AerrGauss}, \ref{fig:AerrHalfcube}, \ref{fig:X}} the final reconstruction is more accurate when we  rely on global rigidity as a postprocessing step for the partitions obtained via spectral clustering. The intuition behind our approach is as follows. In the case when distance information is available, testing for global rigidity is a crucial step in making sure that each of the local patches has a unique embedding in its own reference frame, approximatively consistent with the ground truth, up to a rigid transformation. Since in the kNN-recovery problem, we do not have distance information but only ordinal data, thus we are faced with solving even a harder problem, we expect that the global rigidity check will improve the accuracy of the local patch embeddings.  One specific example where our current rigidity heuristics improved results was in performing ASAP LOE BFGS with max patch size 300, on $n=5000$ points drawn from the constant half-plane distribution, letting $k=18$.  In that example, performing the rigidity check and pruning gave a runtime of 107.056 s, an ordinal error of 0.00107096, and 0.0585465 Procrustes error, while skipping the rigidity check and pruning gave a runtime of 192.606 s, an ordinal error of 0.00154208 A error, and 0.175992 Procrustes error.

A {\em bar and joint framework} in $\mathbb{R}^d$ is defined as an undirected graph $G=(V,E)$ ($|V|=n, |E|=m$) together with a {\em configuration} $p$ which assigns a point $p_i$ in $\mathbb{R}^d$ to each vertex $i$ of the graph. The edges of the graph correspond to distance constraints, that is, $(i,j)\in E$ if an only there is a bar of length $d_{ij}$ between points $p_i$ and $p_j$. 
We say that a framework $G(p)$ is {\em locally rigid} if there exists a neighborhood $U$ of $G(p)$ such that $G(p)$ is the only framework in $U$ with the same set of edge lengths, up to rigid transformations. In other words, there is no continuous deformation that preserves the given edge lengths. A configuration is {\em generic} if the coordinates do not satisfy any non-zero polynomial equation with integer coefficients (or equivalently algebraic coefficients). 

Local rigidity in $\mathbb{R}^d$ has been shown to be a generic property, in the sense that either all generic frameworks of the graph $G$ are locally rigid, or none of them are. 
A consequence of the seminal results of Gluck \cite{Gluck} and Asimow and Roth \cite{Asimow} asserts that the dimension of the null space of the rigidity matrix is the same at every generic point, and hence local rigidity in $\mathbb{R}^d$ is a generic property, meaning that either all generic frameworks of the graph $G$ are locally rigid, or none of them are.  With probability one, the rank of the rigidity matrix that corresponds to the unknown true displacement vectors equals the rank of the randomized rigidity matrix. A similar randomized algorithm for generic local rigidity was described in \cite[Algorithm 3.2]{Harvard1}.
In other words, {\em generic local rigidity} in $\mathbb{R}^d$ can be considered a combinatorial property of the graph $G$ itself, independent of the particular realization. 
Using this observation, generic local rigidity can therefore be tested efficiently in any dimension using a randomized algorithm \cite{conditions_UGR}: one can just randomize the displacement vectors $p_1,\ldots,p_n$ while ignoring the prescribed distance constraints that they have to satisfy, construct the so called \textit{rigidity matrix} corresponding to the framework of the original graph with the randomized points and check its rank. This is approach we use to make sure the obtained patches are local rigid.

Since local generic rigidity does not imply unique realization of the framework, it is possible that there exist multiple non-congruent realizations that satisfy the prescribed distances (which we do not even have available in the kNN recovery problem)
One may consider for example, the $2D$-rigid graph with $n=4$ vertices and $m=5$ edges consisting of two triangles that can be folded with respect to their joint edge. We call a framework $G(p)$  {\em globally rigid} in $\mathbb{R}^d$ if all frameworks $G(q)$ in $\mathbb{R}^d$ which are $G(p)$-equivalent (have all bars the same length as $G(p)$) are congruent to $G(p)$ (i.e., related by a rigid transformation).
Hendrickson proved two key necessary conditions for global rigidity of a framework at a generic configuration:
\begin{theorem}[Hendrickson \cite{conditions_UGR}] \label{Th-Hendrickson}
If a framework $G(p)$, other than a simplex, is globally rigid for a generic configuration $p$ in $\mathbb{R}^d$ then:
\begin{itemize}
\item The graph $G$ is vertex $(d + 1)$-connected;
\item The framework $G(p)$ is edge-2-rigid (or, redundantly rigid), in the sense that removing any one edge leaves a graph which is infinitesimally rigid.
\end{itemize}
\end{theorem}

We say that a graph $G$ is {\em generically globally rigid} in $\mathbb{R}^d$ if $G(p)$ is globally rigid at
all generic configurations $p$ \cite{Connelly3,Connelly}. 
Though it has been conjectured for many years that global rigidity is a generic property, this fact was shown to be true only very recently. The seminal work of \cite{Connelly,Harvard1} proves that global rigidity is a generic property of the graph in each dimension. The conditions of Hendrickson as stated in Theorem \ref{Th-Hendrickson} are necessary for generic global rigidity. They are also sufficient on the line, and in the plane \cite{JacksonJordan}. However, by a result of Connelly \cite{Connelly3}, $K_{5,5}$ in 3-space is generically edge-2-rigid and 5-connected but is not generically globally rigid.

One of the tools used in testing for global rigidity of frameworks relies on the notions on stress matrices, more popular perhaps in the engineering community. 
A {\em stress} is defined an assignment of scalars $w_{ij}$ to the edges of the given graph $G$ such that for every node $i\in V$ it holds that 
\begin{equation}
\sum_{j:\,(i,j)\in E} \omega_{ij} (p_i-p_j) = 0.
\end{equation}
Alternatively, it can be show that a stress is a vector $w$ in the left null space of the rigidity matrix: $R_G(p)^T w = 0$. A stress vector can be rearranged into an $n\times n$ symmetric matrix $\Omega$, known as the {\em stress matrix}, such that for $i\neq j$, the $(i,j)$ entry of $\Omega$ is $\Omega_{ij}=-\omega_{ij}$, and the diagonal entries for $(i,i)$ are $\Omega_{ii} = \sum_{j:\, j\neq i} \omega_{ij}$.
Since all row and column sums are zero, it follows that the all-ones vector $(1 \; 1 \; \cdots \; 1)^T$ is in the null space of $\Omega$ as well as each of the coordinate vectors of the configuration $p$. Therefore, it follows that for generic configurations the rank of the stress matrix is at most $n-(d+1)$. 
The following pairs of theorems give sufficient and necessary conditions for generic global rigidity:

\begin{theorem}[Connelly \cite{Connelly}] \label{Th-Connelly}
If $p$ is a generic configuration in $\mathbb{R}^d$, such that
there is a stress, where the rank of the associated stress matrix
$\Omega$ is $n-(d+1)$, then $G(p)$ is globally rigid in $\mathbb{R}^d$.
\end{theorem}

\begin{theorem}
[Gortler, Healy, and Thurston \cite{Harvard1}] \label{Th-Harvard}
Suppose that $p$ is a generic configuration in $\mathbb{R}^d$, such that $G(p)$ is globally rigid in $\mathbb{R}^d$. Then either $G(p)$ is a simplex or there is a stress where the rank of the associated
stress matrix $\Omega$ is $n-(d+1)$.
\end{theorem}

Based on the latter theorem, the authors of \cite{Harvard1} also provided a randomized polynomial algorithm for checking generic global rigidity of a graph \cite[Algorithm 3.3]{Harvard1}, which we use to test for global rigidity of the patches in the kNN-recovery problem. If a given patch is generically locally rigid then their algorithm picks a random stress vector of the left null space of the rigidity matrix associated to this patch, and converts it into a stress matrix. If the rank of the stress matrix is exactly $n-(d+1)$, then we conclude that the patch is globally rigid, and if the rank is lower, then the respective patch is not globally rigid.

\FloatBarrier

\vspace{-3mm}
\bibliography{bib}
\bibliographystyle{IEEEbib}

\end{document}